\documentclass[runningheads]{llncs}

\usepackage{eccv}

\usepackage{eccvabbrv}

\usepackage{url}
\usepackage{microtype}
\usepackage{nicefrac}
\usepackage{lipsum}

\usepackage{graphicx}
\usepackage{subcaption}
\usepackage{wrapfig}
\usepackage{caption}
\usepackage{float}

\usepackage{booktabs}
\usepackage{multirow}
\usepackage{threeparttable}
\usepackage{arydshln}
\usepackage[table]{xcolor}

\usepackage{amsmath}
\usepackage{amssymb}
\usepackage{amsfonts}
\usepackage{bm}
\usepackage{dsfont}
\DeclareMathOperator*{\argmin}{arg\,min}

\usepackage{algorithm}
\usepackage{algorithmic}
\usepackage{listings}
\usepackage{verbatim}
\usepackage{comment}

\usepackage{pifont}

\usepackage{textcomp}
\usepackage{relsize}

\usepackage{cancel}

\usepackage{tikz}
\usetikzlibrary{decorations.pathreplacing, positioning}

\usepackage{appendix}
\usepackage{afterpage}

\usepackage{etoolbox}
\makeatletter
\AfterEndEnvironment{algorithm}{\let\@algcomment\relax}
\AtEndEnvironment{algorithm}{\kern2pt\hrule\relax\vskip3pt\@algcomment}
\let\@algcomment\relax
\newcommand\algcomment[1]{\def\@algcomment{\footnotesize#1}}
\renewcommand\fs@ruled{\def\@fs@cfont{\bfseries}\let\@fs@capt\floatc@ruled
  \def\@fs@pre{\hrule height.8pt depth0pt \kern2pt}%
  \def\@fs@post{}%
  \def\@fs@mid{\kern2pt\hrule\kern2pt}%
  \let\@fs@iftopcapt\iftrue}
\makeatother
\newcommand{\model}{DCD\xspace}

\usepackage{hyperref}
\usepackage[capitalize,nameinlink]{cleveref}

\usepackage[accsupp]{axessibility}

\begin{document}

\title{
Discriminative and Consistent \\Representation Distillation
}

\titlerunning{
Discriminative and Consistent Representation Distillation
}

\author{
Nikos Giakoumoglou \and
Tania Stathaki
}

\authorrunning{
N.~Giakoumoglou and T.~Stathaki
}

\institute{
Imperial College London \\
\email{\{nikos, tania\}@imperial.ac.uk} \\
Code: \url{https://github.com/giakoumoglou/rrd}
}

\maketitle


\begin{abstract}

Knowledge Distillation (KD) transfers knowledge from a large teacher to a smaller student model. While contrastive objectives have proven effective for learning structured representations in self-supervised settings, their use in distillation is hindered by two practical shortcomings: the reliance on external memory banks for negative sampling, and fixed temperature hyperparameters that limit adaptability across training stages and teacher-student pairs. We therefore propose \textbf{\underline{D}}iscriminative and \textbf{\underline{C}}onsistent Representation \textbf{\underline{D}}istillation (\model), which combines contrastive instance discrimination with a consistency regularization term over the cross-model similarity matrix. The contrastive term aligns each student representation with its teacher counterpart, while the consistency term penalizes asymmetry between the row-normalized and column-normalized views of that matrix, constraining the off-diagonal structure that instance discrimination alone leaves free; we show that it vanishes precisely when this matrix is symmetric. We further introduce an efficient in-batch sampling that eliminates external memory banks, and learnable scale and bias parameters that adapt during training to control the sharpness and offset of the distillation signal. The method matches the training speed of standard KD while adding only 66K additional parameters. Through extensive experiments on CIFAR-100, ImageNet, and MS-COCO, together with cross-dataset transfer to STL-10 and Tiny ImageNet, we show that our approach achieves competitive performance in classification, object detection, and transfer, while substantially reducing memory consumption and training time compared to existing contrastive distillation methods.

\keywords{Knowledge distillation \and Knowledge transfer \and Representation learning \and Contrastive learning }

\end{abstract}

\section{Introduction}
\label{sec:intro}

Knowledge Distillation (KD) enables the transfer of knowledge from large, high-capacity \textit{teacher} models to compact \textit{student} models~\cite{hinton2015distilling}. As state-of-the-art vision models for image classification~\cite{liu2021swin, dosovitskiy2021vit}, object detection~\cite{ren2016fasterrcnn, lin2017feature}, and semantic segmentation~\cite{chen2017deeplab, chen2017deeplabv3} continue to grow in size and computational cost~\cite{goyal2019scaling, kornblith2019do}, efficient model compression has become a practical necessity~\cite{bucilua2006model, polino2018model}. The foundational works of Bucilu\u{a} \etal~\cite{bucilua2006model} and Hinton \etal~\cite{hinton2015distilling} formulated distillation as minimizing the Kullback-Leibler (KL) divergence between teacher and student output distributions. While this is natural when the output is a categorical distribution over classes, it does not capture the richer internal knowledge encoded in intermediate representations, including visual semantics and inter-class relations.

Representational knowledge is inherently \textit{structured}: feature dimensions exhibit non-trivial correlations and higher-order dependencies that logit matching alone cannot preserve. To address this, feature-based methods~\cite{romero2014fitnets, zagoruyko2016paying, yim2017gift, peng2019correlation} extend distillation to intermediate layers. However, Tian \etal~\cite{tian2022crd} demonstrated that such approaches still fail to capture the full structural knowledge in the teacher's representations. Building on this observation, contrastive objectives~\cite{gutmann2010noise, oord2019cpc, hjelm2019dim} have been adapted for distilling structured knowledge between teacher and student networks~\cite{tian2022crd, fang2021seed}. By treating each instance as its own class and learning to discriminate between them, contrastive distillation can transfer fine-grained representational information that goes beyond what logit-based or feature-matching methods capture.

However, contrastive distillation in its current form faces notable \textbf{limitations}. First, existing contrastive distillation methods such as CRD~\cite{tian2022crd} require a large external memory bank holding one feature vector per training sample, introducing significant memory overhead and implementation complexity. Second, fixed temperature hyperparameters limit the adaptability of the contrastive objective across different training stages and teacher-student pairs~\cite{radford2021clip}. Third, instance-discrimination objectives supervise only the diagonal of the cross-model similarity matrix, leaving its off-diagonal structure unconstrained: nothing prevents the student from matching its own teacher counterpart while relating inconsistently to the remaining samples in the batch.

\begin{figure}[t]
    \centering
    \includegraphics[width=0.7\linewidth]{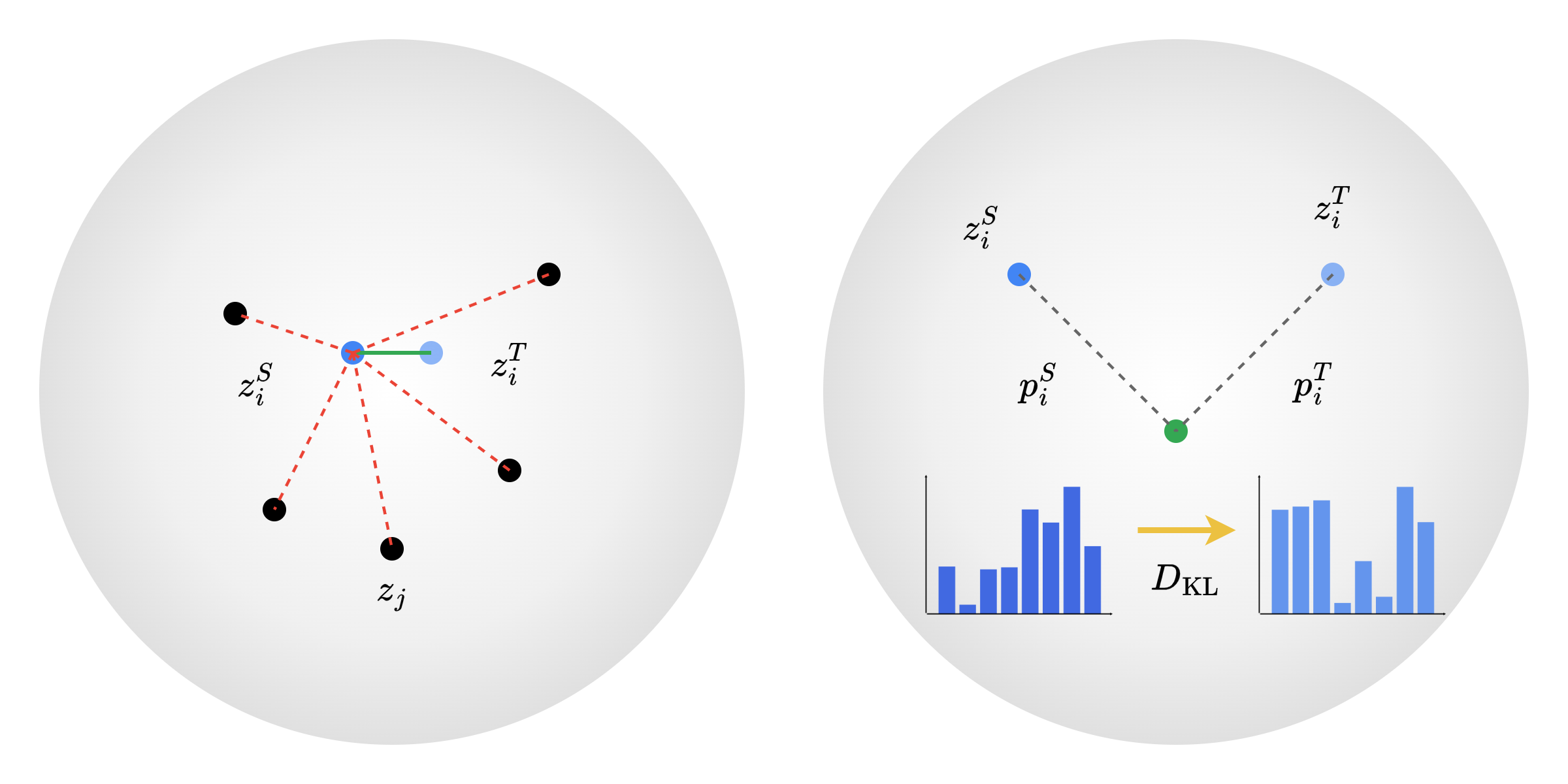}
    \caption{\textbf{Overview of \model.} \textbf{(left)} Discriminative learning through contrastive distillation encourages student features (solid blue) to differentiate between instances by pulling them closer to their corresponding teacher features (transparent blue) while pushing away from other instances as negative samples (black dots). \textbf{(right)} Structural consistency through consistency regularization aligns the two complementary views of the same cross-model similarity matrix (represented by dotted lines) through KL divergence minimization, so that the affinity of student $i$ towards teacher $j$ agrees with the affinity of teacher $i$ towards student $j$.}
    \label{fig:framework}
\end{figure}

To overcome these limitations, we propose \textbf{\underline{D}}iscriminative and \textbf{\underline{C}}onsistent Representation \textbf{\underline{D}}istillation (\model), illustrated in \Cref{fig:framework}, a method that combines contrastive instance discrimination with a consistency regularization term. The contrastive component aligns teacher and student representations at the instance level, while the consistency term, formulated as a KL divergence between the row-normalized and column-normalized views of the cross-model similarity matrix, constrains the off-diagonal structure that the contrastive term leaves free, and vanishes exactly when that matrix is symmetric. Furthermore, our method replaces external memory banks with an efficient in-batch sampling strategy, using only the negative samples that naturally co-exist within each mini-batch, reducing the negative storage from roughly 655MB on ImageNet (CRD) to 0.13MB per iteration while matching the training speed of standard KD at 8ms per batch. We also introduce a learnable scale and bias that adapt during training, automatically adjusting the sharpness and offset of the distillation signal rather than relying on fixed hyperparameters.

Our main \textbf{contributions} are as follows: \textbf{(i)} We propose a unified distillation framework (\Cref{sec:methodology}) that combines contrastive instance discrimination (\Cref{eq:contrastive}) with a KL divergence-based consistency regularization over the cross-model similarity matrix (\Cref{eq:consistency}), enabling the student to learn representations that are both discriminative and bidirectionally consistent with the teacher, and we characterize exactly when the consistency term is minimized (\Cref{eq:symmetry}). \textbf{(ii)} We introduce an efficient in-batch sampling strategy that eliminates the need for large external memory banks, introducing only 66K additional parameters, all discarded at inference, and matching the training throughput of simple logit-based methods (\Cref{subsec:implementation_details} and \Cref{tab:efficiency}). \textbf{(iii)} We employ a learnable scale and bias that adapt during training, providing flexible control over the distillation objective without manual tuning (\Cref{subsec:implementation_details} and \Cref{fig:ablation_study}). \textbf{(iv)} We validate our method through extensive experiments on CIFAR-100 (\Cref{tab:cifar100}), ImageNet (\Cref{tab:imagenet}), and MS-COCO (\Cref{tab:coco}) across 13 teacher-student pairs, demonstrating competitive results in classification, object detection, and cross-dataset transferability (\Cref{tab:transfer}).

\section{Related Work}
\label{sec:related_work}

\subsection{Knowledge Distillation}
\label{subsec:rw_kd}

The original formulation of knowledge distillation~\cite{hinton2015distilling} transfers knowledge through softened logit outputs using temperature scaling in the softmax function. Subsequent work has developed along two main branches: \textit{logit-based} distillation, which operates on the output distributions of the network, and \textit{feature-based} distillation, which leverages intermediate representations.

\paragraph{\bf Logit-based methods.}
Logit-based methods transfer knowledge by aligning the output distributions of the teacher and student networks. Several directions have been explored to improve upon the original KL divergence formulation. One line of work focuses on restructuring the distillation objective itself, including decoupling the KL divergence into target-class and non-target-class components~\cite{zhao2022dkd}, reweighting soft labels from a bias-variance perspective~\cite{zhou2021wsld}, addressing the transfer gap through probability reweighting~\cite{niu2022ipwd}, and normalizing logits before applying softmax and KL divergence to improve gradient behavior~\cite{sun2024logitstandardization}. Another direction explores adaptive temperature control, either by learning the temperature adversarially under a curriculum that gradually increases the difficulty of the distillation objective~\cite{li2022ctkd}, or by relaxing exact distribution matching to correlation-based agreement so as to accommodate teacher-student capacity gaps~\cite{huang2022dist}. Further efforts include applying a power transform to the teacher distribution in place of temperature scaling, which acts as an implicit regularizer on the student~\cite{zheng2024ttm}, and softmax regression-based representation learning~\cite{yang2021srrl}. Our method shares the use of a learned scaling factor with CTKD~\cite{li2022ctkd}, but differs in both the space and the optimization: CTKD learns a temperature over class logits and trains it adversarially through a gradient reversal layer, whereas we learn a scale and an additive bias over feature-space similarities, optimized cooperatively with the student and shared between our two loss terms, following the temperature parameterization of Radford \etal~\cite{radford2021clip}.

\paragraph{\bf Feature-based methods.}
Feature-based methods transfer knowledge from intermediate representations rather than output logits, allowing the student to learn richer structural information from the teacher's internal layers. Early work introduced intermediate feature hints~\cite{romero2014fitnets} and spatial attention alignment~\cite{zagoruyko2016paying} to guide student training. Subsequent work focused on preserving structural relationships, including geometric relations between sample pairs~\cite{park2019relational}, correlation matrices across feature dimensions~\cite{peng2019correlation}, pairwise similarity structures~\cite{tung2019similarity}, and probabilistic embeddings~\cite{passalis2018probabilistic}, alongside factor transfer~\cite{kim2018paraphrasing}, activation boundary transfer~\cite{heo2019knowledge}, flow of solution procedures~\cite{yim2017gift}, neuron selectivity transfer~\cite{huang2017like}, variational information distillation~\cite{ahn2019variational}, and comprehensive feature overhaul~\cite{heo2019ofd}. Recent methods introduced cross-stage review mechanisms~\cite{chen2021reviewkd}, classifier head reuse~\cite{chen2022simkd}, cross-layer semantic calibration~\cite{chen2021semckd}, many-to-one representation matching~\cite{liu2023norm}, feature correlation-based distillation~\cite{liu2023fcfd}, and class attention transfer~\cite{guo2023cat}. Of particular relevance, CRD~\cite{tian2022crd} adapted contrastive objectives to maximize teacher-student mutual information, showing feature-matching alone cannot capture the teacher's full structural knowledge, though it requires a memory bank spanning the training set and fixed hyperparameters. Extensions include Wasserstein-based~\cite{chen2021wcord}, complementary relational~\cite{zhu2021crcd}, kernel-based~\cite{he2022fkd}, and information-theoretic~\cite{miles2022itrd} contrastive distillation. Closest to ours, MCL~\cite{yang2023online} applies contrastive alignment between jointly trained peer networks, whereas we target the offline setting with a frozen teacher and add a consistency term with a learnable scale and bias.

Our method falls within the feature-based family but addresses the key limitations of existing contrastive approaches: we replace external memory banks with in-batch sampling, introduce learnable scale and bias parameters, and augment instance-level contrastive alignment with a consistency term that constrains the off-diagonal structure of the cross-model similarity matrix.

\subsection{Instance Discrimination}
\label{subsec:rw_id}

Instance discrimination methods in self-supervised learning learn representations by treating each individual sample as its own class~\cite{wu2018instdis, ye2019uel, giakoumoglou2024review}, building on foundations in metric learning~\cite{chopra2005learning, hadsell2006dimensionality} and noise contrastive estimation (NCE)~\cite{gutmann2010noise}. These methods transform unsupervised learning into a classification problem, and theoretical analysis has shown that such objectives maximize a lower bound on mutual information between views~\cite{oord2019cpc, arora2019theoretical, hjelm2019dim}. Recent advances include momentum-based encoders~\cite{he2020moco}, with synthetic hard negative generation strategies to strengthen the negative pool~\cite{kalantidis2020mochi, giakoumoglou2024synco}, large-batch contrastive learning with stronger augmentations~\cite{chen2020simclr}, clustering objectives~\cite{caron2020swav,giakoumoglou2025cluster}, and methods that eliminate explicit negative samples entirely through asymmetric architectures~\cite{grill2020byol, chen2020simsiam, caron2021dino} or relational structure-preserving objectives~\cite{giakoumoglou2024rrd,zheng2021ressl}, redundancy reduction~\cite{zbontar2021barlowtwins}, or variance-covariance regularization~\cite{bardes2022vicreg}. Some approaches further explore invariance regularizers~\cite{mitrovic2020relic} or rethink the role of the momentum encoder itself~\cite{michel2024rethinkingmomentum}. Our method combines instance-level discrimination~\cite{wu2018instdis, ye2019uel} with a consistency regularization term that penalizes asymmetry in the cross-model similarity matrix, constraining its off-diagonal entries while retaining the discriminative benefits of instance-level learning. Our approach avoids fixed negative samples or momentum encoders~\cite{michel2024rethinkingmomentum}, using dynamic adaptation during training.

\section{Methodology}
\label{sec:methodology}

Here, we introduce our objective which transfers knowledge from a pre-trained teacher network to a student network by combining instance-level contrastive alignment with a consistency regularization term. \Cref{subsec:preliminaries} outlines the fundamental principles of knowledge distillation, \Cref{subsec:dcd} details the formulation of our dual objective, and \Cref{subsec:implementation_details} describes the implementation details.

\subsection{Preliminaries on Knowledge Distillation}
\label{subsec:preliminaries}

Knowledge distillation involves transferring knowledge from a high-capacity teacher neural network $f^T_\theta$ to a more compact student neural network $f^S_\theta$~\cite{hinton2015distilling, bucilua2006model}. Consider $\mathbf{x}_i$ as the input to these networks, typically an image. We represent the outputs at the penultimate layer, just before the final classification layer, as $\mathbf{z}_i^T = f^T_\theta(\mathbf{x}_i)$ and $\mathbf{z}_i^S = f^S_\theta(\mathbf{x}_i)$ for the teacher and student models, respectively. The primary objective of knowledge distillation is to enable the student model to approximate the performance of the teacher model, which can be expressed as:

\begin{equation}
    \hat{\theta}_{S} = \argmin_{\theta_{S}} \sum_{i=1}^{N} \left(
    \mathcal{L}_{\text{sup}}(\mathbf{x}_i, \theta_{S}, y_i) +
    \lambda \cdot \mathcal{L}_{\text{distill}}(\mathbf{x}_i, \theta_{S}, \theta_{T}) \right),
    \label{eq:standard_kd}
\end{equation}

\noindent where $y_i$ represents the true label for the input $\mathbf{x}_i$, $\theta_S$ and $\theta_T$ are the parameter sets for the student and teacher networks, and $\lambda$ is a hyperparameter that balances the supervised loss and the distillation loss. The supervised loss $\mathcal{L}_{\text{sup}}$ is the task-specific alignment error between the network prediction and annotation. For image classification~\cite{mishra2017apprentice, shen2020amalgamating, polino2018model, cho2019efficacy}, this is typically cross-entropy loss, while for object detection~\cite{liu2019learning, chen2017learning}, it additionally includes bounding box regression. The distillation loss $\mathcal{L}_{\text{distill}}$ is the mimic error of the student network towards the teacher network, typically implemented as KL divergence between student and teacher outputs~\cite{hinton2015distilling}.

\subsection{Discriminative and Consistent Distillation}
\label{subsec:dcd}

Our method extends \Cref{eq:standard_kd} with an additional feature-based distillation term $\mathcal{L}_{\text{dcd}}$:

\begin{equation}
    \hat{\theta}_{S} = \argmin_{\theta_{S}} \sum_{i=1}^{N} \left(
    \mathcal{L}_{\text{sup}}(\mathbf{x}_i, \theta_{S}, y_i) +
    \lambda \cdot \mathcal{L}_{\text{distill}}(\mathbf{x}_i, \theta_{S}, \theta_{T}) +
    \beta \cdot \mathcal{L}_{\text{dcd}}(\mathbf{x}_i, \theta_{S}, \theta_{T}) \right).
    \label{eq:full_loss}
\end{equation}

\noindent \Cref{eq:full_loss} trains the student to jointly minimize the supervised task loss, the logit-space KL distillation loss weighted by $\lambda$, and our proposed feature-space loss weighted by $\beta$, with the teacher backbone parameters $\theta_T$ kept frozen. Our loss $\mathcal{L}_{\text{dcd}}$ consists of two complementary terms:

\begin{equation}
    \mathcal{L}_{\text{dcd}} = \mathcal{L}_{\text{ctr}} + \alpha \cdot \mathcal{L}_{\text{cst}},
    \label{eq:dcd_loss}
\end{equation}

\noindent where $\mathcal{L}_{\text{ctr}}$ is a contrastive objective (\Cref{eq:contrastive}) for instance-level discrimination, $\mathcal{L}_{\text{cst}}$ is a consistency regularization term (\Cref{eq:consistency}) that constrains the off-diagonal structure of the cross-model similarity matrix, and $\alpha$ balances the two components (\cf \Cref{subsec:ablations}). The combination provides complementary supervision: KL divergence over class logits offers direct class-level guidance, while our proposed loss operates entirely in feature space.

\paragraph{\bf Similarity matrix.}
All quantities below are derived from a single matrix. For a mini-batch of $N$ samples we compute the $\ell_2$-normalized embeddings $\mathbf{z}_i^S$ and $\mathbf{z}_i^T$ (\cf \Cref{subsec:implementation_details}) and form

\begin{equation}
    \ell_{ij} = s \cdot \langle \mathbf{z}_i^S, \mathbf{z}_j^T \rangle + b, \qquad s = \exp(\tau),
    \label{eq:logits}
\end{equation}

\noindent where $\langle \cdot, \cdot \rangle$ denotes the inner product, rows of $\ell \in \mathbb{R}^{N \times N}$ are indexed by students and columns by teachers, and $\tau$ and $b$ are the learnable scale and bias parameters described in \Cref{subsec:implementation_details}.

\paragraph{\bf Contrastive objective.}
We employ contrastive learning to align teacher and student representations at the instance level. Each student representation $\mathbf{z}_i^S$ must identify its corresponding teacher representation $\mathbf{z}_i^T$ among all $N$ teacher representations in the batch, which we formulate as an $N$-way classification problem using noise contrastive estimation~\cite{gutmann2010noise}:

\begin{equation}
    \mathcal{L}_{\text{ctr}} = -\frac{1}{N} \sum_{i=1}^{N} \log \frac{\exp(\ell_{ii})}{\sum_{j=1}^{N} \exp(\ell_{ij})}.
    \label{eq:contrastive}
\end{equation}

\noindent The negative samples are all other teacher representations within the same mini-batch, eliminating the need for external memory banks. In practice this reduces to a cross-entropy loss in which the scaled similarities act as logits and the positive pair indices act as class labels, so the objective is a softmax classification over the rows of $\ell$ with the diagonal as the target. This is the InfoNCE objective~\cite{oord2019cpc} applied across models rather than across augmented views, with the scale and bias shared with the consistency term below.

\paragraph{\bf Consistency objective.}
The contrastive objective supervises only the diagonal of $\ell$ through its row-wise normalization, leaving the off-diagonal entries free up to the row constraints. To constrain them, we introduce a consistency regularization term that enforces agreement between two complementary normalizations of the same matrix. The student view normalizes each row over the teacher axis, giving, for each student $i$, its relative affinity to all teacher representations; the teacher view normalizes each column over the student axis, giving, for each teacher $i$, the relative affinity of all student representations towards it:

\begin{equation}
    p_i^S(j) = \frac{\exp(\ell_{ij})}{\sum_{k=1}^{N} \exp(\ell_{ik})}, \qquad
    p_i^T(j) = \frac{\exp(\ell_{ji})}{\sum_{k=1}^{N} \exp(\ell_{ki})}.
    \label{eq:distributions}
\end{equation}

\noindent Both $\mathbf{p}_i^S = [p_i^S(1), \ldots, p_i^S(N)]$ and $\mathbf{p}_i^T = [p_i^T(1), \ldots, p_i^T(N)]$ sum to one over $j$ and are therefore probability vectors of length $N$, so the KL divergence between them is well-defined. Intuitively, $\mathbf{p}_i^S$ answers ``which teacher representation best matches student $i$?'', while $\mathbf{p}_i^T$ answers ``which student representation best matches teacher $i$?''. The consistency objective enforces agreement between the two by minimizing:

\begin{equation}
    \mathcal{L}_{\text{cst}} = \frac{1}{N} \sum_{i=1}^{N} D_{\text{KL}}\!\left(\mathbf{p}_i^T \,\big\|\, \mathbf{p}_i^S\right) = \frac{1}{N} \sum_{i=1}^{N} \sum_{j=1}^{N} p_i^T(j) \log \frac{p_i^T(j)}{p_i^S(j)}.
    \label{eq:consistency}
\end{equation}

\noindent The term is non-negative and vanishes if and only if $\mathbf{p}_i^S = \mathbf{p}_i^T$ for every $i$, and this condition admits an exact characterization. Taking logarithms of $p_i^S(j) = p_i^T(j)$ gives $\ell_{ij} - \ell_{ji} = c_i$ with $c_i = \log \sum_k \exp(\ell_{ik}) - \log \sum_k \exp(\ell_{ki})$ depending on $i$ alone; exchanging $i$ and $j$ gives $\ell_{ji} - \ell_{ij} = c_j$, so $c_j = -c_i$ for all $i, j$, which forces every $c_i$ to equal a single constant $c$ satisfying $c = -c$, hence $c = 0$. Since $s > 0$ and $b$ is shared, this yields

\begin{equation}
    \mathcal{L}_{\text{cst}} = 0 \iff \langle \mathbf{z}_i^S, \mathbf{z}_j^T \rangle = \langle \mathbf{z}_j^S, \mathbf{z}_i^T \rangle \quad \forall\, i, j,
    \label{eq:symmetry}
\end{equation}

\noindent that is, the term is minimized precisely when the cross-model similarity matrix is symmetric. It therefore asks that the affinity of student $i$ towards teacher $j$ agree with the affinity of student $j$ towards teacher $i$, a condition that is satisfied when the student reproduces the teacher's embeddings and that is otherwise violated in a way the diagonal-only contrastive term cannot detect. Note that both views are functions of the student, so neither acts as a fixed target and gradients propagate through both; we do not apply a stop-gradient to $\mathbf{p}_i^T$. Unlike relational methods such as RKD~\cite{park2019relational} that preserve geometric relations between sample pairs, or CC~\cite{peng2019correlation} that matches teacher-only against student-only correlation statistics, our term operates on the cross-model matrix directly and requires no separate intra-teacher or intra-student similarity computation.



\subsection{Implementation Details}
\label{subsec:implementation_details}

We implement the objective using mini-batch stochastic gradient descent. The representations $\mathbf{z}_i^T$ and $\mathbf{z}_i^S$ are taken from the penultimate layer of the teacher and student models, respectively, and are mapped into a shared 128-dimensional embedding space by a single linear projection layer per network. Both projections are trained jointly with the student by stochastic gradient descent, including the teacher-side projection, even though the teacher backbone itself remains frozen, and both are discarded after training so that they add no inference cost. We $\ell_2$-normalize the projected outputs before computing \Cref{eq:logits}, so that the representations lie on a unit hypersphere and the inner product reduces to a cosine similarity.

\paragraph{\bf Memory-efficient in-batch sampling.}
Instead of maintaining a large memory buffer for negative sampling as in CRD~\cite{tian2022crd}, we use the negative samples that naturally co-exist within the mini-batch. This substantially reduces memory requirements. CRD maintains a memory bank holding one 128-dimensional feature per training image for each network, which on ImageNet amounts to roughly 655MB per bank ($1.28$M features $\times$ $128$ dimensions $\times$ $4$ bytes), from which $16$k negatives are sampled per anchor; our in-batch sampling instead materializes only the current batch, requiring 0.13MB per iteration ($256 \times 128 \times 4$ bytes at the ImageNet batch size of 256, and a quarter of that at the CIFAR-100 batch size of 64). This also eliminates the complexity of memory bank management, including challenges related to feature staleness and queue maintenance~\cite{he2020moco}, and ensures that all negative representations are up-to-date within the current training iteration. This efficiency extends to training time: on a 4-GPU machine, our method completes ImageNet training in approximately 72 hours compared to CRD's 88 hours, representing an 18\% reduction in training time.

\paragraph{\bf Learnable scale and bias.}
Unlike contrastive objectives that use a fixed temperature, we parameterize the multiplicative scale in \Cref{eq:logits} as $s = \exp(\tau)$ with $\tau$ a learnable scalar, following the temperature parameterization of Radford \etal~\cite{radford2021clip}, and add a learnable bias $b$. The exponential parameterization is used rather than a raw temperature divisor for two reasons. First, it guarantees that the effective scale remains strictly positive for every $\tau \in \mathbb{R}$, so no lower bound or constrained optimization is required. Second, it permits unconstrained gradient-based optimization of $\tau$ while producing smooth, well-behaved gradients, since $\exp(\tau)$ is its own derivative and never vanishes. Larger $\tau$ sharpens both distributions in \Cref{eq:distributions} and smaller $\tau$ softens them. We initialize $\tau = 1$, giving an initial scale of $s = e \approx 2.72$, and $b = 0$. The scale is bounded from above at $s_{\max} = 10$ for numerical stability, which is equivalent to constraining $\tau \le \log s_{\max}$; we apply this bound to the parameter after each optimizer step rather than inside the forward pass, so that the gradient of $\tau$ is not zeroed once the bound is reached. The learnable bias $b$ provides an additive degree of freedom that shifts the logit scale globally, allowing the model to compensate for any systematic offset between teacher and student similarity magnitudes. Both $\tau$ and $b$ are shared across the contrastive and consistency objectives (\Cref{eq:contrastive,eq:consistency}), coupling the two terms through a common similarity parameterization. This adaptive approach allows the model to tune the contrast level and logit offset during training, rather than requiring these to be selected per teacher-student pair. We ablate the learnable $\tau$ and $b$ in \Cref{subsec:ablations}.

\section{Experiments}
\label{sec:experiments}

We evaluate our method\footnote{Following standard practice, we denote our method as ``\textit{\model}'' when using only supervised and proposed losses (\ie, $\lambda=0$), and ``\textit{\model+KD}'' when incorporating all objectives (\ie, $\lambda \neq 0$).} on both image classification and object detection tasks across multiple benchmarks and teacher-student configurations. \Cref{subsec:setup} outlines the experimental setup, \Cref{subsec:results} presents quantitative results across benchmarks, and \Cref{subsec:visualizations} analyzes the learned representations. Ablations are discussed in \Cref{subsec:ablations}.

\subsection{Experimental Setup}
\label{subsec:setup}


\paragraph{\bf Datasets.}
We evaluate the proposed framework on both image classification and object detection tasks using five standard benchmarks: CIFAR-100~\cite{krizhevsky2009cifar} and ImageNet ILSVRC-2012~\cite{deng2009imagenet} for classification, STL-10~\cite{coates2011importance} and Tiny ImageNet~\cite{deng2009imagenet} for cross-dataset transfer, and MS-COCO~\cite{li2015coco} for object detection.

\paragraph{\bf Architectures.}
Following prior work~\cite{tian2022crd}, we experiment with 13 teacher-student combinations of varying capacity using CIFAR-style ResNet (rn) and ImageNet-style ResNet (RN)~\cite{he2015resnet}, Wide ResNet (WRN)~\cite{zagoruyko2017wide}, VGG~\cite{simonyan2015vgg}, MobileNet-v2 (MN-v2)~\cite{sandler2018mobilenetv2} with a width multiplier of 0.5, and ShuffleNet-v1 (SN-v1)~\cite{zhang2018shufflenet} and ShuffleNet-v2 (SN-v2)~\cite{ma2018shufflenet}. We follow the implementation protocol of~\cite{tian2022crd} for image classification and~\cite{zhao2022dkd, chen2021reviewkd} for object detection.

\paragraph{\bf Hyperparameters.}
We set $\alpha=0.5$, $\beta=1$, $s_{\max}=10.0$, and initialize $\tau=1.0$ and $b=0.0$. The hyperparameter $\lambda$ is set to $1.0$ for the KL divergence loss to maintain consistency with~\cite{tian2022crd, chen2021semckd, chen2022simkd}, while we provide ablations in \Cref{subsec:ablations}. All CIFAR-100 results are reported as means over five trials; ImageNet results are from a single trial.

\paragraph{\bf Baselines.}
We compare against a wide range of distillation methods spanning logit-based, feature-based, and contrastive approaches, including KD~\cite{hinton2015distilling}, FitNet~\cite{romero2014fitnets}, AT~\cite{zagoruyko2016paying}, SP~\cite{tung2019similarity}, CC~\cite{peng2019correlation}, VID~\cite{ahn2019variational}, RKD~\cite{park2019relational}, PKT~\cite{passalis2018probabilistic}, AB~\cite{heo2019knowledge}, FT~\cite{kim2018paraphrasing}, FSP~\cite{yim2017gift}, CRD~\cite{tian2022crd}, OFD~\cite{heo2019ofd}, WSLD~\cite{zhou2021wsld}, IPWD~\cite{niu2022ipwd}, and CTKD~\cite{li2022ctkd}.

\subsection{Main Results}
\label{subsec:results}

We benchmark our method on image classification and object detection tasks.

\begin{table*}[!t]
\centering
\setlength{\tabcolsep}{0.3mm}
\caption{\textbf{Main results on CIFAR-100.} Test top-1 accuracy (\%) for various teacher-student architecture combinations. Results for our method and those adapted from \cite{tian2022crd} are averaged over five independent runs.}
\label{tab:cifar100}
\resizebox{\textwidth}{!}{
\begin{tabular}{lccccccccccccc}
\toprule
& \multicolumn{7}{c}{Same architecture} & \multicolumn{6}{c}{Different architecture} \\
\cmidrule(lr){2-8} \cmidrule(lr){9-14}
Teacher & WRN-40-2 & WRN-40-2 & rn-56 & rn-110 & rn-110 & rn-32x4 & VGG-13 & VGG-13 & RN-50 & RN-50 & RN-32x4 & RN-32x4 & WRN-40-2 \\
Student & WRN-16-2 & WRN-40-1 & rn-20 & rn-20 & rn-32 & rn-8x4 & VGG-8 & MN-v2 & MN-v2 & VGG-8 & SN-v1 & SN-v2 & SN-v1 \\ 
\midrule
\textit{Teacher} & 75.61 & 75.61 & 72.34 & 74.31 & 74.31 & 79.42 & 74.64 & 74.64 & 79.34 & 79.34 & 79.42 & 79.42 & 75.61 \\
\textit{Student} & 73.26 & 71.98 & 69.06 & 69.06 & 71.14 & 72.50 & 70.36 & 64.60 & 64.60 & 70.36 & 70.50 & 71.82 & 70.50 \\ 
KD \cite{hinton2015distilling} & 74.92 & 73.54 & 70.66 & 70.67 & 73.08 & 73.33 & 72.98 & 67.37 & 67.35 & 73.81 & 74.07 & 74.45 & 74.83 \\
FitNet \cite{romero2014fitnets} & 73.58 & 72.24 & 69.21 & 68.99 & 71.06 & 73.50 & 71.02 & 64.14 & 63.16 & 70.69 & 73.59 & 73.54 & 73.73  \\
AT \cite{zagoruyko2016paying} & 74.08 & 72.77 & 70.55 & 70.22 & 72.31 & 73.44 & 71.43 & 59.40 & 58.58 & 71.84 & 71.73 & 72.73 & 73.32 \\
SP \cite{tung2019similarity} & 73.83 & 72.43 & 69.67 & 70.04 & 72.69 & 72.94 & 72.68 & 66.30 & 68.08 & 73.34 & 73.48 & 74.56 & 74.52  \\
CC \cite{peng2019correlation} & 73.56 & 72.21 & 69.63 & 69.48 & 71.48 & 72.97 & 70.81 & 64.86 & 65.43 & 70.25 & 71.14 & 71.29 & 71.38 \\
VID \cite{ahn2019variational} & 74.11 & 73.30 & 70.38 & 70.16 & 72.61 & 73.09 & 71.23 & 65.56 & 67.57 & 70.30 & 73.38 & 73.40 & 73.61  \\
RKD \cite{park2019relational} & 73.35 & 72.22 & 69.61 & 69.25 & 71.82 & 71.90 & 71.48 & 64.52 & 64.43 & 71.50 & 72.28 & 73.21 & 72.21  \\
PKT \cite{passalis2018probabilistic} & 74.54 & 73.45 & 70.34 & 70.25 & 72.61 & 73.64 & 72.88 & 67.13 & 66.52 & 73.01 & 74.10 & 74.69 & 73.89  \\
AB \cite{heo2019knowledge} & 72.50 & 72.38 & 69.47 & 69.53 & 70.98 & 73.17 & 70.94 & 66.06 & 67.20 & 70.65 & 73.55 & 74.31 & 73.34 \\
FT \cite{kim2018paraphrasing} & 73.25 & 71.59 & 69.84 & 70.22 & 72.37 & 72.86 & 70.58 & 61.78 & 60.99 & 70.29 & 71.75 & 72.50 & 72.03 \\
FSP \cite{yim2017gift} & 72.91 & n/a & 69.95 & 70.11 & 71.89 & 72.62 & 70.33 & 58.16 & 64.96 & 71.28 & 74.12 & 74.68 & 76.09 \\
CRD \cite{tian2022crd} & 75.48 & 74.14 & 71.16 & 71.46 & 73.48 & \underline{75.51} & 73.94 & 69.73 & 69.11 & 74.30 & 75.11 & 75.65 & 76.05 \\
CRD+KD \cite{tian2022crd} & \underline{75.64} & 74.38 & \underline{71.63} & \underline{71.56} & \underline{73.75} & 75.46 & \textbf{74.29} & \textbf{69.94} & 69.54 & \underline{74.58} & 75.12 & 76.05 & 76.27 \\ 
OFD \cite{heo2019ofd} & 75.24 & 74.33 & 70.38 & n/a & 73.23 & 74.95 & \underline{73.95} & 69.48 & 69.04 & n/a & 75.98 & \underline{76.82} & 75.8 \\
WSLD \cite{zhou2021wsld} & n/a & 73.74 & 71.53 & n/a & 73.36 & 74.79 & n/a & n/a & 68.79 & 73.80 & 75.09 & n/a & 75.23 \\
IPWD \cite{niu2022ipwd} & n/a & \underline{74.64} & 71.32 & n/a & \textbf{73.91} & \textbf{76.03} & n/a & n/a & \textbf{70.25} & \textbf{74.97} & \textbf{76.03} & n/a & \underline{76.44} \\
CTKD \cite{li2022ctkd} & 75.45 & 73.93 & 71.19 & 70.99 & 73.52 & n/a & 73.52 & 68.46 & 68.47 & n/a & 74.78 & 75.31 & 75.78 \\
\model (ours) & \cellcolor{gray!12}74.99 & \cellcolor{gray!12}73.69 & \cellcolor{gray!12}71.18 & \cellcolor{gray!12}71.00 & \cellcolor{gray!12}73.12 & \cellcolor{gray!12}74.23 & \cellcolor{gray!12}73.22 & \cellcolor{gray!12}68.35 & \cellcolor{gray!12}67.39 & \cellcolor{gray!12}73.85 & \cellcolor{gray!12}74.26 & \cellcolor{gray!12}75.26 & \cellcolor{gray!12}74.98 \\
\model+KD (ours) & \cellcolor{gray!12}\textbf{76.06} & \cellcolor{gray!12}\textbf{74.76} & \cellcolor{gray!12}\textbf{71.81} & \cellcolor{gray!12}\textbf{72.03} & \cellcolor{gray!12}73.62 & \cellcolor{gray!12}75.09 & \cellcolor{gray!12}\underline{73.95} & \cellcolor{gray!12}\underline{69.77} & \cellcolor{gray!12}\underline{70.03} & \cellcolor{gray!12}74.08 & \cellcolor{gray!12}\underline{76.01} & \cellcolor{gray!12}\textbf{76.95} & \cellcolor{gray!12}\textbf{76.51} \\
\bottomrule
\end{tabular}
}
\end{table*}
\begin{table*}[!t]
\centering
\setlength{\tabcolsep}{0.8mm}
\caption{\textbf{Main results on ImageNet.} Test top-1 accuracy (\%) on the ILSVRC-2012 validation set across diverse distillation methods. We report results from a single run across representative teacher-student pairs.}
\begin{tabular}{lccc}
\toprule
Teacher & RN-34 & RN-50 & RN-50 \\
Student & RN-18 & RN-18 & MN-v2 \\
\midrule
Teacher & 73.31 & 76.16 & 76.16 \\
Student & 69.75 & 69.75 & 69.63 \\
KD \cite{hinton2015distilling} & 70.67 & 71.29 & 70.49 \\
AT \cite{zagoruyko2016paying} & 71.03 & 71.18 & 70.18 \\
SP \cite{tung2019similarity} & 70.62 & 71.08 & n/a \\
CC \cite{peng2019correlation} & 69.96 & n/a & n/a \\
RKD \cite{park2019relational} & 70.40 & n/a & 68.50 \\
CRD \cite{tian2022crd} & 71.17 & 71.25 & 69.07 \\
\model (ours) & \cellcolor{gray!12}71.10 & \cellcolor{gray!12}71.38 & \cellcolor{gray!12}70.51 \\
\model+KD (ours) & \cellcolor{gray!12}\textbf{71.71} & \cellcolor{gray!12}\textbf{71.65} & \cellcolor{gray!12}\textbf{71.55} \\
\bottomrule
\end{tabular}
\label{tab:imagenet}
\end{table*}
\begin{table*}[!t]
\centering
\caption{\textbf{Object detection performance on MS-COCO.} Evaluation using Faster R-CNN with an FPN backbone on the $\texttt{val2017}$ set. We report mean Average Precision (AP) and AP at IoU thresholds of 0.5 and 0.75 for single-run experiments.}
\label{tab:coco}
\resizebox{\textwidth}{!}{
\begin{tabular}{lcccccccccc}
\toprule
\multirow{2}{*}{Method} & \multicolumn{3}{c}{RN-101 $\rightarrow$ RN-18} & \multicolumn{3}{c}{RN-101 $\rightarrow$ RN-50} & \multicolumn{3}{c}{RN-50 $\rightarrow$ MN-v2} \\
\cmidrule(lr){2-4} \cmidrule(lr){5-7} \cmidrule(lr){8-10}
& AP & AP$_{50}$ & AP$_{75}$ & AP & AP$_{50}$ & AP$_{75}$ & AP & AP$_{50}$ & AP$_{75}$ \\
\midrule
\textit{Teacher} & 42.04 & 62.48 & 45.88 & 42.04 & 62.48 & 45.88 & 40.22 & 61.02 & 43.81 \\
\textit{Student} & 33.26 & 53.61 & 35.26 & 37.93 & 58.84 & 41.05 & 29.47 & 48.87 & 30.90 \\
KD \cite{hinton2015distilling} & 33.97 & 54.66 & 36.62 & 38.35 & 59.41 & 41.71 & 30.13 & 50.28 & 31.35 \\
FitNet \cite{romero2014fitnets} & 34.13 & 54.16 & 36.71 & 38.76 & 59.62 & 41.80 & 30.20 & 49.80 & 31.69 \\
ReviewKD \cite{chen2021reviewkd} & 36.75 & 56.72 & 39.00 & 40.36 & 60.97 & 44.08 & 33.71 & 53.15 & 36.13 \\
DKD \cite{zhao2022dkd} & 35.05 & 56.60 & 37.54 & 39.25 & 60.90 & 42.73 & 32.34 & 53.77 & 34.01 \\
\model (ours) & \cellcolor{gray!12}\textbf{37.12} & \cellcolor{gray!12}\textbf{57.58} & \cellcolor{gray!12}\textbf{39.93} & \cellcolor{gray!12}\textbf{40.48} & \cellcolor{gray!12}\textbf{61.14} & \cellcolor{gray!12}\textbf{44.21} & \cellcolor{gray!12}\textbf{33.89} & \cellcolor{gray!12}\textbf{53.84} & \cellcolor{gray!12}\textbf{36.28} \\
\bottomrule
\end{tabular}
}
\end{table*}
\begin{table*}[!t]
\centering
\setlength{\tabcolsep}{0.8mm}
\caption{\textbf{Cross-dataset generalization performance.} Test top-1 accuracy (\%) of a WRN-16-2 student distilled from a WRN-40-2 teacher. Representations learned on CIFAR-100 are transferred and evaluated on the STL-10 and Tiny ImageNet datasets.}
\label{tab:transfer}
\resizebox{\textwidth}{!}{
\begin{tabular}{lccccccccc}
\toprule
& \textit{Teacher} & \textit{Student} & KD \cite{hinton2015distilling}  & AT \cite{zagoruyko2016paying}  & FitNet \cite{romero2014fitnets} & CRD \cite{tian2022crd} & CRD+KD \cite{tian2022crd} & \model & \model+KD\\
\midrule
STL-10 & 68.6 & 69.7  & 70.9 & 70.7 & 70.3 & 71.6 & 72.2 & \cellcolor{gray!12}71.2 & \cellcolor{gray!12}\textbf{72.5}  \\
Tiny ImageNet & 31.5 & 33.7  & 33.9 & 34.2 & 33.5 & 35.6 & 35.5 & \cellcolor{gray!12}35.0 & \cellcolor{gray!12}\textbf{36.2} \\
\bottomrule
\end{tabular}
}
\end{table*}

\begin{table}[!t]
\centering
\setlength{\tabcolsep}{0.7mm}
\caption{\textbf{Efficiency analysis on CIFAR-100.} Comparison of training latency (ms per batch) and parameter counts (in millions, M) for different distillation frameworks. We use a ResNet-32x4 teacher and ResNet-8x4 student on an NVIDIA RTX 6000 GPU.}
\label{tab:efficiency}
\resizebox{\textwidth}{!}{
\begin{tabular}{lccccccccc}
\toprule
 & KD \cite{hinton2015distilling} & FitNet \cite{romero2014fitnets} & AT \cite{zagoruyko2016paying} & RKD \cite{park2019relational} & CRD \cite{tian2022crd} & OFD \cite{heo2019ofd} & RewiewKD \cite{chen2021reviewkd} & DKD \cite{zhao2022dkd} & \model \\
\midrule
Time (ms) & 8 & 8 & 10 & 17 & 19 & 20 & 14 & 9 & \cellcolor{gray!12}8 \\
Params (M) & 0 & 0.017 & 0 & 0 & 12.866 & 0.087 & 1.809 & 0 & \cellcolor{gray!12}0.066 \\
\bottomrule
\end{tabular}
}
\end{table}

\paragraph{\bf Results on CIFAR-100.}
We evaluate top-1 classification accuracy on CIFAR-100 across 13 teacher-student pairs covering both same-architecture and cross-architecture configurations. \Cref{tab:cifar100} compares our method against existing distillation approaches. \model combined with KD achieves strong performance, surpassing the teacher network by +0.45\% in the same-architecture setting (WRN-40-2 to WRN-16-2) and by +0.90\% in the cross-architecture setting (WRN-40-2 to ShuffleNet-v1). Over baseline students, the improvements reach +2.82\% for same-architecture and +5.25\% for cross-architecture pairs. \model alone performs slightly below CRD, which draws its negatives from a memory bank spanning the full training set rather than the current batch. Combining \model with KD recovers and exceeds this gap, consistent with the two objectives supervising different spaces.

\paragraph{\bf Results on ImageNet.}
We evaluate top-1 classification accuracy on ImageNet ILSVRC-2012 across multiple teacher-student configurations. \Cref{tab:imagenet} presents the results. Our method consistently improves upon baselines~\cite{hinton2015distilling, zagoruyko2016paying, tung2019similarity, peng2019correlation, park2019relational, tian2022crd} and achieves competitive performance across different architectures, including challenging cross-architecture transfer scenarios. These results confirm that the gains observed on CIFAR-100 extend to large-scale settings.

\paragraph{\bf Results on COCO.}
We evaluate object detection performance on MS-COCO using Faster R-CNN~\cite{ren2016fasterrcnn} with Feature Pyramid Network (FPN)~\cite{lin2017feature} as the detection framework, following~\cite{zhao2022dkd}. We report AP, AP$_{50}$, and AP$_{75}$ across three teacher-student scenarios: ResNet-101 to ResNet-18, ResNet-101 to ResNet-50, and ResNet-50 to MobileNet-v2. \Cref{tab:coco} presents the results. Our method consistently improves upon baselines~\cite{hinton2015distilling, zhao2022dkd, romero2014fitnets, chen2021reviewkd} across all three settings, demonstrating that the gains of our approach extend beyond classification to the more challenging object detection task, including cross-architecture transfer scenarios.

\paragraph{\bf Transferability of representations.}
We evaluate the cross-dataset transferability of distilled representations by training a WRN-16-2 student (distilled from WRN-40-2 teacher) on CIFAR-100, then using it as a frozen feature extractor with a linear classifier on STL-10 and Tiny ImageNet. \Cref{tab:transfer} reports the top-1 test accuracy for each distillation method. Our method, both standalone and combined with KD, consistently improves transferability over baselines~\cite{hinton2015distilling, zagoruyko2016paying, romero2014fitnets, tian2022crd}, indicating that the learned representations capture generalizable features rather than overfitting to the training distribution.

\paragraph{\bf Efficiency.}
We compare the computational cost of our method against existing approaches in terms of training latency and parameter overhead. \Cref{tab:efficiency} presents the results using a ResNet-32x4 teacher and ResNet-8x4 student on CIFAR-100. Our method matches KD and FitNet as the fastest approach at 8ms per batch, while requiring only 65,794 additional trainable parameters for this pair: 65,792 for the two linear projection layers ($256 \times 128 + 128$ each) and the two scalars $\tau$ and $b$. The count scales with the teacher and student feature dimensions, and all of these parameters are discarded at inference. By contrast, CRD stores a memory bank proportional to the size of the training set and runs at 19ms per batch, making our method over 2$\times$ faster. Compared to other feature-based methods such as ReviewKD~\cite{chen2021reviewkd} (1.81M parameters, 14ms) and OFD~\cite{heo2019ofd} (0.087M parameters, 20ms), our approach is both lighter and faster. Our method therefore achieves the efficiency of simple logit-based methods like KD and DKD while retaining the accuracy gains of contrastive feature-based distillation.

\subsection{Visualizations}
\label{subsec:visualizations}

\begin{figure}[!t]
\centering
\begin{subfigure}[b]{0.17\textwidth}
  \centering
  \includegraphics[width=\linewidth]{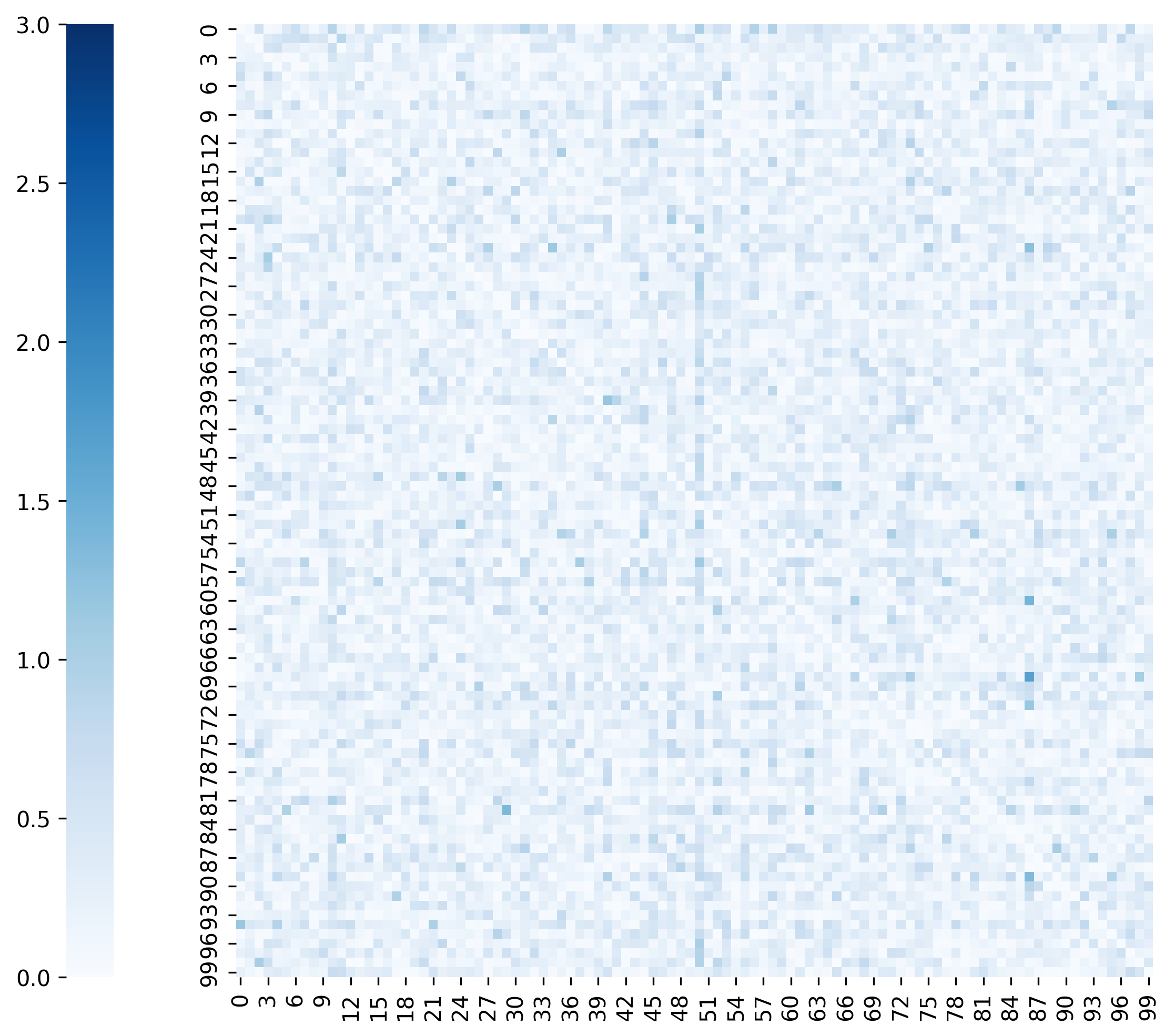}
  \subcaption{Vanilla \\ Mean: 0.24\\ Max: 1.66}
\end{subfigure}
\hfill
\begin{subfigure}[b]{0.17\textwidth}
  \centering
  \includegraphics[width=\linewidth]{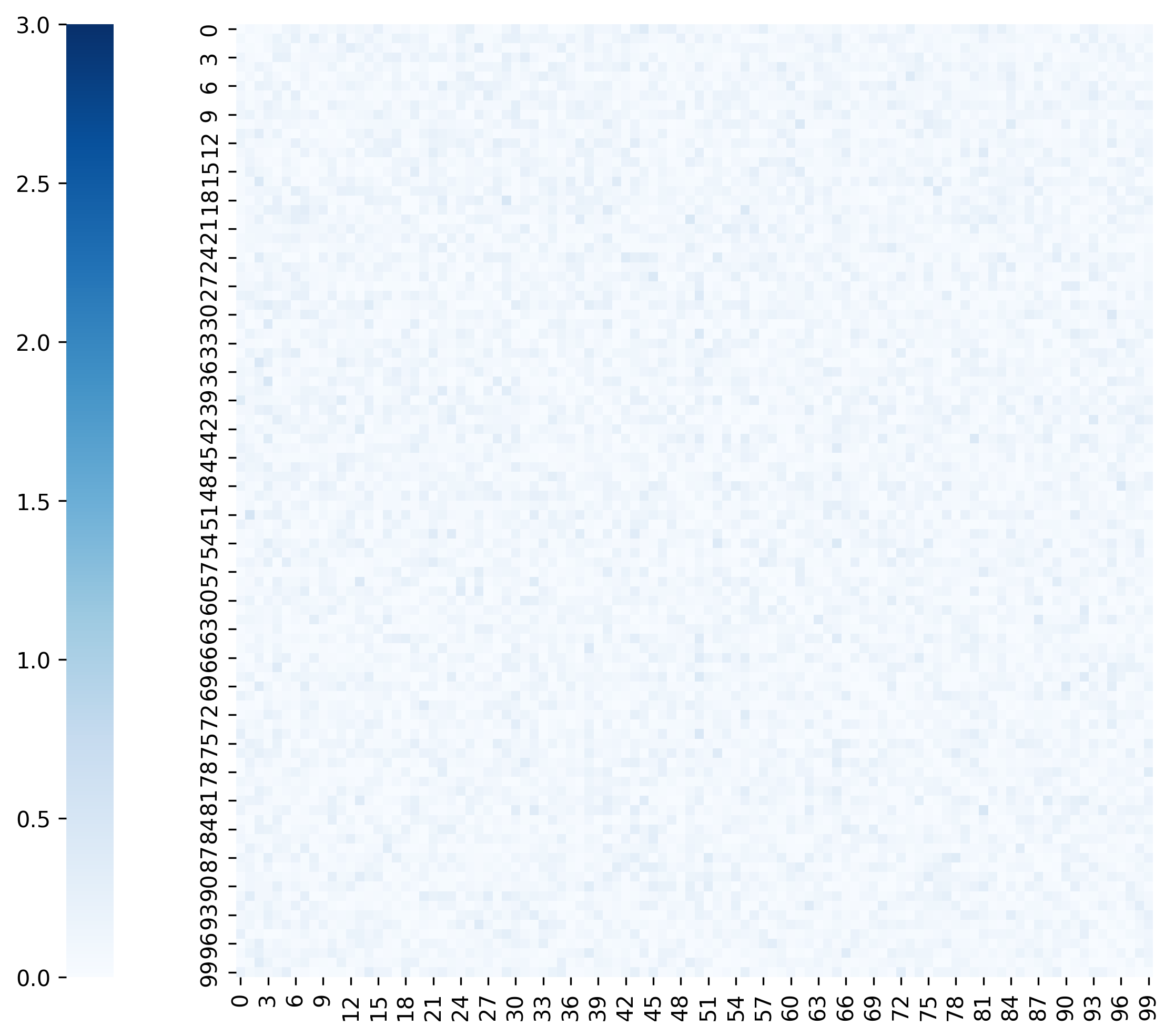}
  \subcaption{KD \cite{hinton2015distilling} \\ Mean: 0.09\\ Max: 0.49}
\end{subfigure}
\hfill
\begin{subfigure}[b]{0.17\textwidth}
  \centering
  \includegraphics[width=\linewidth]{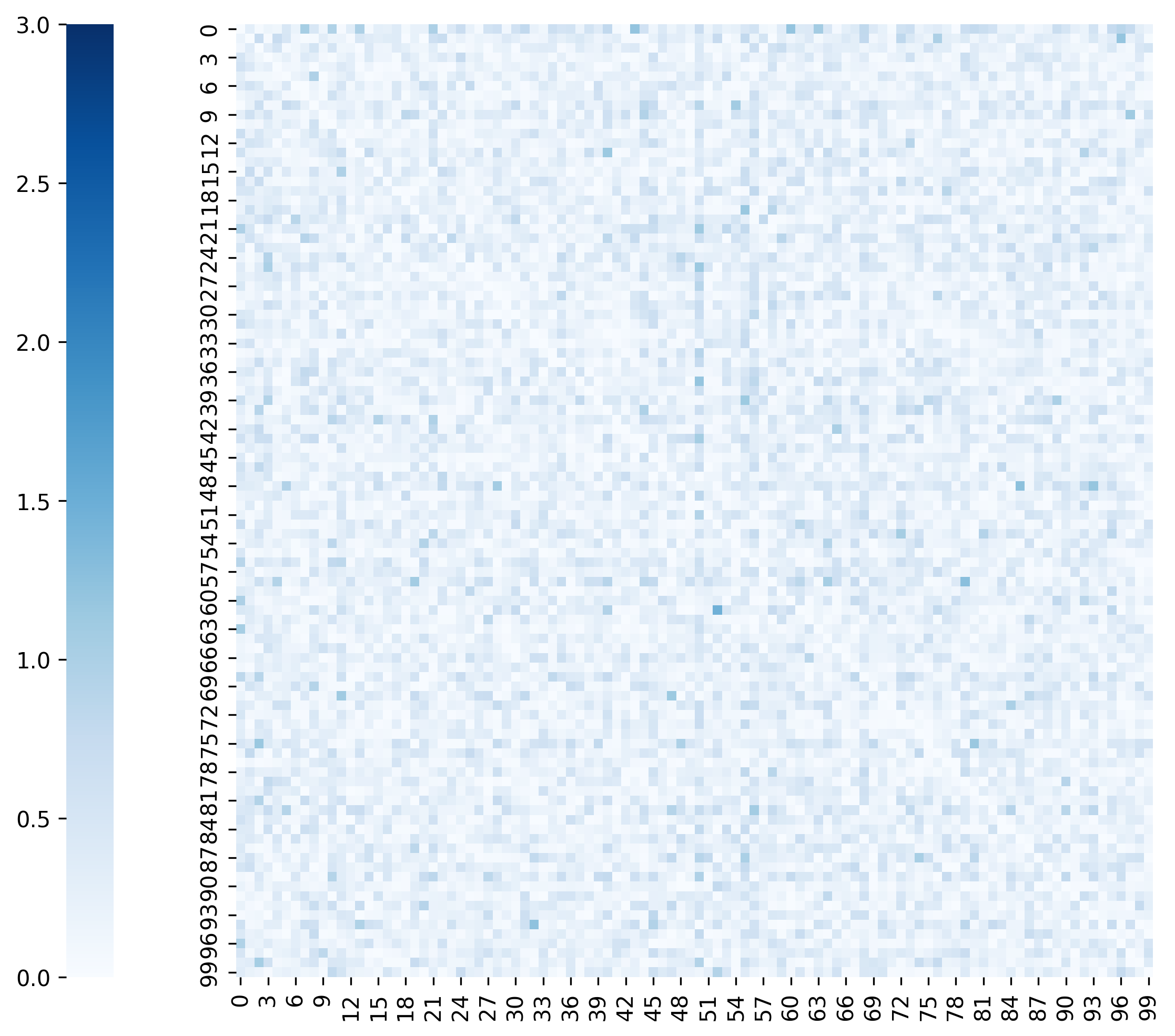}
  \subcaption{CRD \cite{tian2022crd} \\ Mean: 0.23\\ Max: 1.56}
\end{subfigure}
\hfill
\begin{subfigure}[b]{0.17\textwidth}
  \centering
  \includegraphics[width=\linewidth]{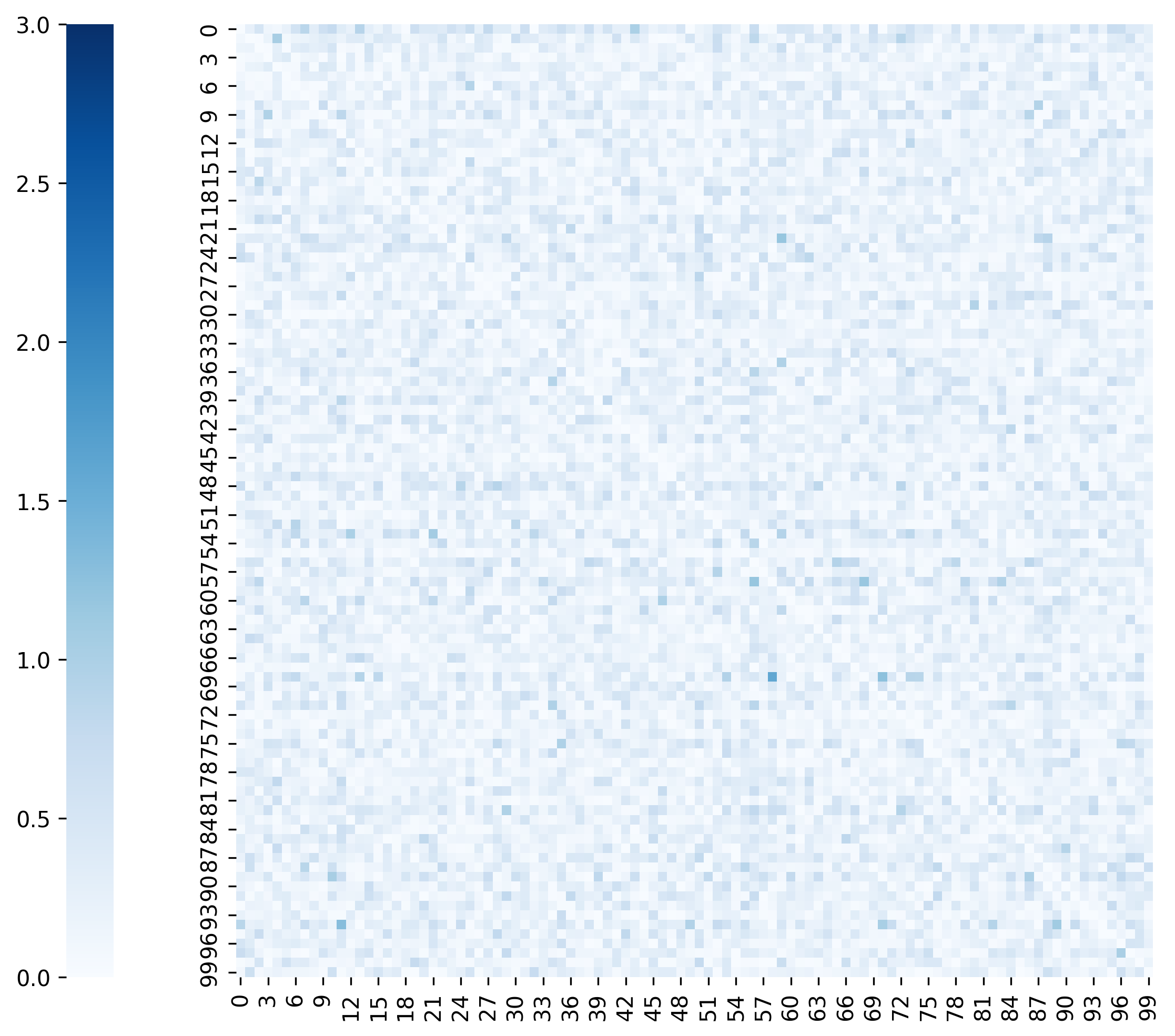}
  \subcaption{\model \\ Mean: 0.22\\ Max: 1.59}
\end{subfigure}
\hfill
\begin{subfigure}[b]{0.17\textwidth}
  \centering
  \includegraphics[width=\linewidth]{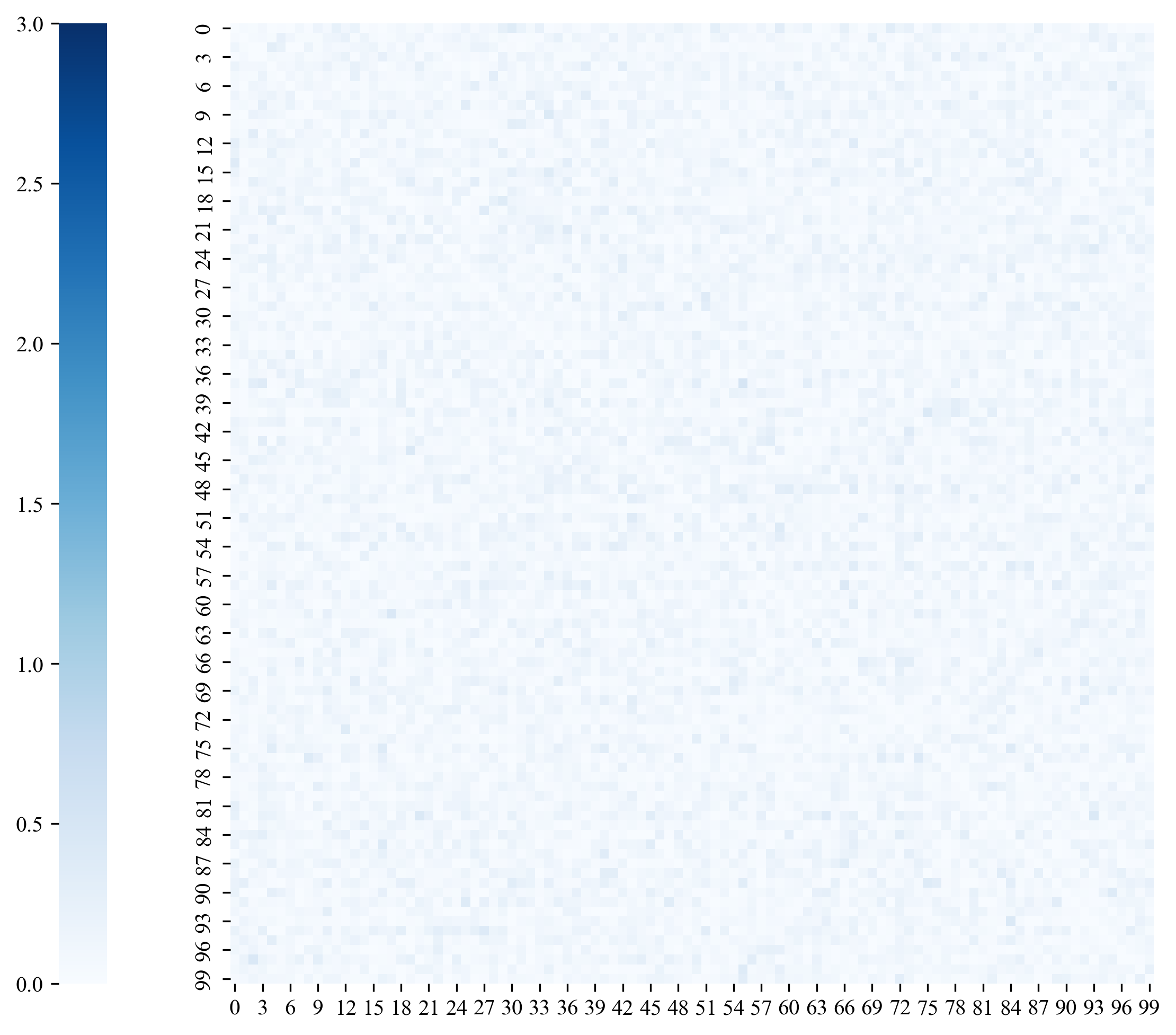}
  \subcaption{\model+KD \\ Mean: 0.09\\ Max: 0.51}
\end{subfigure}
\caption{\textbf{Logit correlation analysis.} Matrix of the average logit difference between teacher and student outputs on CIFAR-100 (lower values indicate higher similarity). Results are based on a WRN-40-2 teacher and a WRN-40-1 student.}
\label{fig:correlation}
\end{figure}
\begin{figure}[!t]
\centering
\begin{subfigure}[b]{0.16\textwidth}
  \centering
  \includegraphics[width=\linewidth]{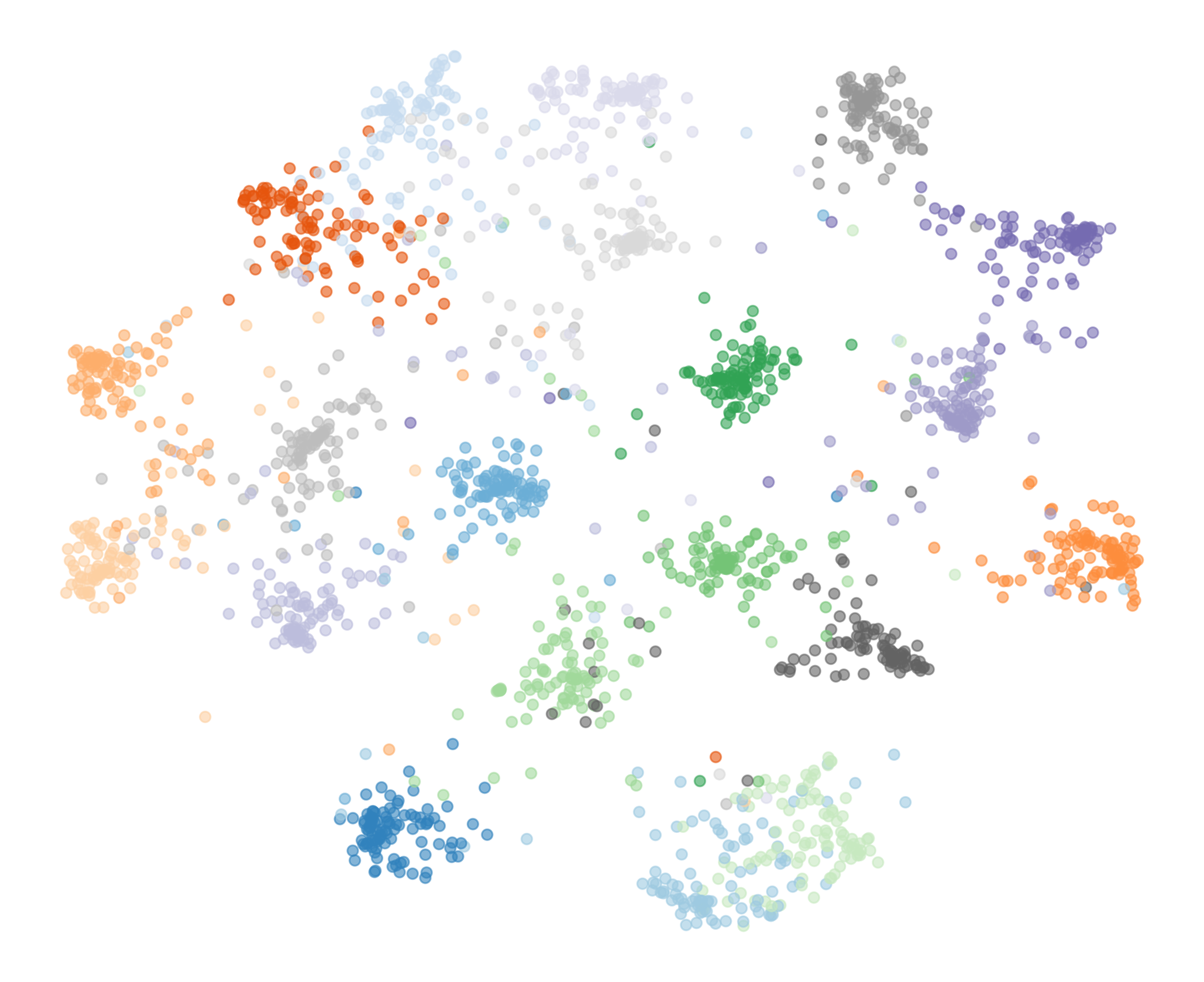}
  \subcaption{Teacher}
\end{subfigure}
\hfill
\begin{subfigure}[b]{0.16\textwidth}
  \centering
  \includegraphics[width=\linewidth]{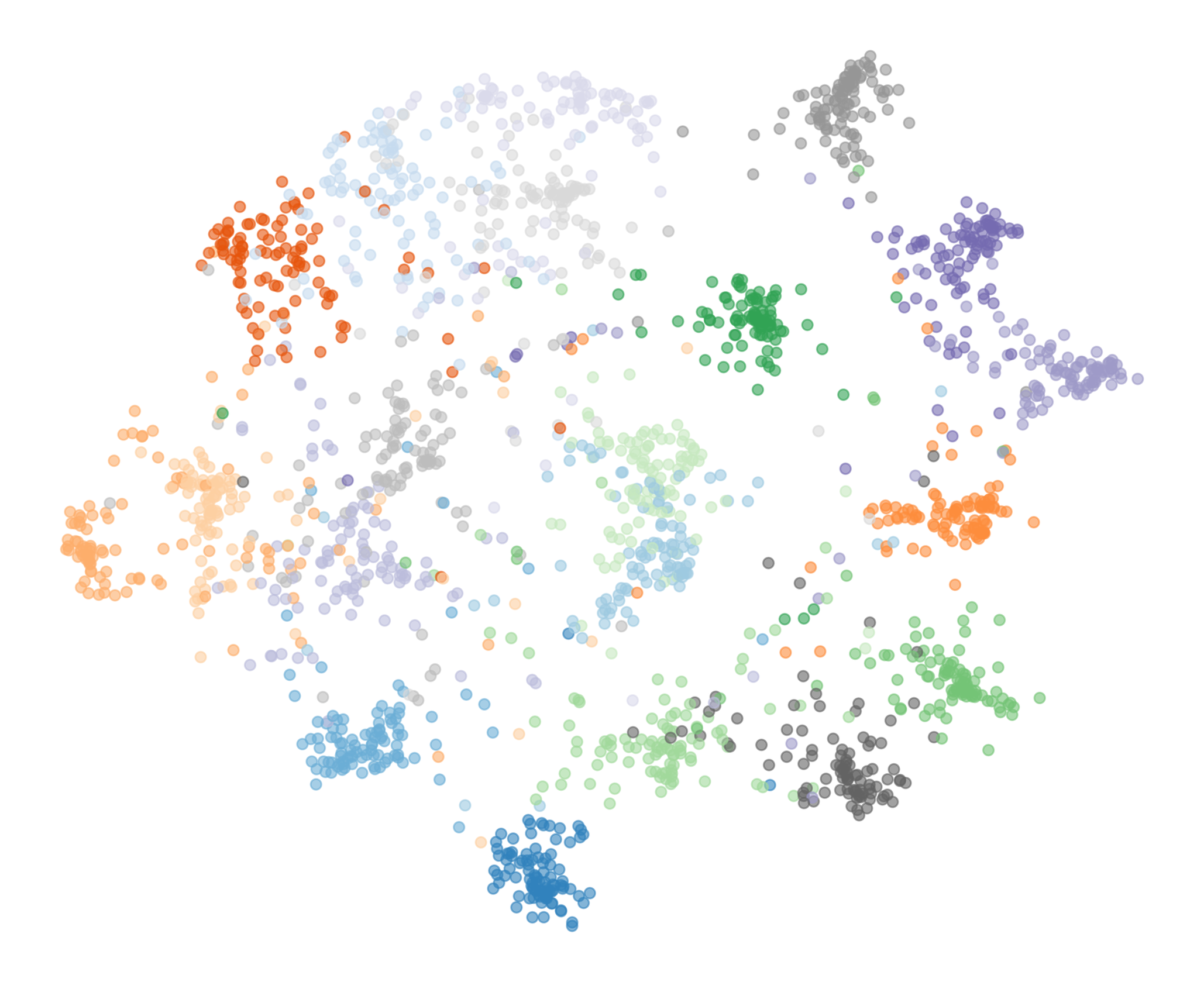}
  \subcaption{Vanilla}
\end{subfigure}
\hfill
\begin{subfigure}[b]{0.16\textwidth}
  \centering
  \includegraphics[width=\linewidth]{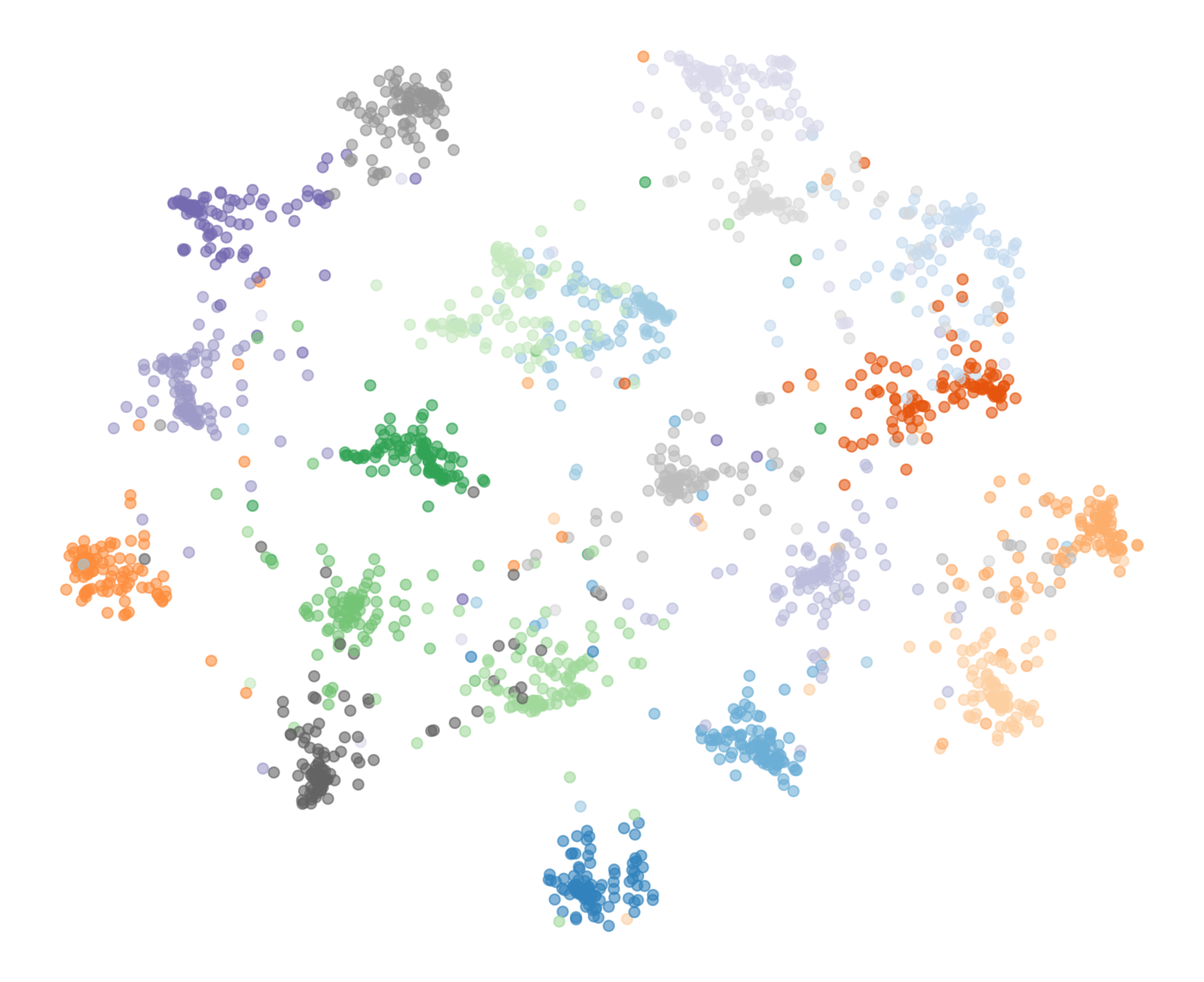}
  \subcaption{KD \cite{hinton2015distilling}}
\end{subfigure}
\hfill
\begin{subfigure}[b]{0.16\textwidth}
  \centering
  \includegraphics[width=\linewidth]{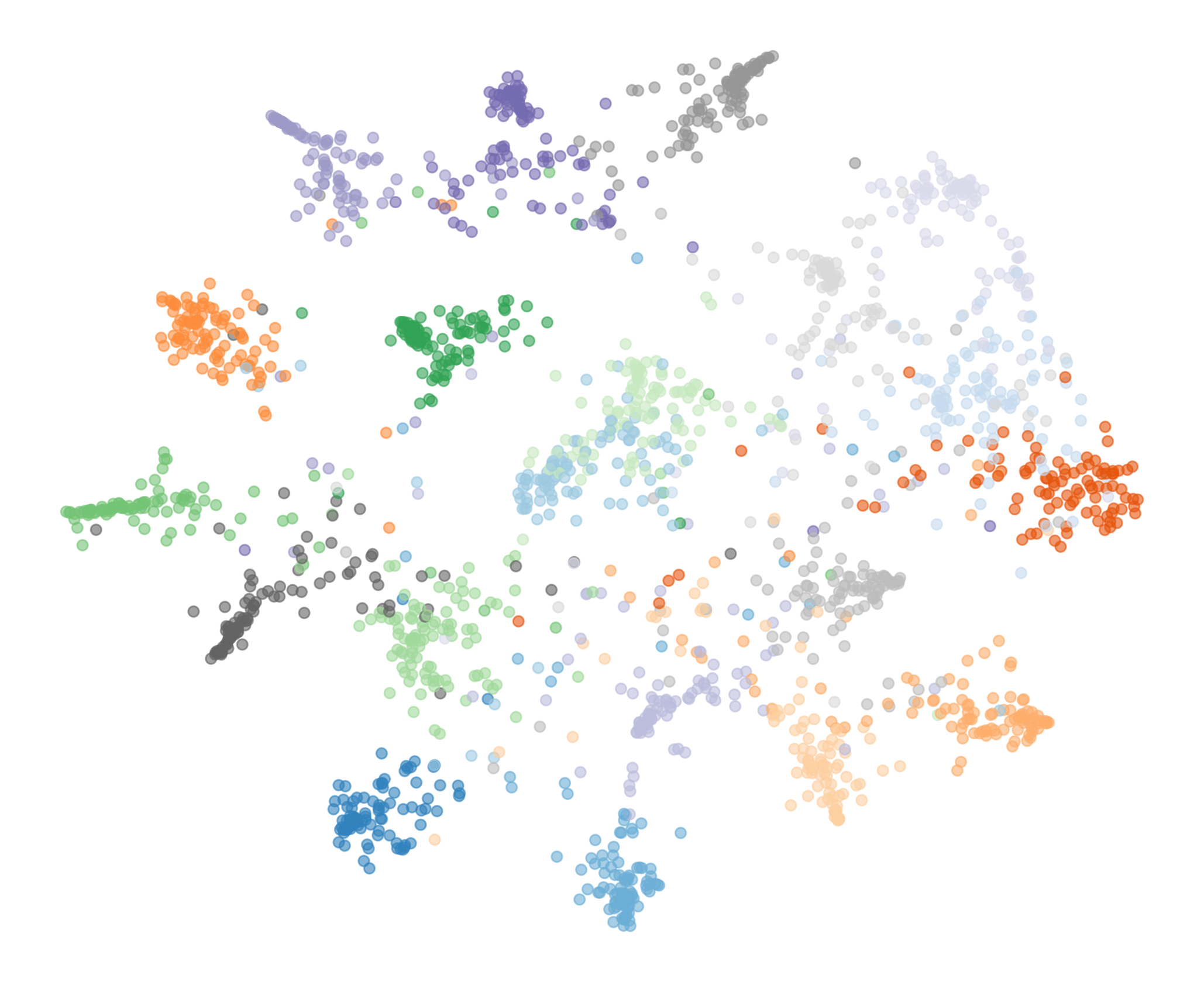}
  \subcaption{CRD \cite{tian2022crd}}
\end{subfigure}
\hfill
\begin{subfigure}[b]{0.16\textwidth}
  \centering
  \includegraphics[width=\linewidth]{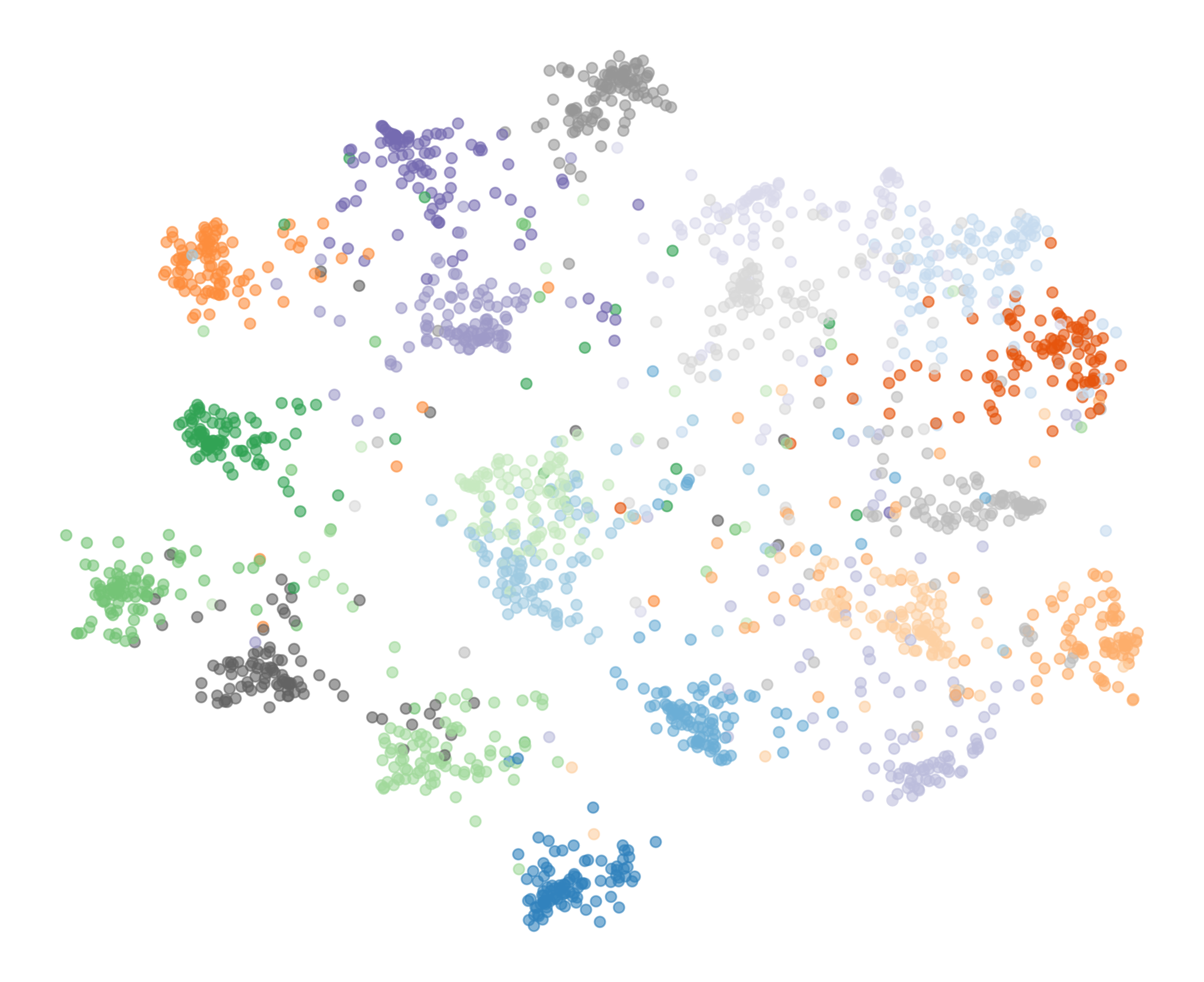}
  \subcaption{\model}
\end{subfigure}
\hfill
\begin{subfigure}[b]{0.16\textwidth}
  \centering
  \includegraphics[width=\linewidth]{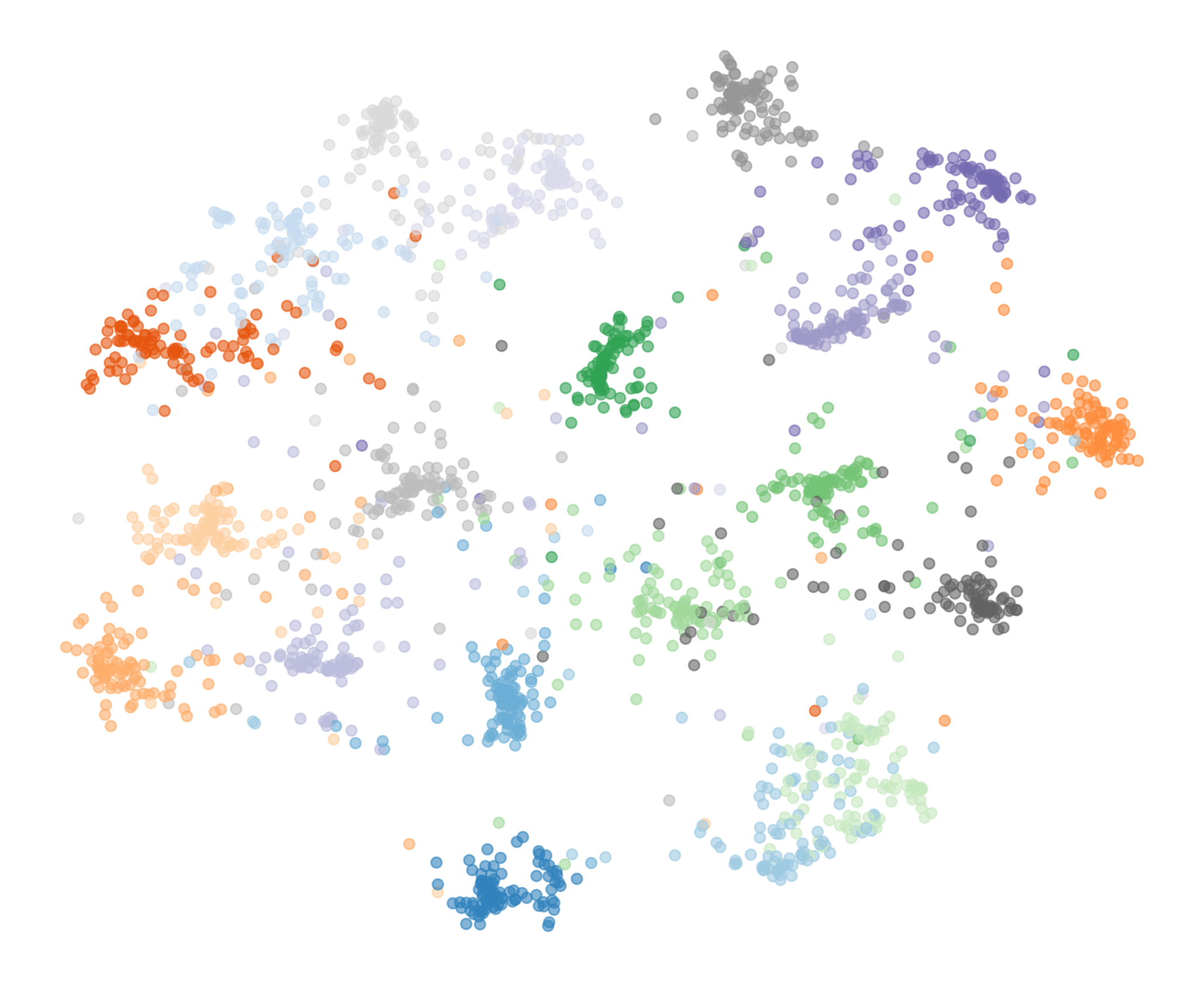}
  \subcaption{\model+KD}
\end{subfigure}
\caption{\textbf{t-SNE visualization of embedding spaces.} Comparison of feature distributions from the teacher and student networks for the first 20 classes of CIFAR-100. We use a WRN-40-2 teacher and a WRN-40-1 student.}
\label{fig:tsne}
\end{figure}

We analyze the learned representations through qualitative visualizations that illustrate the knowledge transfer patterns of different distillation approaches.

\paragraph{\bf Inter-class correlations.}
We visualize the logit correlation matrices~\cite{tian2022crd} on CIFAR-100 using WRN-40-2 (teacher) and WRN-40-1 (student). \Cref{fig:correlation} shows the difference between teacher and student correlation structures for various distillation methods. Compared to vanilla training and standard KD, our method substantially reduces the discrepancy between teacher and student correlations, demonstrating stronger and more consistent alignment in logit space.

\paragraph{\bf t-SNE.}
We visualize the feature embeddings on CIFAR-100 using t-SNE~\cite{vandermaaten08tsne} for the same teacher-student pair. \Cref{fig:tsne} shows that our method achieves closer alignment of student features with the teacher while preserving meaningful semantic class structure across clusters, outperforming standard training, KD, and several strong contrastive distillation baselines.

\subsection{Ablation Study}
\label{subsec:ablations}

\begin{figure}[!t]
\centering
\begin{subfigure}[b]{0.24\linewidth}
  \centering
  \includegraphics[width=\linewidth]{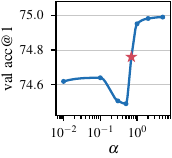}
  \subcaption{}
  \label{fig:ablation_alpha}
\end{subfigure}
\hfill
\begin{subfigure}[b]{0.24\linewidth}
  \centering
  \includegraphics[width=\linewidth]{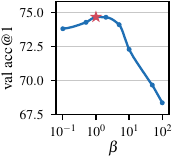}
  \subcaption{}
  \label{fig:ablation_beta}
\end{subfigure}
\hfill
\begin{subfigure}[b]{0.24\linewidth}
  \centering
  \includegraphics[width=\linewidth]{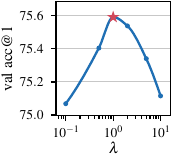}
  \subcaption{}
  \label{fig:ablation_lambda}
\end{subfigure}
\hfill
\begin{subfigure}[b]{0.24\linewidth}
  \centering
  \includegraphics[width=\linewidth]{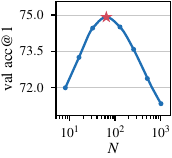}
  \subcaption{}
  \label{fig:ablation_batch}
\end{subfigure}
\caption{\textbf{Hyper-parameter ablation on CIFAR-100.} Test top-1 accuracy (\%) for the WRN-40-2 teacher and WRN-16-2 student pair. (a) Loss weight $\alpha$. (b) Loss weight $\beta$. (c) Loss weight $\lambda$. (d) Batch size $N$. Stars mark the selected setting, which is used in all other experiments. Curves are smoothed with a Gaussian kernel ($\sigma = 1.0$ samples).}
\label{fig:ablation_study}
\end{figure}

We ablate our method on CIFAR-100 in \Cref{tab:ablation_config,fig:ablation_study}. Starting from a purely contrastive variant ($\alpha=0$) with a fixed scale, we add the consistency term and then make the scale and bias learnable, reporting both with and without KD for the ResNet-50 to MobileNet-v2 and WRN-40-2 to WRN-16-2 pairs (\Cref{tab:ablation_mnv2,tab:ablation_wrn}). We then study coefficient sensitivity on WRN-40-2 to WRN-16-2: $\alpha$, balancing $\mathcal{L}_{\text{ctr}}$ and $\mathcal{L}_{\text{cst}}$ (\Cref{eq:dcd_loss}), is stable across $\{0.01, \dots, 5\}$ with $\beta=1$, $\lambda=0$ (\Cref{fig:ablation_alpha}), consistent with the learnable scale absorbing changes in the relative magnitude of the two terms; $\beta$, weighting $\mathcal{L}_{\text{dcd}}$ against the supervised loss, is best between $0.5$ and $10$, with very high values degrading performance as the feature loss overwhelms the supervised term (\Cref{fig:ablation_beta}); and $\lambda$, weighting $\mathcal{L}_{\text{distill}}$, is stable across $\{0.1, \dots, 10\}$ and best at $\lambda=1.0$, while $\lambda=50$ or $100$ causes collapse, confirming the $\lambda=1.0$ of prior work~\cite{tian2022crd, chen2021semckd, chen2022simkd} (\Cref{fig:ablation_lambda}). Finally, since negatives are sampled in-batch, the batch size determines their number, with optimal results at 64 and degradation at smaller and larger sizes (\Cref{fig:ablation_batch}). This is the principal limitation of removing the memory bank: unlike CRD, whose negative pool is decoupled from the batch, our negative count is fixed by the training configuration.

\begin{table}[!t]
\centering
\setlength{\tabcolsep}{1.2mm}
\caption{\textbf{Configuration ablation on CIFAR-100.} Test top-1 accuracy (\%) as the components of \model{} are enabled one at a time, without and with the KD term.}
\label{tab:ablation_config}

\begin{subtable}[b]{0.48\linewidth}
\centering
\begin{tabular}{ccccc}
\toprule
\multicolumn{3}{c}{Configuration} & \multicolumn{2}{c}{acc@1} \\
\cmidrule(lr){1-3} \cmidrule(lr){4-5}
$\alpha$ & $\tau$ & $b$ & w/o KD & w/ KD \\
\midrule
0   & 0.07    & 0       & 65.52 & 69.09 \\
0.5 & 0.07    & 0       & 66.35 & 69.53 \\
0.5 & learned & learned & \cellcolor{gray!12}\textbf{67.39} & \cellcolor{gray!12}\textbf{70.03} \\
\bottomrule
\end{tabular}
\subcaption{T: MobileNet-v2, S: ResNet-50}
\label{tab:ablation_mnv2}
\end{subtable}
\hfill
\begin{subtable}[b]{0.48\linewidth}
\centering
\begin{tabular}{ccccc}
\toprule
\multicolumn{3}{c}{Configuration} & \multicolumn{2}{c}{acc@1} \\
\cmidrule(lr){1-3} \cmidrule(lr){4-5}
$\alpha$ & $\tau$ & $b$ & w/o KD & w/ KD \\
\midrule
0   & 0.07    & 0       & 74.37 & 75.51 \\
0.5 & 0.07    & 0       & 74.44 & 75.43 \\
0.5 & learned & learned & \cellcolor{gray!12}\textbf{74.99} & \cellcolor{gray!12}\textbf{76.06} \\
\bottomrule
\end{tabular}
\subcaption{T: WRN-16-2, S: WRN-40-2}
\label{tab:ablation_wrn}
\end{subtable}

\end{table}

\section{Conclusion}
\label{sec:conclusion}

We presented an objective combining instance-level contrastive alignment with a consistency regularization term, where the contrastive term discriminates instances on the diagonal of the cross-model similarity matrix while the consistency term constrains its off-diagonal structure by penalizing asymmetry between its two normalizations, and a learnable scale and bias adapt the sharpness and offset of distillation without manual tuning. In-batch negative sampling replaces external memory banks, adding just 66K parameters, all discarded at inference, and matches logit-based methods at 8ms per batch, making ours the fastest feature-based method evaluated, with competitive results in classification, object detection, and cross-dataset transfer. Unlike CRD~\cite{tian2022crd}, which needs a memory bank spanning the training set and fixed hyperparameters, we operate within the mini-batch with learnable parameters; unlike RKD~\cite{park2019relational} or CC~\cite{peng2019correlation}, which compare teacher-only against student-only geometry or correlation statistics, our consistency term acts on the cross-model matrix the contrastive loss already computes, at no extra cost; and while AT~\cite{zagoruyko2016paying} and FitNets~\cite{romero2014fitnets} target specific aspects of the teacher's knowledge and ReviewKD~\cite{chen2021reviewkd} and SimKD~\cite{chen2022simkd} rely on architectural innovations, our objective captures both discriminative and structural information through algorithmic refinements alone, with no architectural changes.

\section*{Acknowledgements}
We acknowledge the computational resources and support provided by the Imperial College Research Computing Service (\url{http://doi.org/10.14469/hpc/2232}), which enabled our experiments.

%
%
\bibliographystyle{splncs04}
\bibliography{main}

\appendix
\clearpage

\section{Algorithm}
\label{app:algorithm}

\Cref{alg:pseudocode} provides the pseudo-code of \model, corresponding to $\mathcal{L}_{\text{dcd}}$ in \Cref{eq:dcd_loss} of the main paper. The supervised loss $\mathcal{L}_{\text{sup}}$, the logit-space term $\lambda \cdot \mathcal{L}_{\text{distill}}$, and the outer weight $\beta$ are omitted for clarity. Two details are worth highlighting because they determine the form of the loss. First, the column-wise softmax is transposed before the divergence is taken, so that each row of the resulting matrix is a distribution over the $N$ samples and the KL divergence is taken between two vectors on the same probability simplex. Second, \texttt{kl\_div} in PyTorch takes log-probabilities as its first argument and probabilities as its second, and computes $D_{\text{KL}}(\text{target} \| \text{input})$; with \texttt{reduction="batchmean"} the result therefore equals \Cref{eq:consistency} of the main paper exactly, including the $1/N$ factor.

\begin{algorithm}[!ht]
\caption{Pseudocode of \model in a PyTorch-like style.}
\label{alg:pseudocode}
\algcomment{\fontsize{7.2pt}{0em}\selectfont
\texttt{mm}: matrix multiplication;
\texttt{t}: matrix transpose;
\texttt{normalize}: $\ell_2$ normalization along a dimension;
\texttt{softmax}: exponential normalization across a dimension;
\texttt{log\_softmax}: logarithm of softmax;
\texttt{cross\_entropy}: negative log likelihood loss;
\texttt{kl\_div}: Kullback-Leibler divergence, taking log-probabilities first and probabilities second, so that \texttt{kl\_div(p, q)} computes $D_{\text{KL}}(q \| \exp(p))$
}\definecolor{codeblue}{rgb}{0.25,0.5,0.5}
\lstset{
  backgroundcolor=\color{white},
  basicstyle=\fontsize{7.2pt}{7.2pt}\ttfamily\selectfont,
  columns=fullflexible,
  breaklines=true,
  captionpos=b,
  commentstyle=\fontsize{7.2pt}{7.2pt}\color{codeblue},
  keywordstyle=\fontsize{7.2pt}{7.2pt},
}
\begin{lstlisting}[language=python]
# f_s, f_t: student and teacher backbones (f_t frozen)
# g_s, g_t: linear projection heads, s_dim->128 and t_dim->128 (both trained)
# tau, b: learnable scale and bias, initialized to 1.0 and 0.0
# s_max: upper bound on the logit scale exp(tau), set to 10.0
# alpha: weight of the consistency term, set to 0.5

for x in loader:  # load a minibatch x with N samples
    h_s = f_s.forward(x)  # student features: NxD_s

    with torch.no_grad():  # no gradients for the teacher backbone
        h_t = f_t.forward(x)  # teacher features: NxD_t

    z_s = normalize(g_s(h_s), dim=1)  # student embeddings: Nx128
    z_t = normalize(g_t(h_t), dim=1)  # teacher embeddings: Nx128

    # scaled cross-model similarity: rows index students, columns teachers
    scale = tau.exp()
    logits = torch.mm(z_s, z_t.t()) * scale + b  # NxN

    # contrastive loss: softmax over rows, positives on the diagonal
    labels = torch.arange(N)  # diagonal labels
    contrastive_loss = cross_entropy(logits, labels)

    # consistency loss: row-normalized view vs column-normalized view
    p1 = log_softmax(logits, dim=1)   # rows sum to one
    p2 = softmax(logits, dim=0).t()   # transpose so rows sum to one
    consistency_loss = kl_div(p1, p2, reduction="batchmean")

    # feature-space objective (L_sup and the logit KD term are omitted)
    loss = contrastive_loss + alpha * consistency_loss

    # SGD update: student network, both projections, and both scalars
    loss.backward()
    update(f_s.params)
    update(g_s.params); update(g_t.params)
    update(tau)  # update scale
    update(b)    # update bias

    # bound the scale after the step, so the gradient of tau is never zeroed
    tau.data.clamp_(max=log(s_max))
\end{lstlisting}
\end{algorithm}

\section{Experimental Setup}
\label{app:setup}

We implement \model in PyTorch following the protocol of \texttt{RepDistiller}\cite{tian2022crd}\footnote{Available at: \url{https://github.com/HobbitLong/RepDistiller}.}. This protocol has become a common standard and is widely used among many papers to demonstrate their knowledge distillation methods \cite{hinton2015distilling, romero2014fitnets, zagoruyko2016paying, tung2019similarity, peng2019correlation, ahn2019variational, park2019relational, passalis2018probabilistic, heo2019knowledge, kim2018paraphrasing, yim2017gift, huang2017like, tian2022crd}, allowing for fair comparison across methods. For visualization, we follow the implementation of \cite{sun2024logitstandardization}\footnote{Available at: \url{https://github.com/sunshangquan/logit-standardization-KD}.}. For the efficiency measurements we use the \texttt{MDistiller} framework \cite{zhao2022dkd}\footnote{Available at: \url{https://github.com/megvii-research/mdistiller}}.

\subsection{Baseline Methods}
\label{app:baseline_methods}

We compare our approach to the following methods from the literature: (1) Distilling the Knowledge in a Neural Network (KD) \cite{hinton2015distilling}; (2) FitNets: Hints for Thin Deep Nets \cite{romero2014fitnets}; (3) Paying More Attention to Attention (AT) \cite{zagoruyko2016paying}; (4) Similarity-Preserving Knowledge Distillation (SP) \cite{tung2019similarity}; (5) Correlation Congruence for Knowledge Distillation (CC) \cite{peng2019correlation}; (6) Variational Information Distillation for Knowledge Transfer (VID) \cite{ahn2019variational}; (7) Relational Knowledge Distillation (RKD) \cite{park2019relational}; (8) Learning Deep Representations with Probabilistic Knowledge Transfer (PKT) \cite{passalis2018probabilistic}; (9) Knowledge Transfer via Distillation of Activation Boundaries Formed by Hidden Neurons (AB) \cite{heo2019knowledge}; (10) Paraphrasing Complex Network: Network Compression via Factor Transfer (FT) \cite{kim2018paraphrasing}; (11) A Gift from Knowledge Distillation: Fast Optimization, Network Minimization and Transfer Learning (FSP) \cite{yim2017gift}; (12) Like What You Like: Knowledge Distill via Neuron Selectivity Transfer (NST) \cite{huang2017like}; (13) Contrastive Representation Distillation (CRD) \cite{tian2022crd}; (14) A Comprehensive Overhaul of Feature Distillation (OFD) \cite{heo2019ofd}; (15) Rethinking Soft Labels for Knowledge Distillation: A Bias-Variance Tradeoff Perspective (WSLD) \cite{zhou2021wsld}; (16) Respecting Transfer Gap in Knowledge Distillation (IPWD) \cite{niu2022ipwd}; (17) Knowledge Distillation via Softmax Regression Representation Learning (SRRL) \cite{yang2021srrl}; (18) Cross-Layer Distillation with Semantic Calibration (SemCKD) \cite{chen2021semckd}; (19) Distilling Knowledge via Knowledge Review (ReviewKD) \cite{chen2021reviewkd}; (20) Knowledge Distillation with the Reused Teacher Classifier (SimKD) \cite{chen2022simkd}; (21) DistPro: Searching a Fast Knowledge Distillation Process via Meta Optimization (DistPro) \cite{deng2022distpro}; (22) Knowledge Distillation via N-to-One Representation Matching (NORM) \cite{liu2023norm}; (23) Wasserstein Contrastive Representation Distillation (WCoRD) \cite{chen2021wcord}; (24) Complementary Relation Contrastive Distillation (CRCD) \cite{zhu2021crcd}; (25) Feature Kernel Distillation (FKD) \cite{he2022fkd}; (26) Information Theoretic Representation Distillation (ITRD) \cite{miles2022itrd}; (27) Transformed Teacher Matching (TTM) and its weighted variant (WTTM) \cite{zheng2024ttm}; (28) Decoupled Knowledge Distillation (DKD) \cite{zhao2022dkd}; (29) Function Matching and Feature Consistency for Knowledge Distillation (FCFD) \cite{liu2023fcfd}; (30) Class Attention Transfer Based Knowledge Distillation (CAT-KD) \cite{guo2023cat}; (31) Knowledge Distillation from A Stronger Teacher (DIST) \cite{huang2022dist}; (32) Curriculum Temperature for Knowledge Distillation (CTKD) \cite{li2022ctkd}.

\subsection{Datasets}
\label{app:datasets}

We take five widely researched datasets: (1) CIFAR-100 \cite{krizhevsky2009cifar} is a standard benchmark for knowledge distillation and contains 50,000 training images of size $32\times32$ with 500 images per class and 10,000 test images. (2) ImageNet ILSVRC-2012 \cite{deng2009imagenet}, which is more challenging than CIFAR, and includes 1.2 million images from 1,000 classes for training and 50,000 for validation. (3) STL-10 \cite{coates2011importance} consists of a training set of 5,000 labeled images from 10 classes, and a test set of 8,000 images. (4) Tiny ImageNet \cite{deng2009imagenet} has 200 classes, each with 500 training images and 50 validation images. (5) MS-COCO \cite{li2015coco} is an 80-category general object detection dataset. The $\texttt{train2017}$ split contains 118,000 images, and the $\texttt{val2017}$ split contains 5,000 images.

\subsection{Network Architectures}
\label{app:network_arch}

We use the following network architectures as described in \cite{tian2022crd}: (1) Wide Residual Network (WRN) \cite{zagoruyko2017wide}, where WRN-$d$-$w$ represents a wide ResNet with depth \(d\) and width factor \(w\); (2) ResNet (rn) \cite{he2015resnet}, where rn-$d$ represents a CIFAR-style ResNet with 3 groups of basic blocks having 16, 32, and 64 channels, respectively, and rn-8 \(\times 4\) and rn-32 \(\times 4\) indicate a 4-times wider network with 64, 128, and 256 channels; (3) ResNet (RN) \cite{he2015resnet}, where RN-$d$ represents an ImageNet-style ResNet with Bottleneck blocks and more channels; (4) MobileNet-v2 (MN-v2) \cite{sandler2018mobilenetv2}, using a width multiplier of 0.5 in our experiments; (5) VGG \cite{simonyan2015vgg}, where the VGG network used is adapted from its original ImageNet counterpart; and (6) ShuffleNet-v1 (SN-v1) \cite{zhang2018shufflenet} and ShuffleNet-v2 (SN-v2) \cite{ma2018shufflenet}, which are adapted for efficient training with input sizes of \(32 \times 32\). 
To ensure dimensional consistency between the two networks, the penultimate-layer features of the teacher and the student are each mapped into a shared 128-dimensional embedding space by a single linear \textit{projection layer}, followed by $\ell_2$ normalization. Both projections are trained jointly with the student, including the teacher-side projection, even though the teacher backbone is frozen, and both are discarded after training so that they add no inference cost. For the rn-32$\times$4 to rn-8$\times$4 pair, where both feature dimensions are 256, the two projections account for $2 \times (256 \times 128 + 128) = 65{,}792$ parameters, which together with the scalars $\tau$ and $b$ gives the 65,794 additional trainable parameters reported in the main paper; the count varies with the teacher and student feature dimensions. Prior analysis suggests that such projections do more than reconcile dimensions, and can themselves contribute to the transfer by shaping the geometry of the shared embedding space~\cite{miles2024understanding}.

\subsection{Optimization}
\label{app:optimization}

We closely follow the protocol of \cite{tian2022crd}. All methods evaluated in our experiments use SGD with 0.9 Nesterov momentum. 
For CIFAR-100, we initialize the learning rate as 0.05, and decay it by 0.1 every 30 epochs after the first 150 epochs until the last 240 epoch. For MobileNet-v2, ShuffleNet-v1, and ShuffleNet-v2, we use a learning rate of 0.01 as this learning rate is optimal for these models in a grid search, while 0.05 is optimal for other models. The batch size is set to 64 for CIFAR-100, and the weight decay is set to $5 \times 10^{-4}$. 
For ImageNet, the initial learning rate is set to 0.1 and then divided by 10 at the 30th, 60th, and 90th epochs of the total 120 training epochs. The mini-batch size is set to 256, and the weight decay is set to $1 \times 10^{-4}$. The learnable scale and bias are optimized with the same optimizer and learning rate as the student, and the scale is clamped to $s_{\max} = 10$ after each step. 
Our implementation for MS-COCO follows the settings in~\cite{zhao2022dkd}. We use the two-stage method Faster R-CNN~\cite{ren2016fasterrcnn} with Feature Pyramid Network (FPN)~\cite{lin2017feature} as the detection framework. We evaluate three teacher-student settings: ResNet-101 to ResNet-18, ResNet-101 to ResNet-50, and ResNet-50 to MobileNet-V2~\cite{sandler2018mobilenetv2}. All students are trained with the $1\times$ scheduler, with schedulers and task-specific loss weights following \texttt{Detectron2}~\cite{wu2019detectron2}. 
We take one RTX 6000 GPU to train the model on CIFAR-100 and four L40 GPUs on ImageNet and MS-COCO.

\section{Extended Results}
\label{app:extended_results}

\subsection{CIFAR-100}
\label{app:extended_results_cifar100}

\Cref{tab:cifar100_same_appendix,tab:cifar100_diff_appendix} provide a comprehensive overview of the top-1 accuracies of student networks trained with various distillation techniques across a wide range of teacher-student architectural combinations. Our method benefits from its simplicity: the only parameters it adds are the two linear projections and the two scalars $\tau$ and $b$, all of which are discarded after training, and the only quantities requiring selection are the loss coefficients $\alpha$, $\beta$, and $\lambda$, since the scale that a fixed-temperature objective would require tuning is learned instead.

\subsection{Inter-class Correlations}
\label{app:extended_results_correlations}

\begin{figure*}[!p]
\centering
\begin{subfigure}[b]{0.23\textwidth}
  \centering
  \includegraphics[width=\linewidth]{figures/correlation_VANILLA_wrn_40_1_wrn_40_2.png}
  \subcaption{Vanilla \\ Mean: 0.24, Max: 1.66}
\end{subfigure}
\hfill
\begin{subfigure}[b]{0.23\textwidth}
  \centering
  \includegraphics[width=\linewidth]{figures/correlation_KD_wrn_40_1_wrn_40_2.png}
  \subcaption{KD \cite{hinton2015distilling} \\ Mean: 0.09, Max: 0.49}
\end{subfigure}
\hfill
\begin{subfigure}[b]{0.23\textwidth}
  \centering
  \includegraphics[width=\linewidth]{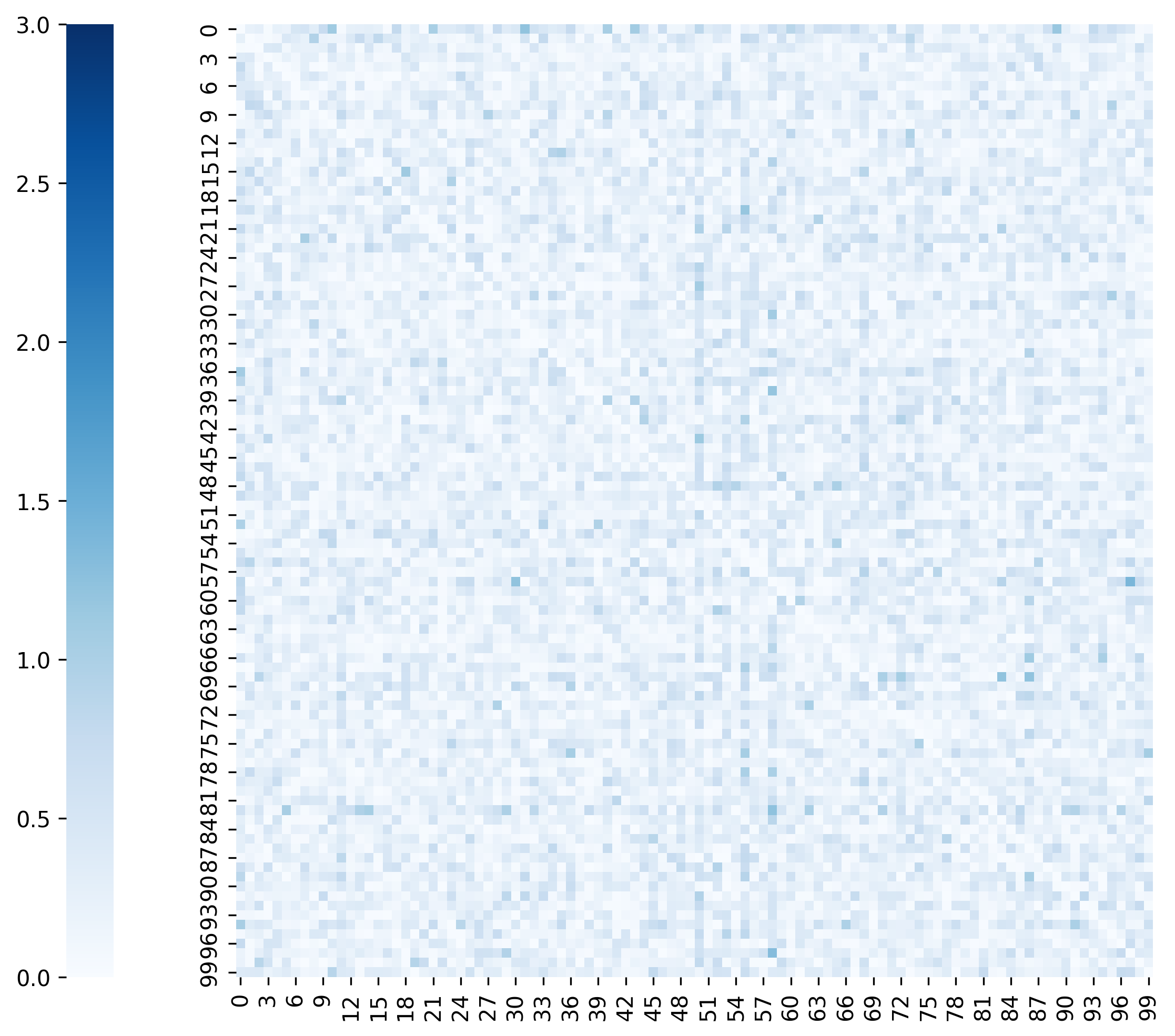}
  \subcaption{FitNets \cite{romero2014fitnets} \\ Mean: 0.25, Max: 1.38}
\end{subfigure}
\hfill
\begin{subfigure}[b]{0.23\textwidth}
  \centering
  \includegraphics[width=\linewidth]{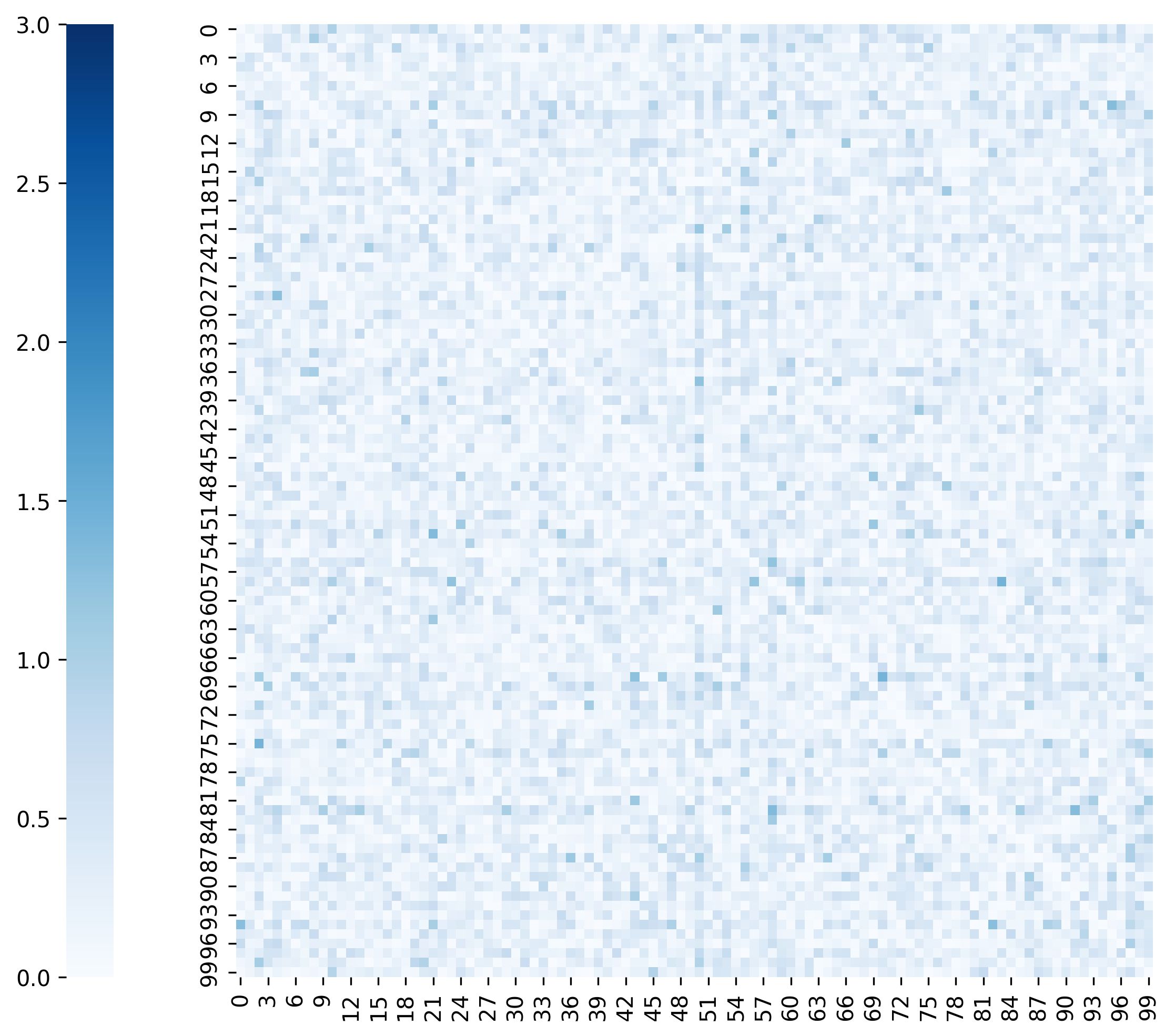}
  \subcaption{AT \cite{zagoruyko2016paying} \\ Mean: 0.26, Max: 1.43}
\end{subfigure}
\begin{subfigure}[b]{0.23\textwidth}
  \centering
  \includegraphics[width=\linewidth]{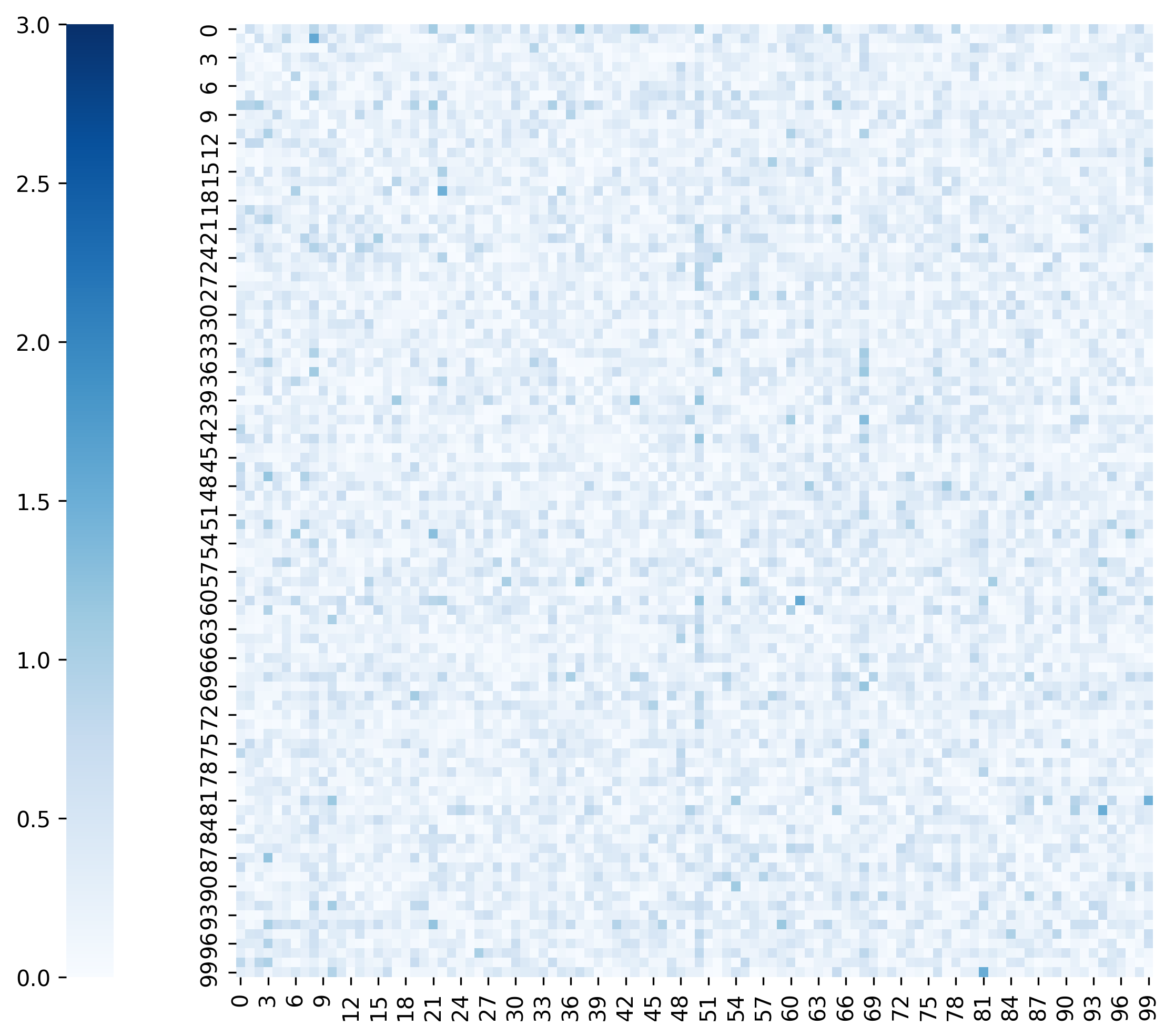}
  \subcaption{SP \cite{tung2019similarity} \\ Mean: 0.25, Max: 1.57}
\end{subfigure}
\hfill
\begin{subfigure}[b]{0.23\textwidth}
  \centering
  \includegraphics[width=\linewidth]{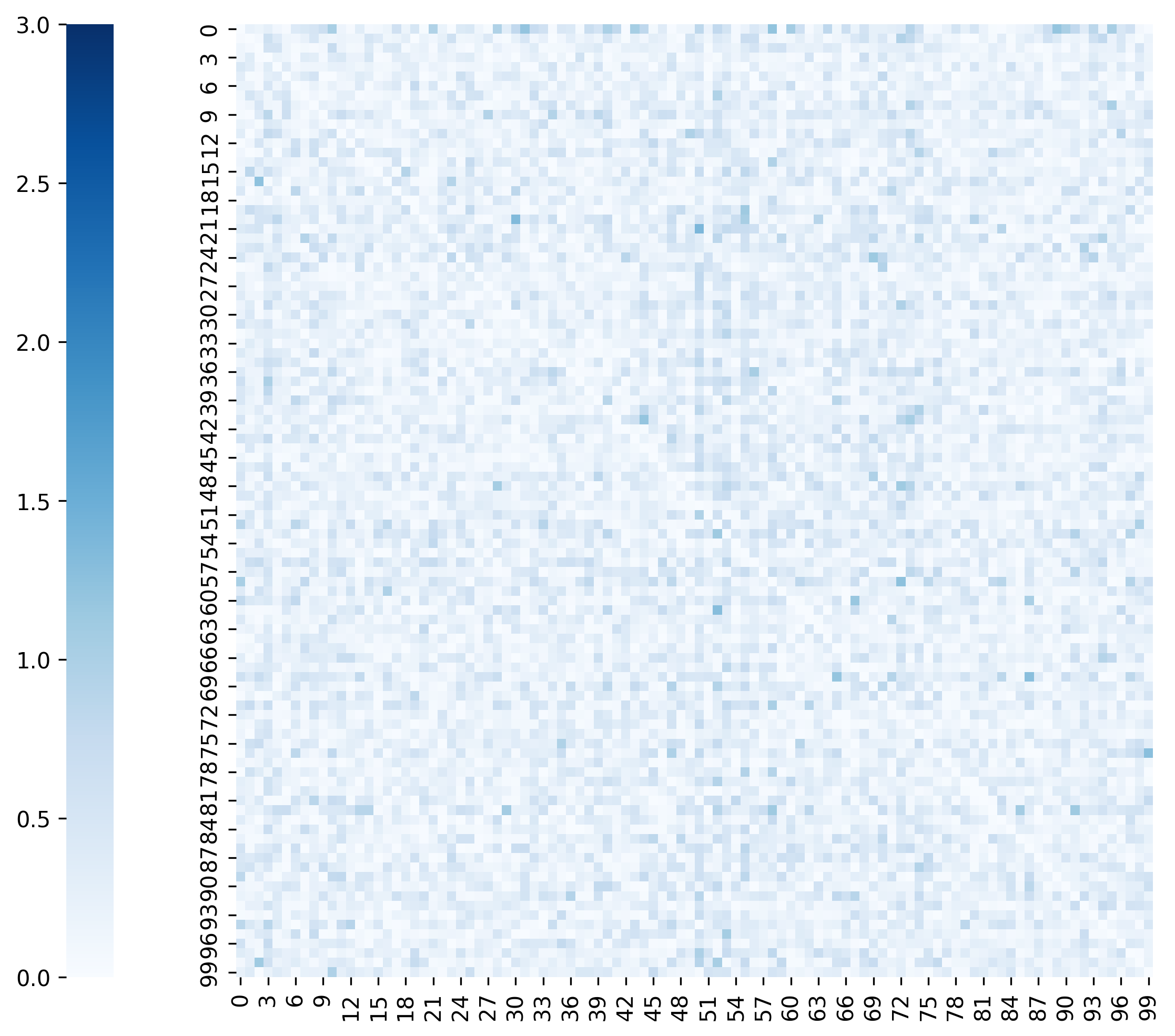}
  \subcaption{CC \cite{peng2019correlation} \\ Mean: 0.234, Max: 1.37}
\end{subfigure}
\hfill
\begin{subfigure}[b]{0.23\textwidth}
  \centering
  \includegraphics[width=\linewidth]{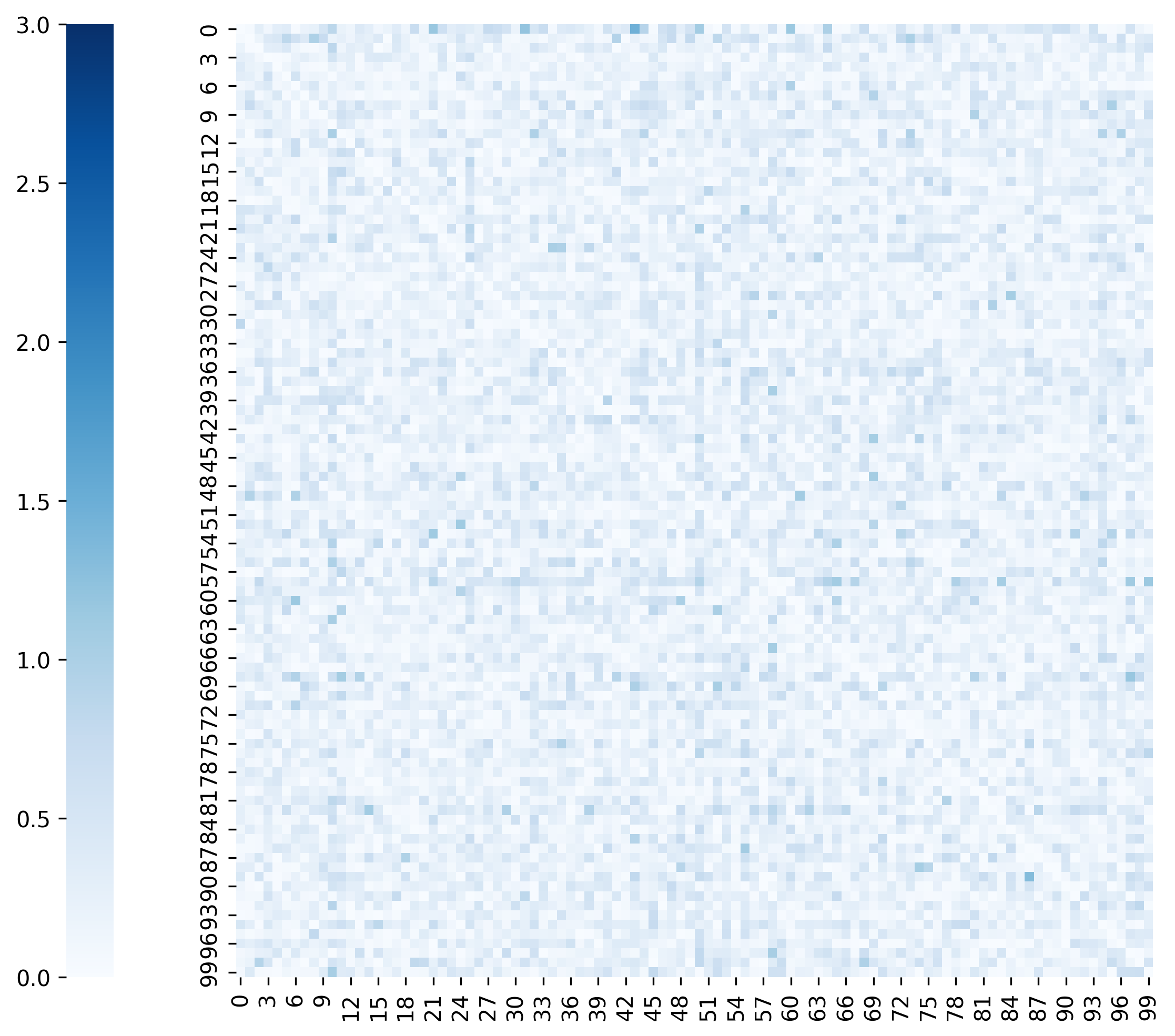}
  \subcaption{VID \cite{ahn2019variational} \\ Mean: 0.23, Max: 1.46}
\end{subfigure}
\hfill
\begin{subfigure}[b]{0.23\textwidth}
  \centering
  \includegraphics[width=\linewidth]{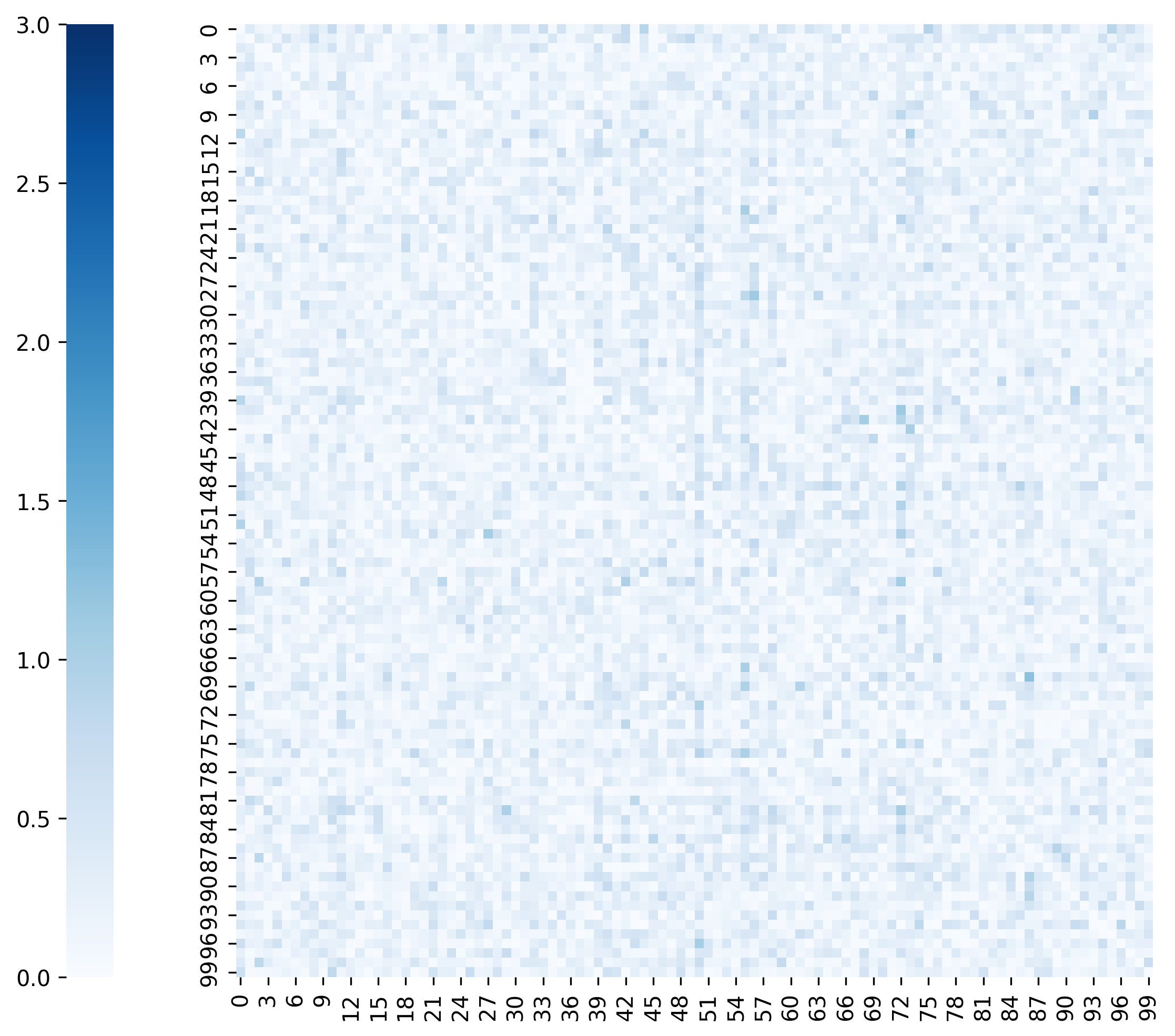}
  \subcaption{RKD \cite{park2019relational} \\ Mean: 0.21, Max: 1.24}
\end{subfigure}
\begin{subfigure}[b]{0.23\textwidth}
  \centering
  \includegraphics[width=\linewidth]{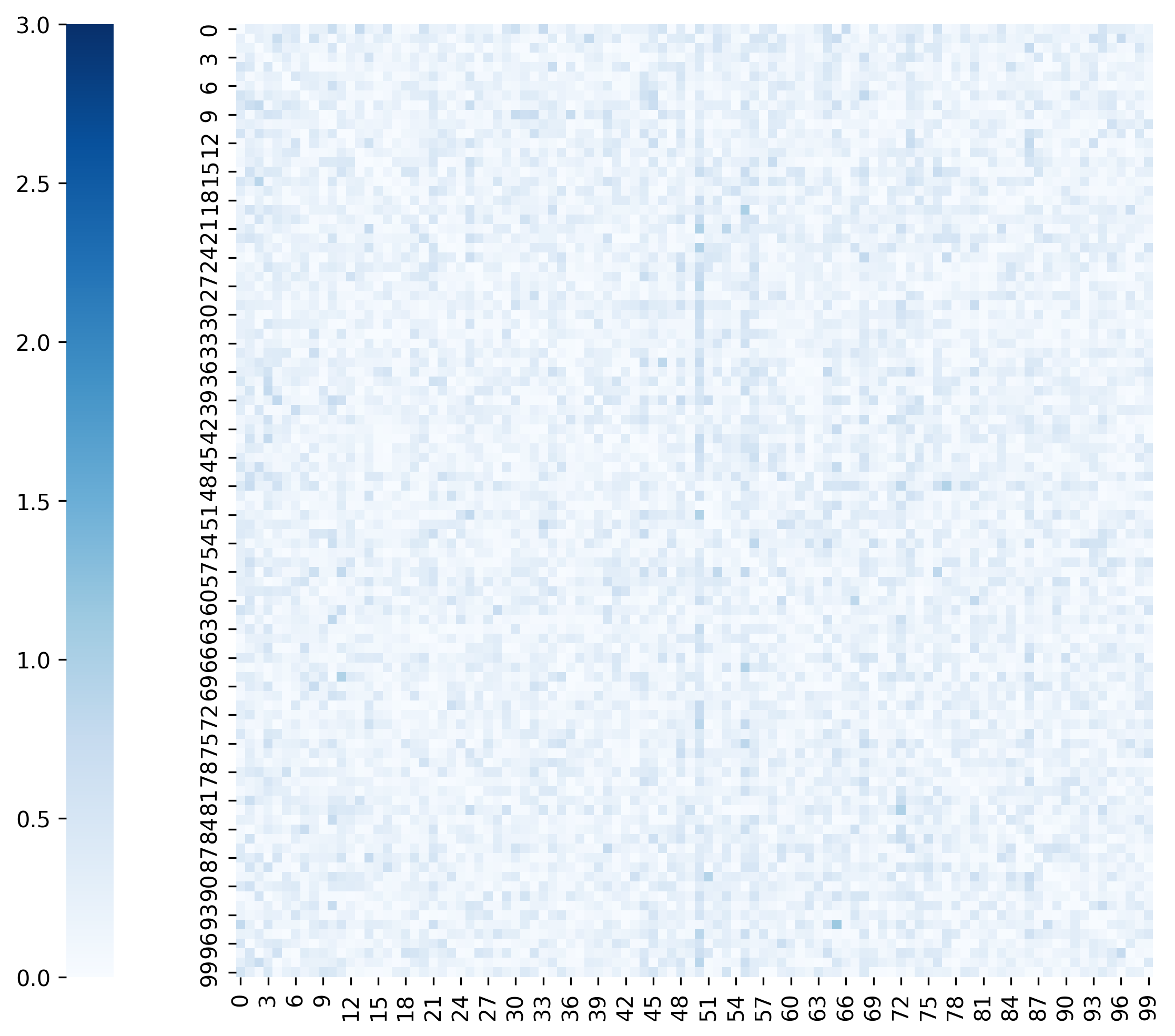}
  \subcaption{PKT \cite{passalis2018probabilistic} \\ Mean: 0.19, Max: 1.14}
\end{subfigure}
\hfill
\begin{subfigure}[b]{0.23\textwidth}
  \centering
  \includegraphics[width=\linewidth]{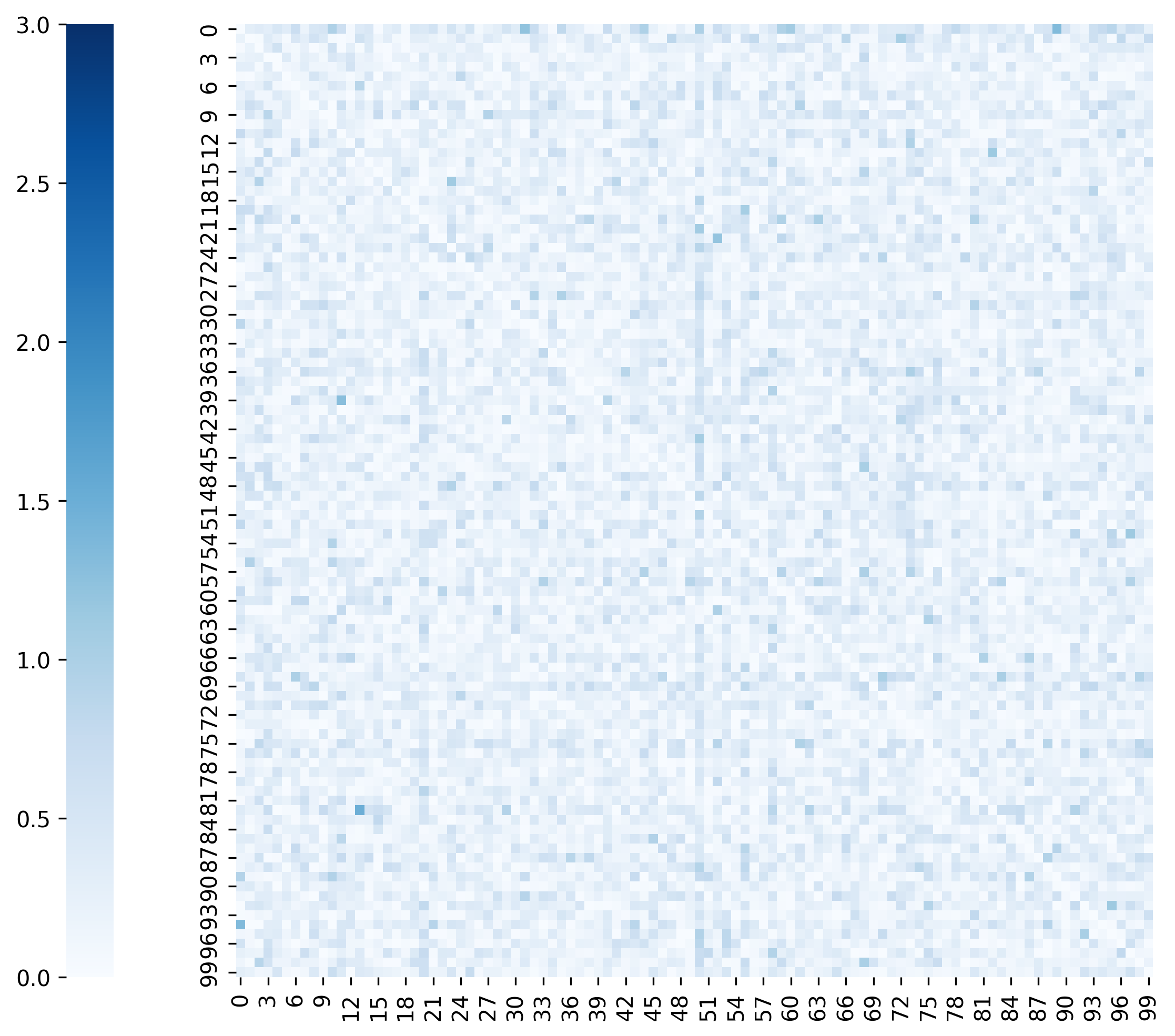}
  \subcaption{AB \cite{heo2019knowledge} \\ Mean: 0.24, Max: 1.49}
\end{subfigure}
\hfill
\begin{subfigure}[b]{0.23\textwidth}
  \centering
  \includegraphics[width=\linewidth]{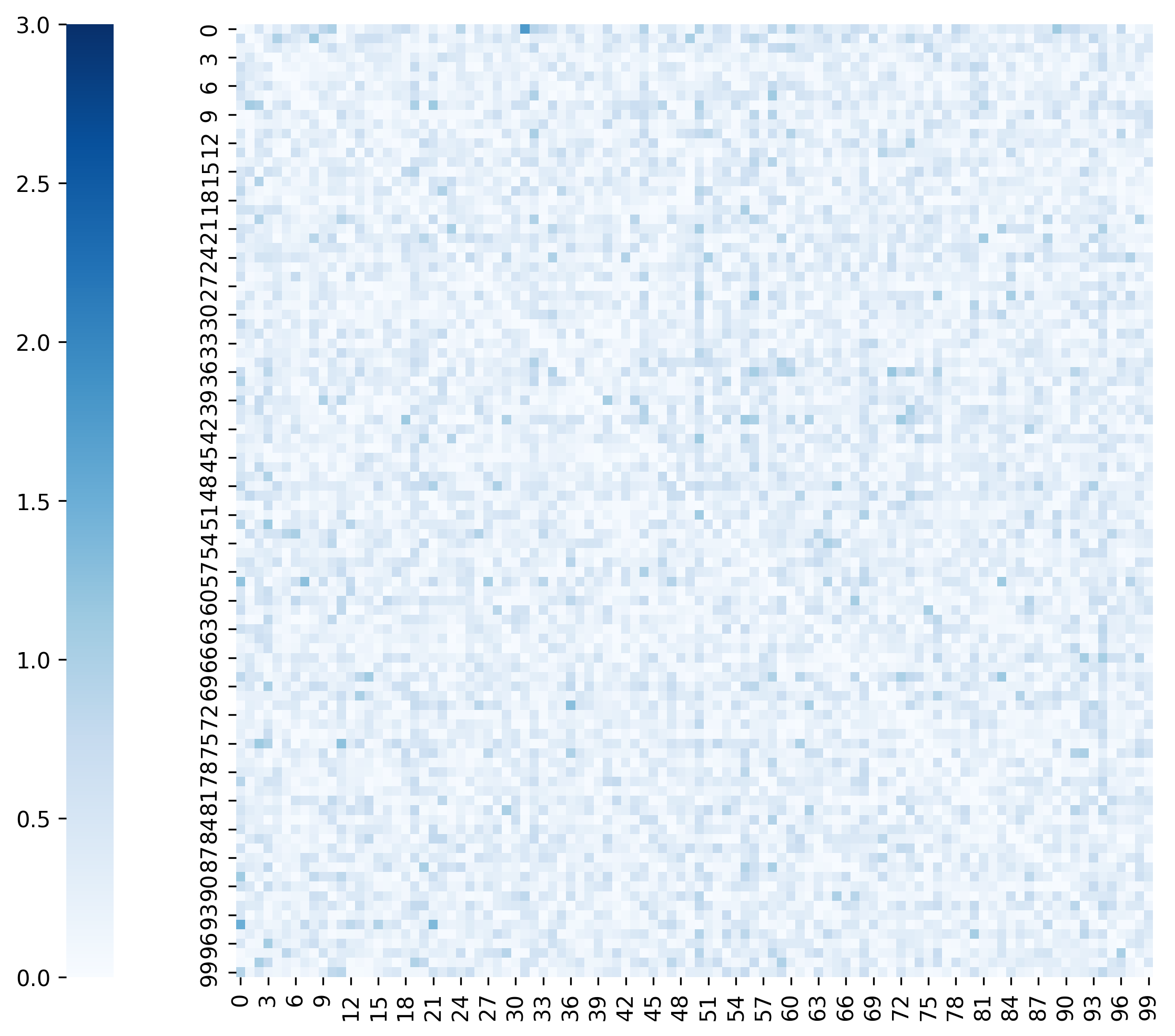}
  \subcaption{FT \cite{kim2018paraphrasing} \\ Mean: 0.26, Max: 1.76}
\end{subfigure}
\hfill
\begin{subfigure}[b]{0.23\textwidth}
  \centering
  \includegraphics[width=\linewidth]{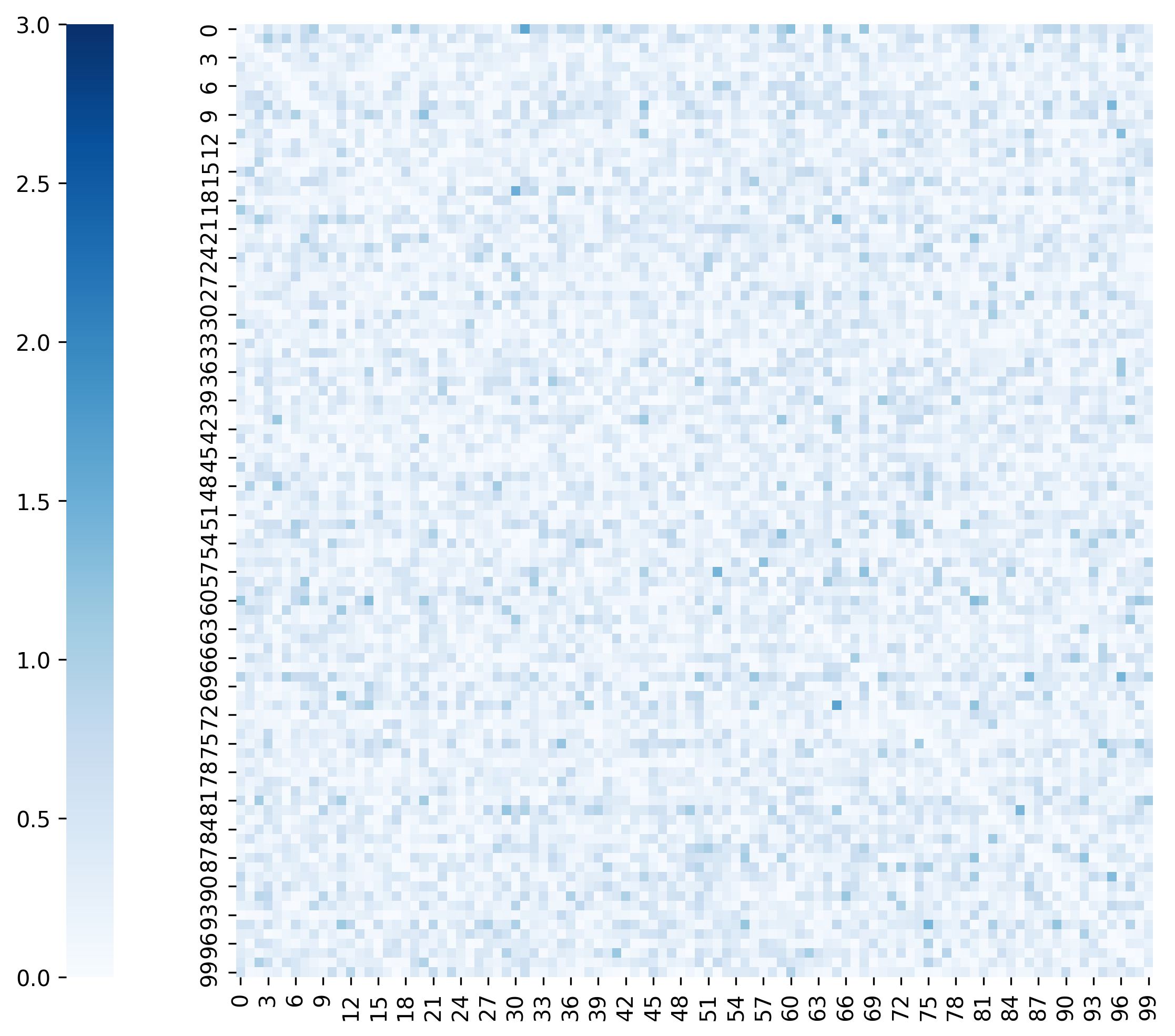}
  \subcaption{NST \cite{huang2017like} \\ Mean: 0.28, Max: 1.65}
\end{subfigure}
\begin{subfigure}[b]{0.23\textwidth}
  \centering
  \includegraphics[width=\linewidth]{figures/correlation_CRD_wrn_40_1_wrn_40_2.png}
  \subcaption{CRD \cite{tian2022crd} \\ Mean: 0.23, Max: 1.56}
\end{subfigure}
\hfill
\begin{subfigure}[b]{0.23\textwidth}
  \centering
  \includegraphics[width=\linewidth]{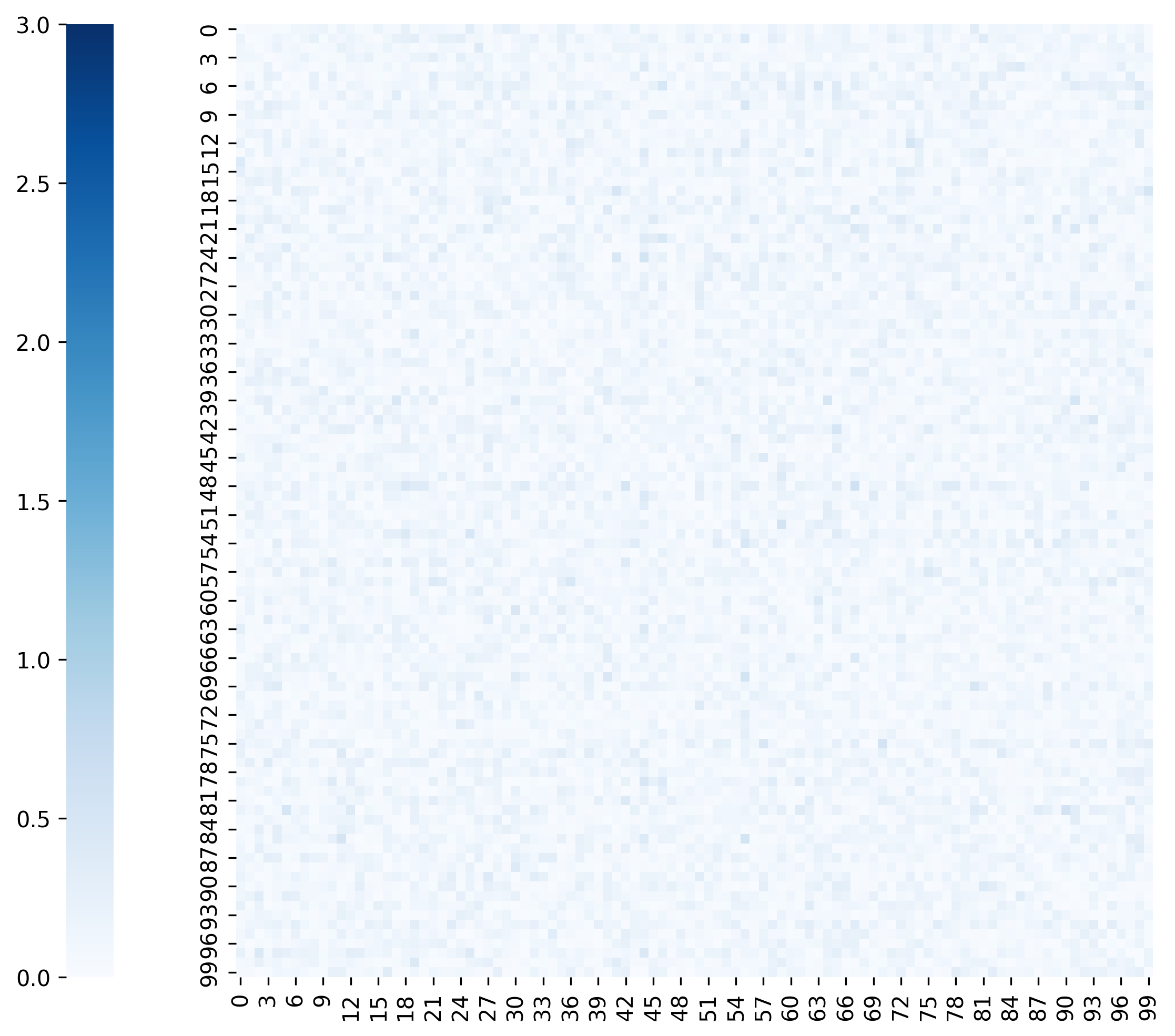}
  \subcaption{CRD+KD \cite{tian2022crd} \\ Mean: 0.10, Max: 0.57}
\end{subfigure}
\hfill
\begin{subfigure}[b]{0.23\textwidth}
  \centering
  \includegraphics[width=\linewidth]{figures/correlation_DCD_wrn_40_1_wrn_40_2.png}
  \subcaption{\model \\ Mean: 0.18, Max: 0.99}
\end{subfigure}
\hfill
\begin{subfigure}[b]{0.23\textwidth}
  \centering
  \includegraphics[width=\linewidth]{figures/correlation_DCD+KD_wrn_40_1_wrn_40_2.png}
  \subcaption{\model+KD \\ Mean: 0.07, Max: 0.55}
\end{subfigure}
\caption{\textbf{Logit correlation analysis.} Matrix of the average logit difference between teacher and student outputs on CIFAR-100 (lower values indicate higher similarity). Results are based on a WRN-40-2 teacher and a WRN-40-1 student.}
\label{fig:correlation_appendix}
\end{figure*}

\Cref{fig:correlation_appendix} compares the correlation matrix differences between teacher (WRN-40-2) and student (WRN-40-1) logits on CIFAR-100. Our method achieves better alignment of correlation structures compared to models trained without distillation or with alternative methods.

\subsection{t-SNE}
\label{app:extended_results_tsne}

\begin{figure*}[!p]
\centering
\begin{subfigure}[b]{0.16\textwidth}
  \centering
  \includegraphics[width=\linewidth]{figures/tsne_TEACHER_wrn_40_1_wrn_40_2_20classes.png}
  \subcaption{Teacher}
\end{subfigure}
\hfill
\begin{subfigure}[b]{0.16\textwidth}
  \centering
  \includegraphics[width=\linewidth]{figures/tsne_VANILLA_wrn_40_1_wrn_40_2_20classes.png}
  \subcaption{Vanilla}
\end{subfigure}
\hfill
\begin{subfigure}[b]{0.16\textwidth}
  \centering
  \includegraphics[width=\linewidth]{figures/tsne_KD_wrn_40_1_wrn_40_2_20classes.png}
  \subcaption{KD \cite{hinton2015distilling}}
\end{subfigure}
\hfill
\begin{subfigure}[b]{0.16\textwidth}
  \centering
  \includegraphics[width=\linewidth]{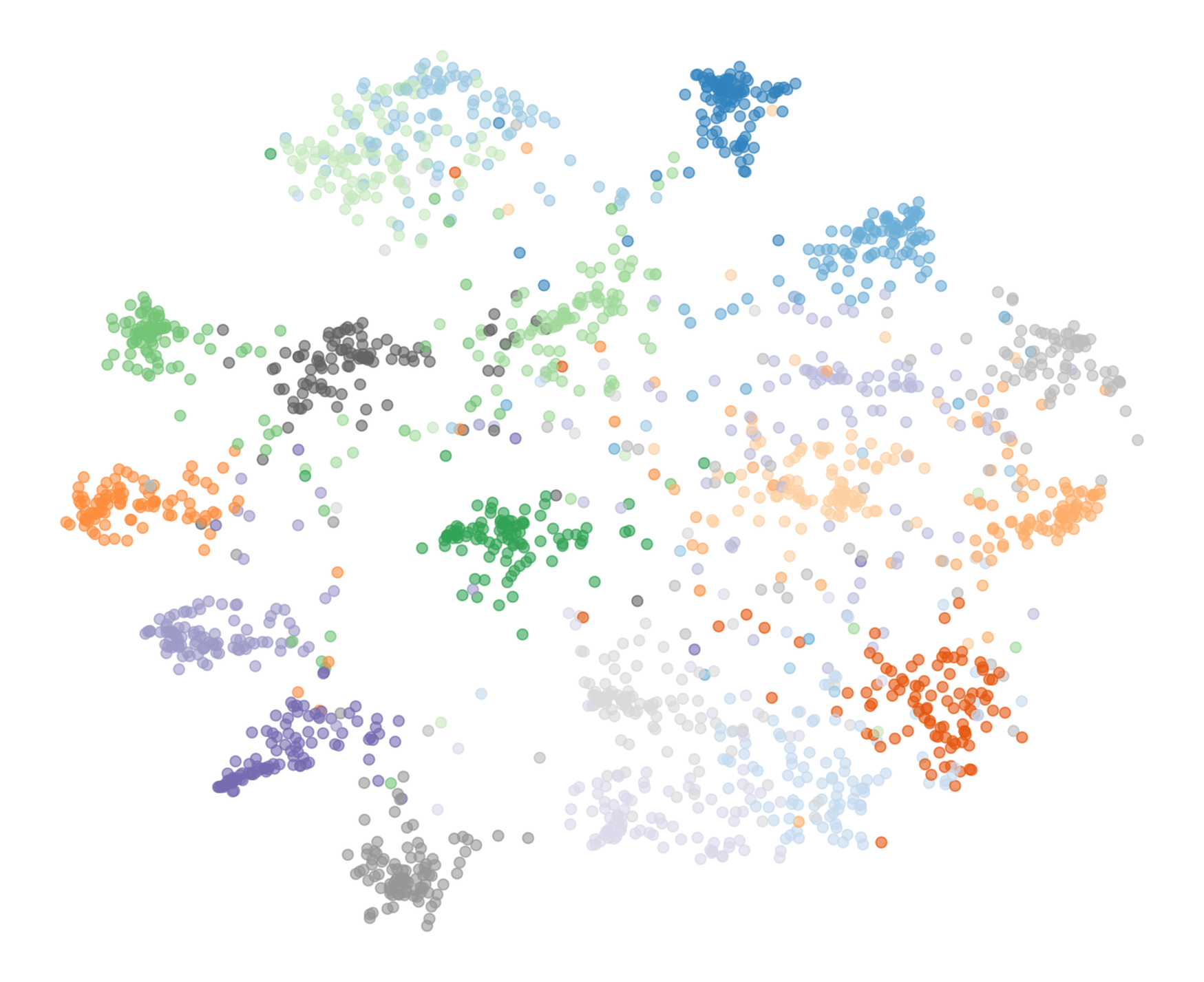}
  \subcaption{FitNets \cite{romero2014fitnets}}
\end{subfigure}
\hfill
\begin{subfigure}[b]{0.16\textwidth}
  \centering
  \includegraphics[width=\linewidth]{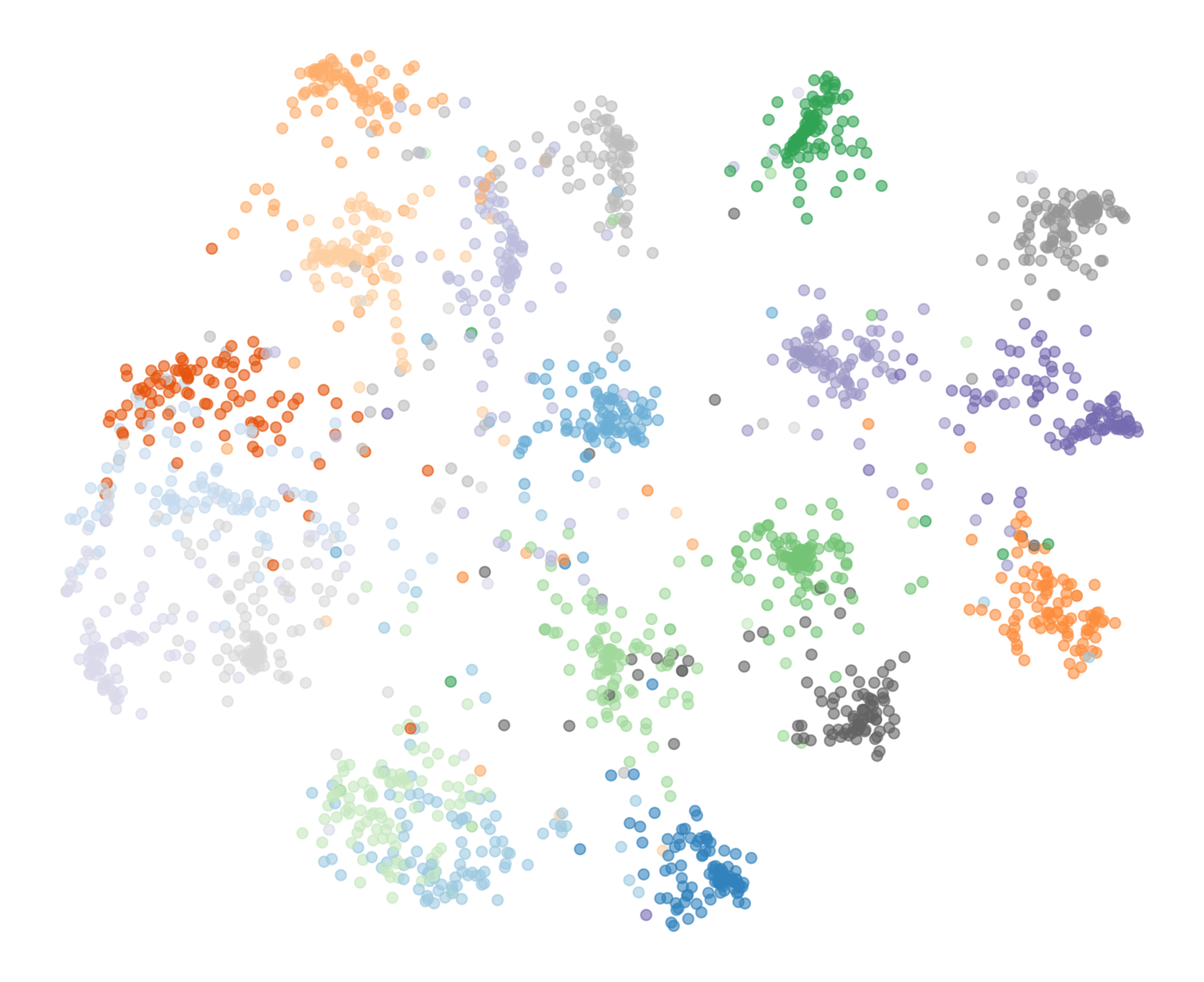}
  \subcaption{AT \cite{zagoruyko2016paying}}
\end{subfigure}
\hfill
\begin{subfigure}[b]{0.16\textwidth}
  \centering
  \includegraphics[width=\linewidth]{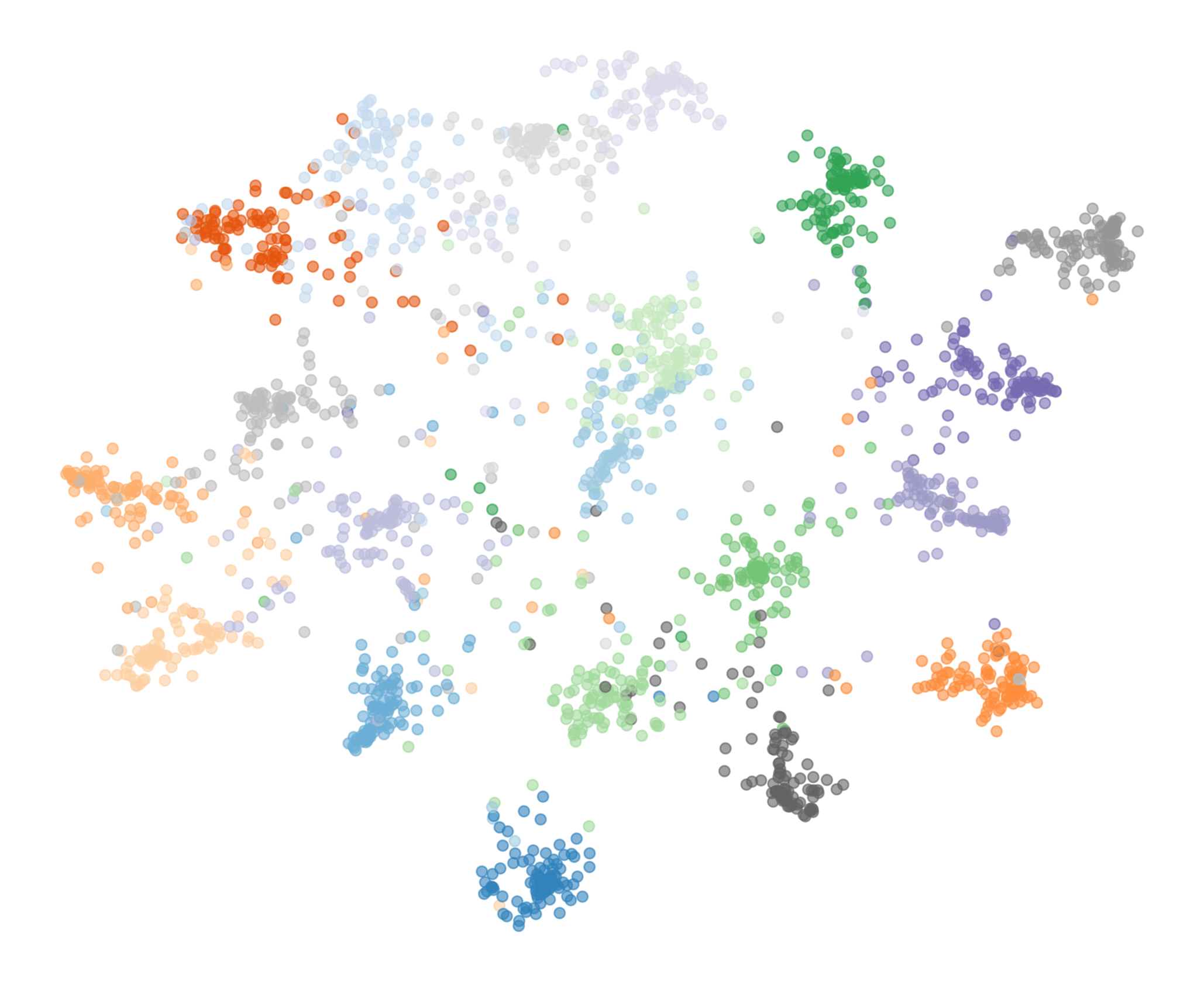}
  \subcaption{SP \cite{tung2019similarity}}
\end{subfigure}
\begin{subfigure}[b]{0.16\textwidth}
  \centering
  \includegraphics[width=\linewidth]{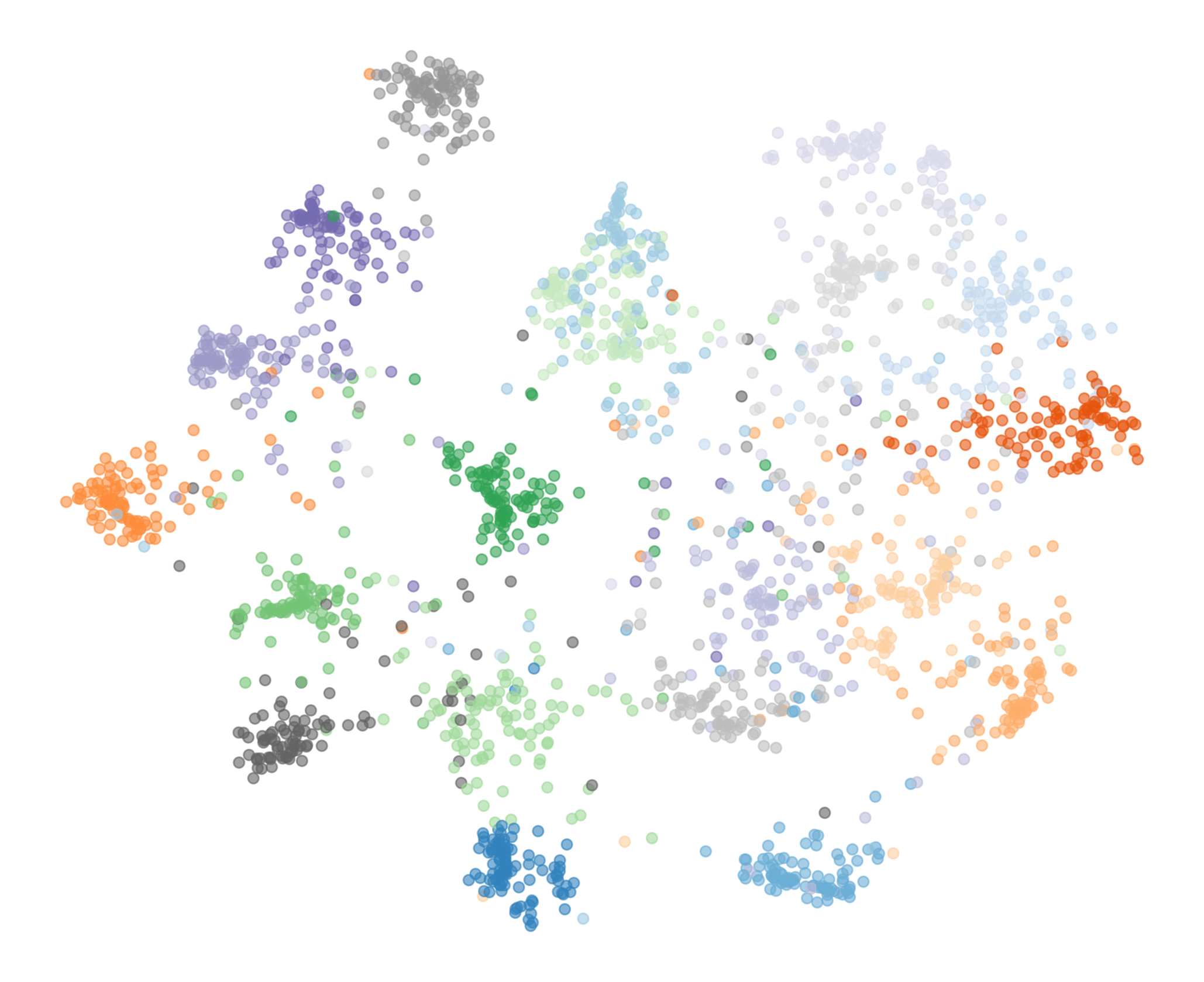}
  \subcaption{CC \cite{peng2019correlation}}
\end{subfigure}
\hfill
\begin{subfigure}[b]{0.16\textwidth}
  \centering
  \includegraphics[width=\linewidth]{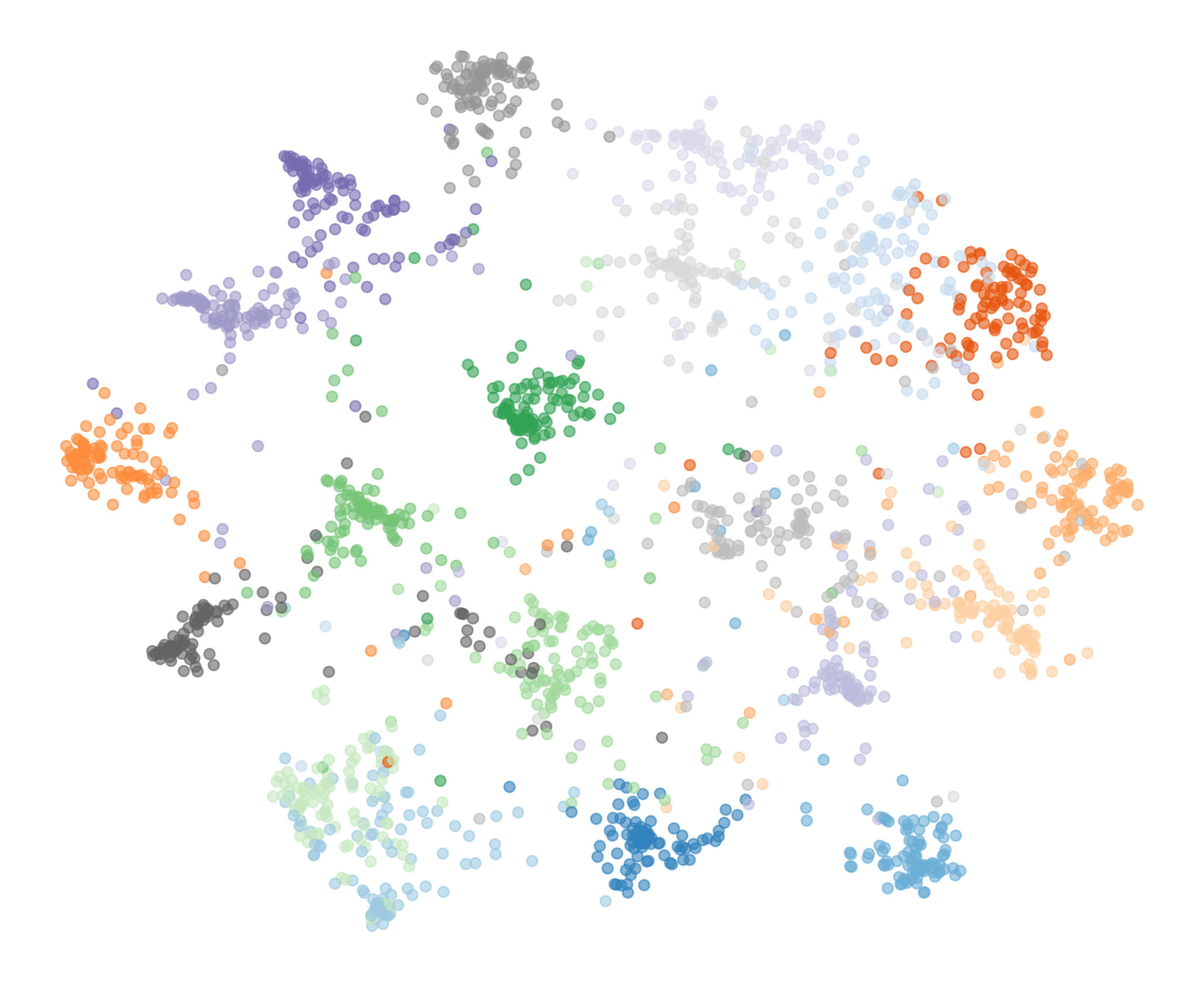}
  \subcaption{VID \cite{ahn2019variational}}
\end{subfigure}
\hfill
\begin{subfigure}[b]{0.16\textwidth}
  \centering
  \includegraphics[width=\linewidth]{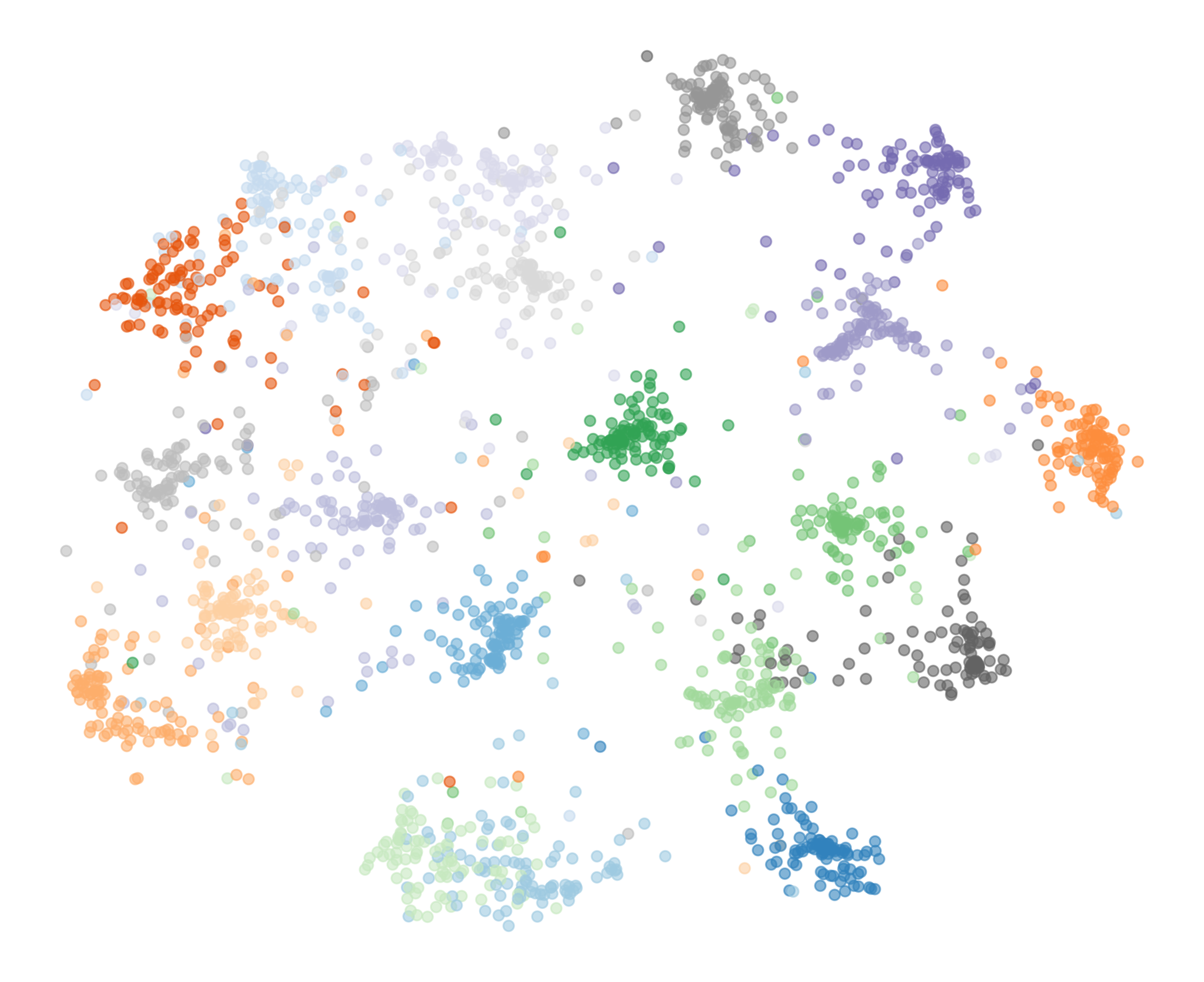}
  \subcaption{RKD \cite{park2019relational}}
\end{subfigure}
\hfill
\begin{subfigure}[b]{0.16\textwidth}
  \centering
  \includegraphics[width=\linewidth]{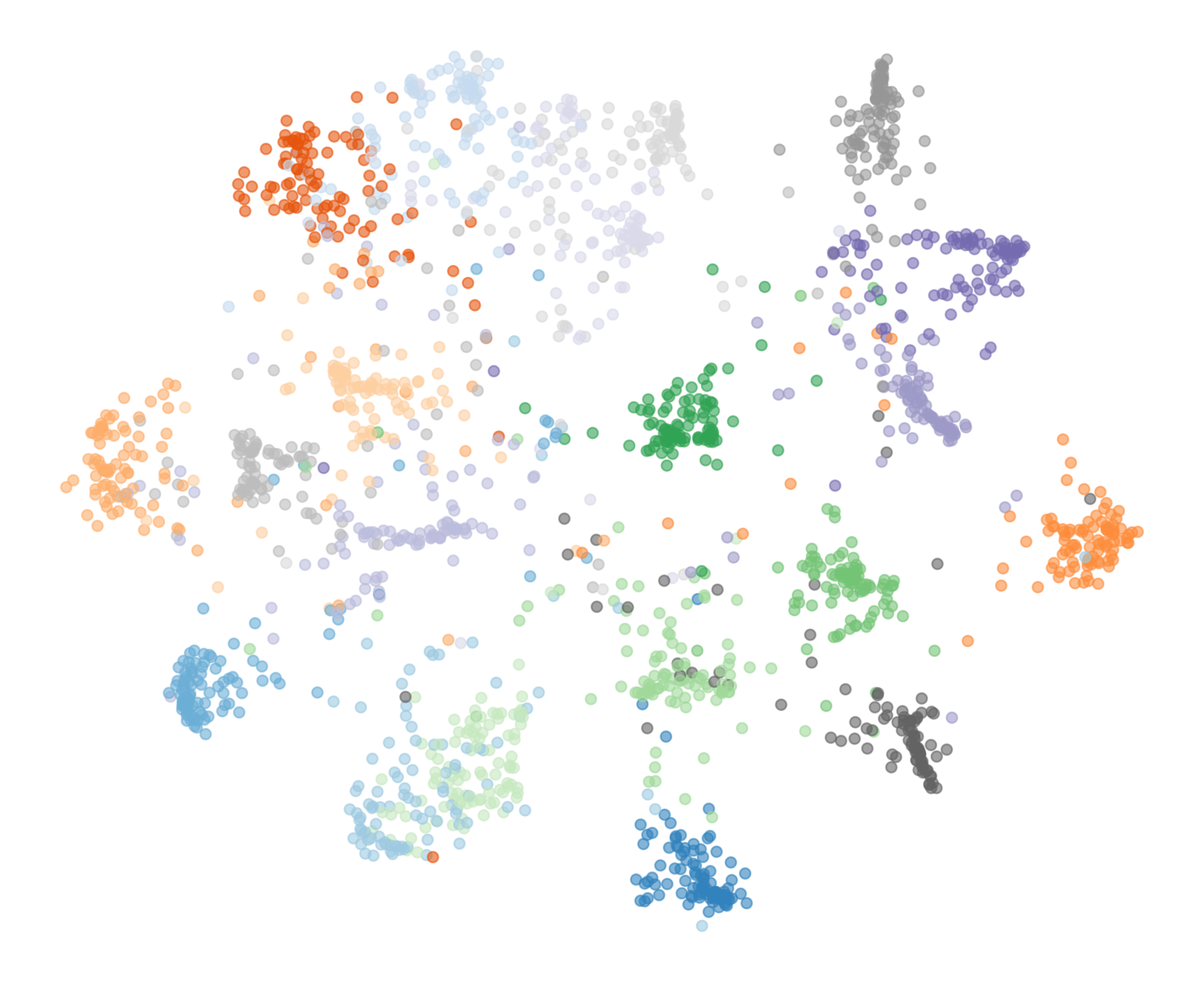}
  \subcaption{PKT \cite{passalis2018probabilistic}}
\end{subfigure}
\hfill
\begin{subfigure}[b]{0.16\textwidth}
  \centering
  \includegraphics[width=\linewidth]{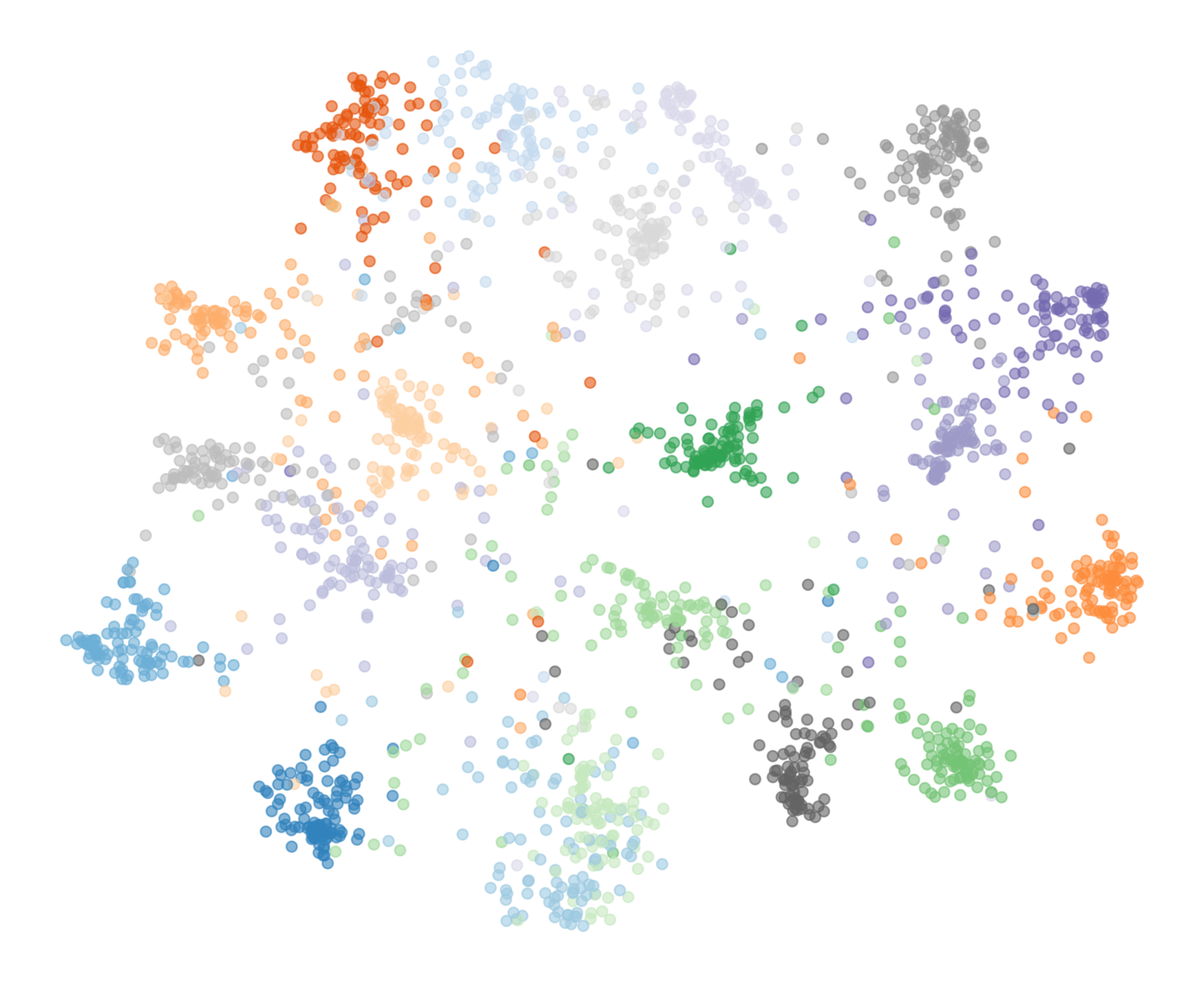}
  \subcaption{AB \cite{heo2019knowledge}}
\end{subfigure}
\hfill
\begin{subfigure}[b]{0.16\textwidth}
  \centering
  \includegraphics[width=\linewidth]{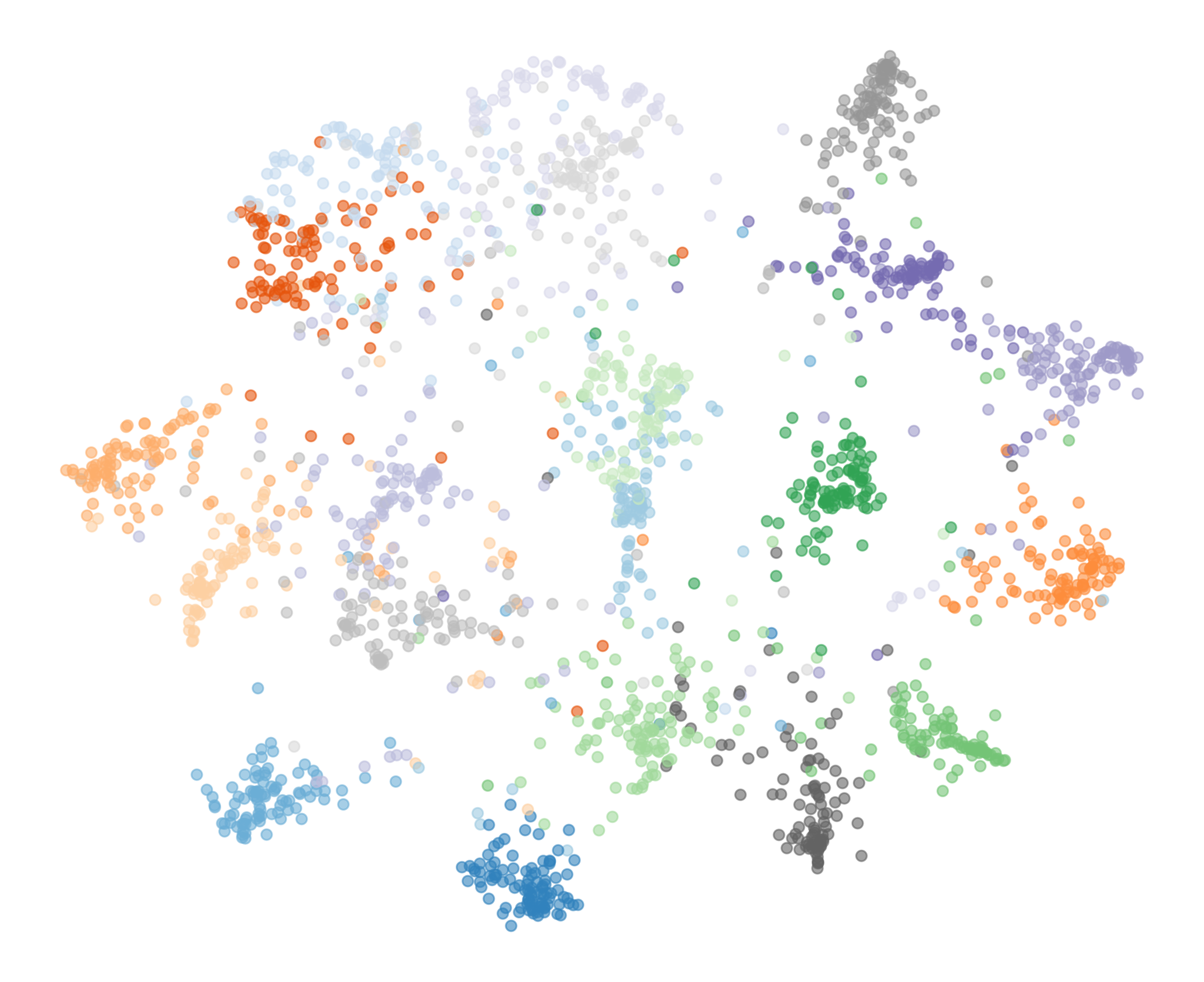}
  \subcaption{FT \cite{kim2018paraphrasing}}
\end{subfigure}
\begin{subfigure}[b]{0.16\textwidth}
  \centering
  \includegraphics[width=\linewidth]{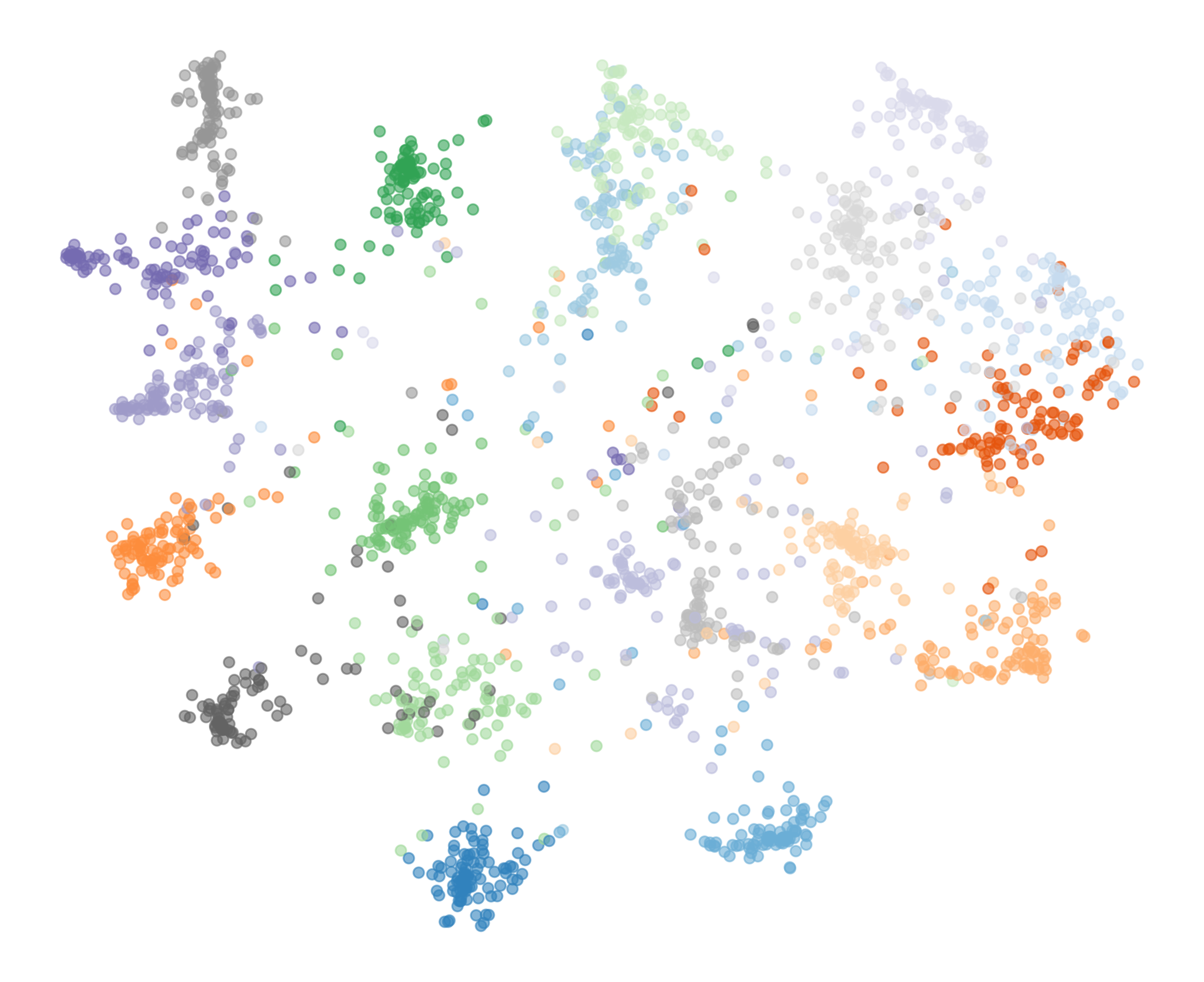}
  \subcaption{NST \cite{huang2017like}}
\end{subfigure}
\hfill
\begin{subfigure}[b]{0.16\textwidth}
  \centering
  \includegraphics[width=\linewidth]{figures/tsne_CRD_wrn_40_1_wrn_40_2_20classes.png}
  \subcaption{CRD \cite{tian2022crd}}
\end{subfigure}
\hfill
\begin{subfigure}[b]{0.16\textwidth}
  \centering
  \includegraphics[width=\linewidth]{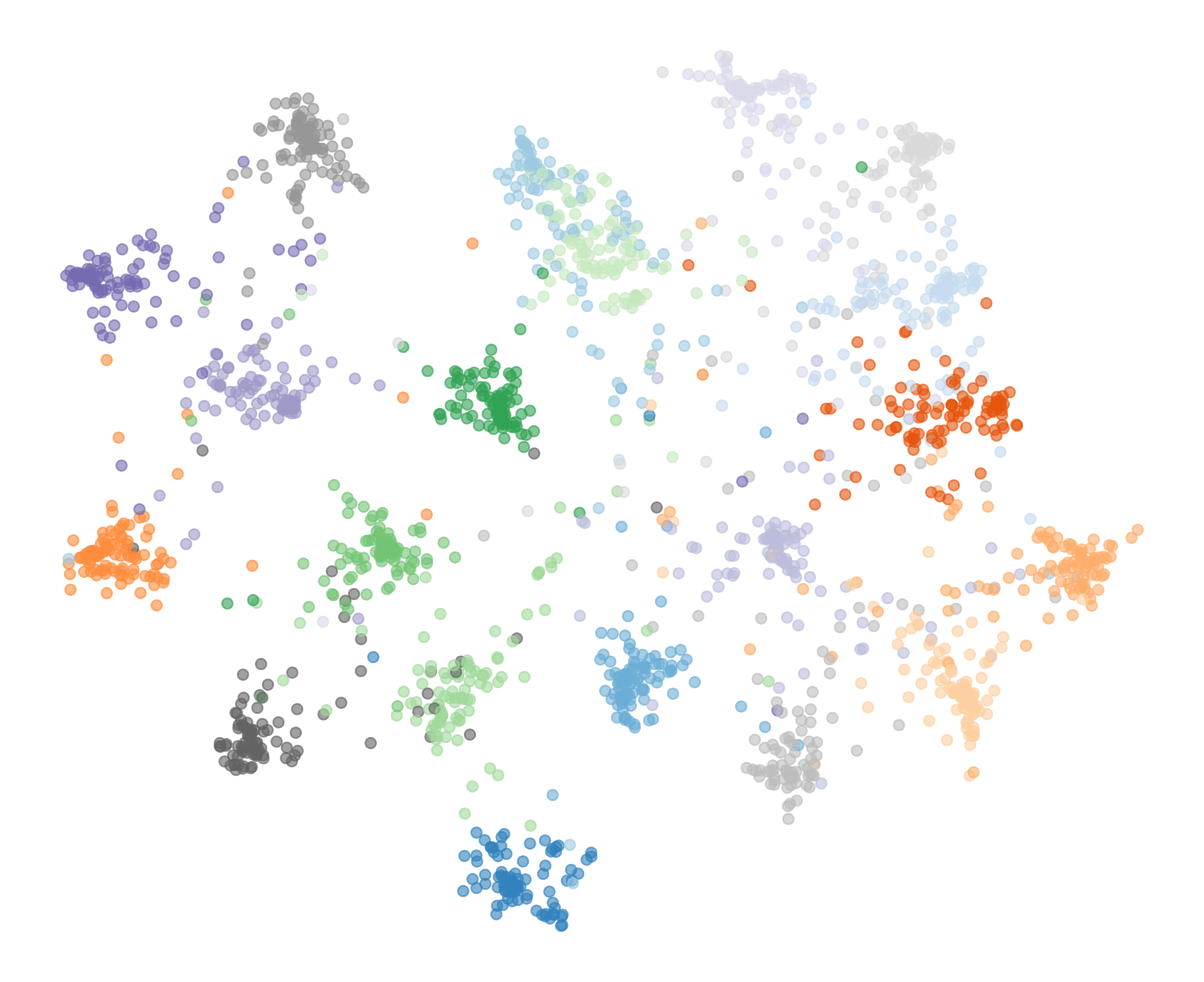}
  \subcaption{CRD+KD \cite{tian2022crd}}
\end{subfigure}
\hfill
\begin{subfigure}[b]{0.16\textwidth}
  \centering
  \includegraphics[width=\linewidth]{figures/tsne_DCD_wrn_40_1_wrn_40_2_20classes.png}
  \subcaption{\model}
\end{subfigure}
\hfill
\begin{subfigure}[b]{0.16\textwidth}
  \centering
  \includegraphics[width=\linewidth]{figures/tsne_DCD+KD_wrn_40_1_wrn_40_2_20classes.png}
  \subcaption{\model+KD}
\end{subfigure}
\caption{\textbf{t-SNE visualization of embedding spaces.} Comparison of feature distributions from the teacher and student networks for the first 20 classes of CIFAR-100. We use a WRN-40-2 teacher and a WRN-40-1 student for the visualization.}
\label{fig:tsne1_appendix}
\end{figure*}

\begin{figure*}[!p]
\centering
\begin{subfigure}[b]{0.16\textwidth}
  \centering
  \includegraphics[width=\linewidth]{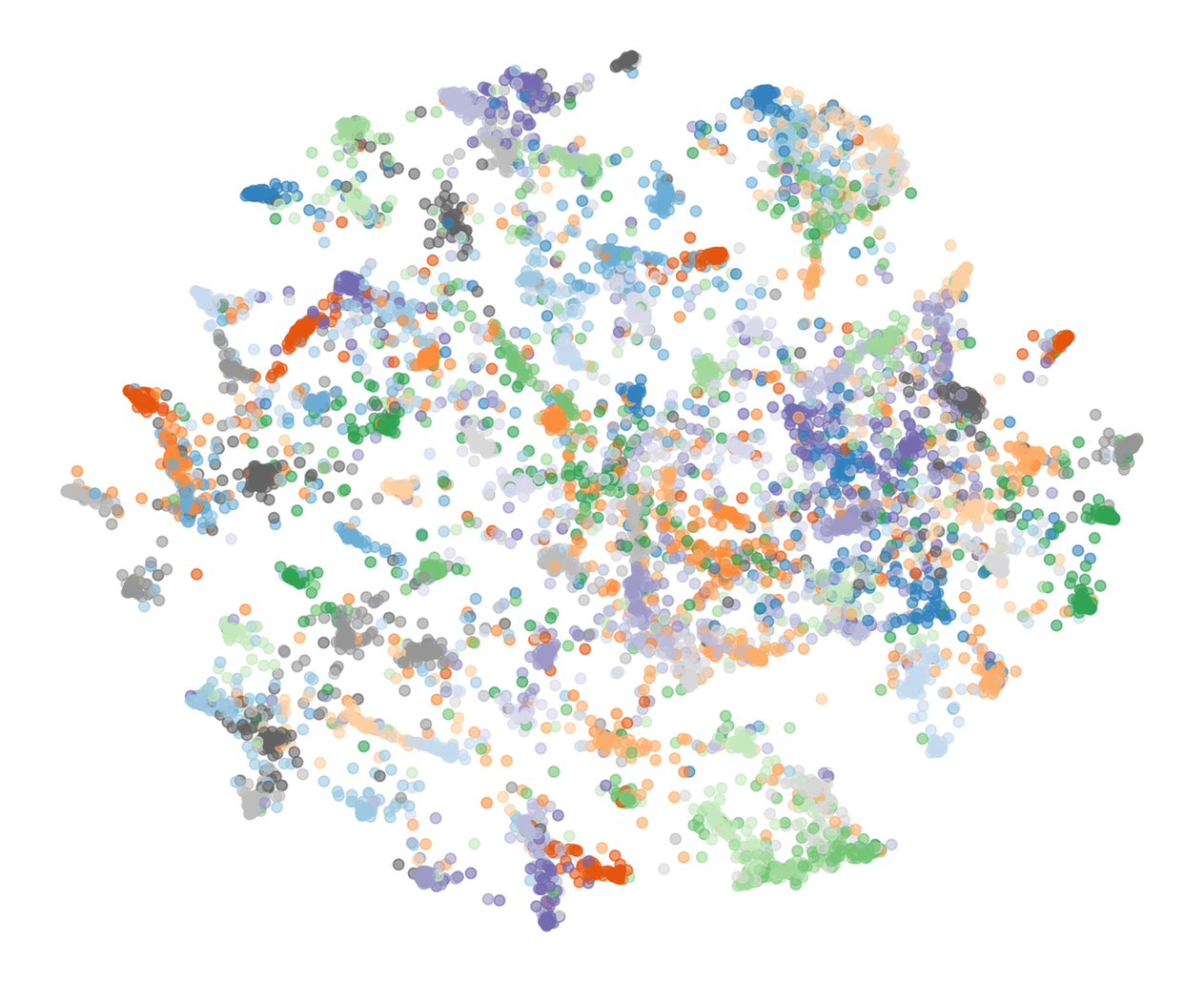}
  \subcaption{Teacher}
\end{subfigure}
\hfill
\begin{subfigure}[b]{0.16\textwidth}
  \centering
  \includegraphics[width=\linewidth]{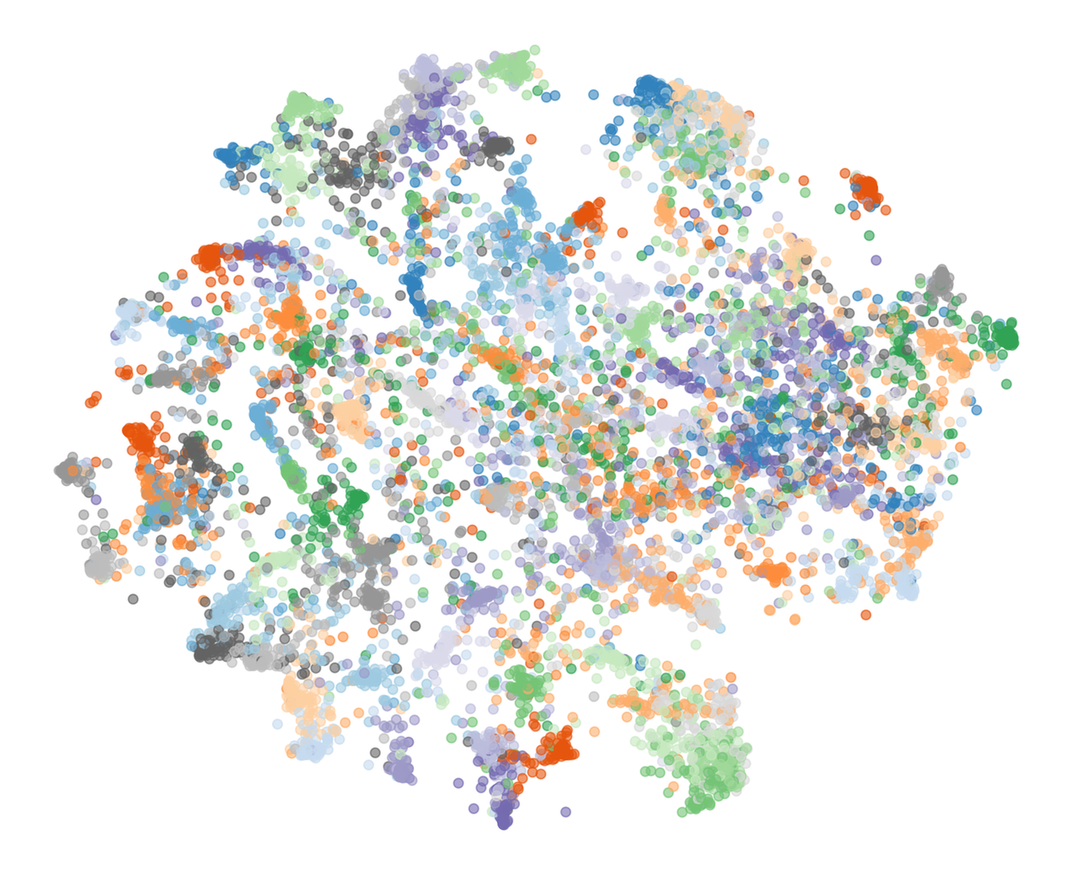}
  \subcaption{Vanilla}
\end{subfigure}
\hfill
\begin{subfigure}[b]{0.16\textwidth}
  \centering
  \includegraphics[width=\linewidth]{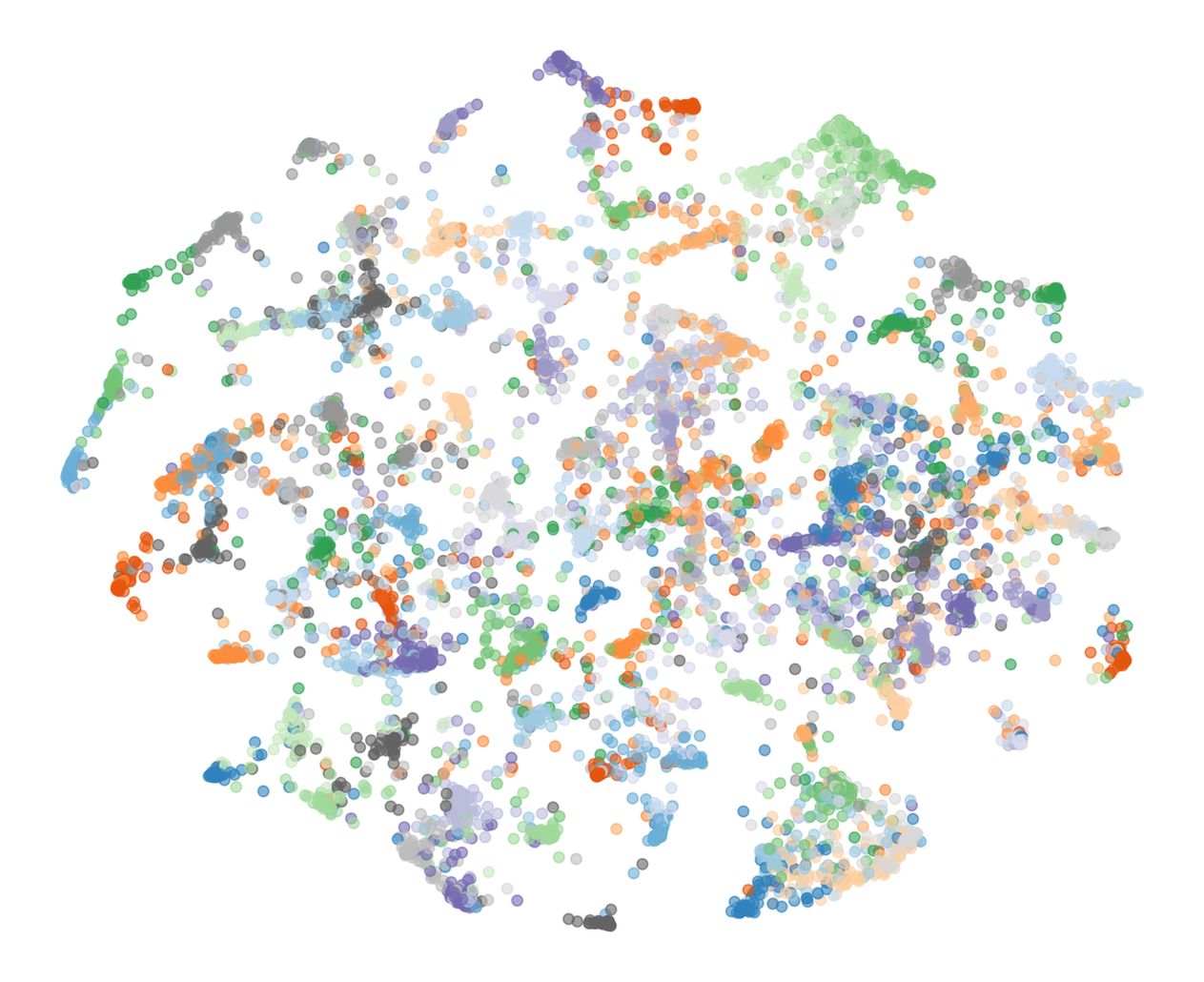}
  \subcaption{KD \cite{hinton2015distilling}}
\end{subfigure}
\hfill
\begin{subfigure}[b]{0.16\textwidth}
  \centering
  \includegraphics[width=\linewidth]{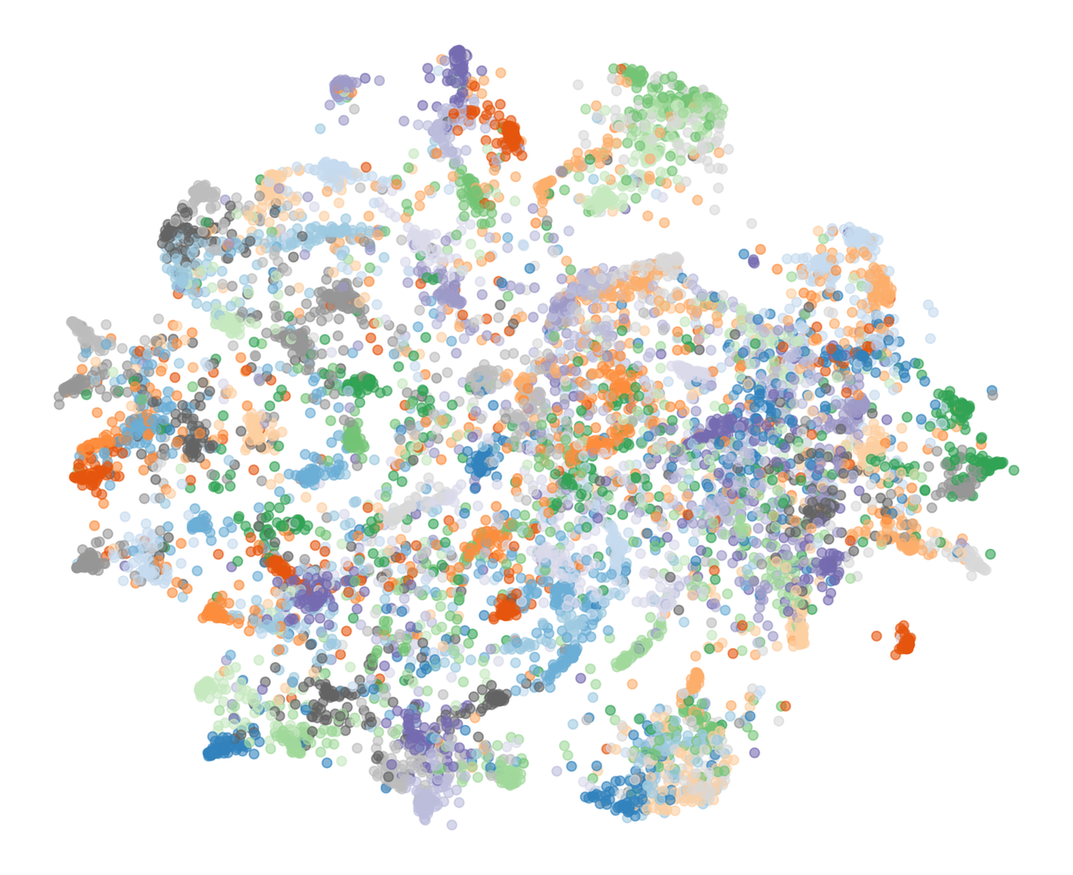}
  \subcaption{FitNets \cite{romero2014fitnets}}
\end{subfigure}
\hfill
\begin{subfigure}[b]{0.16\textwidth}
  \centering
  \includegraphics[width=\linewidth]{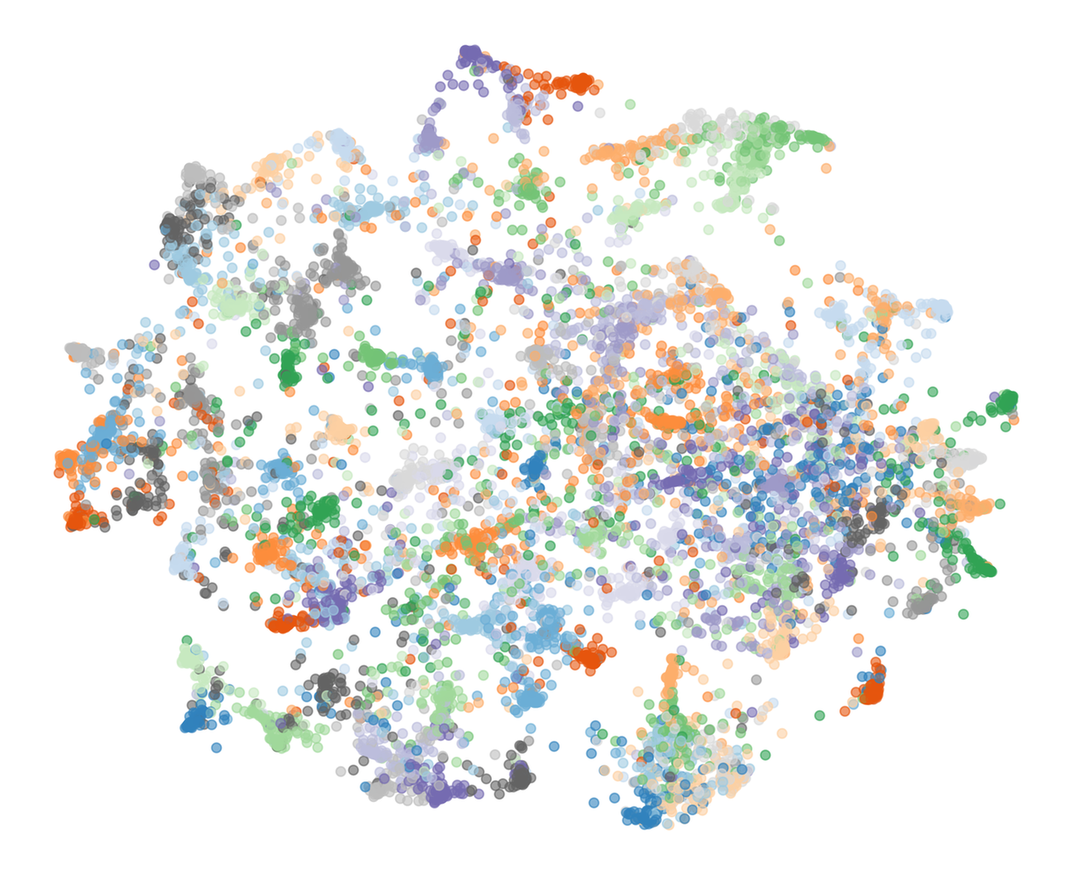}
  \subcaption{AT \cite{zagoruyko2016paying}}
\end{subfigure}
\hfill
\begin{subfigure}[b]{0.16\textwidth}
  \centering
  \includegraphics[width=\linewidth]{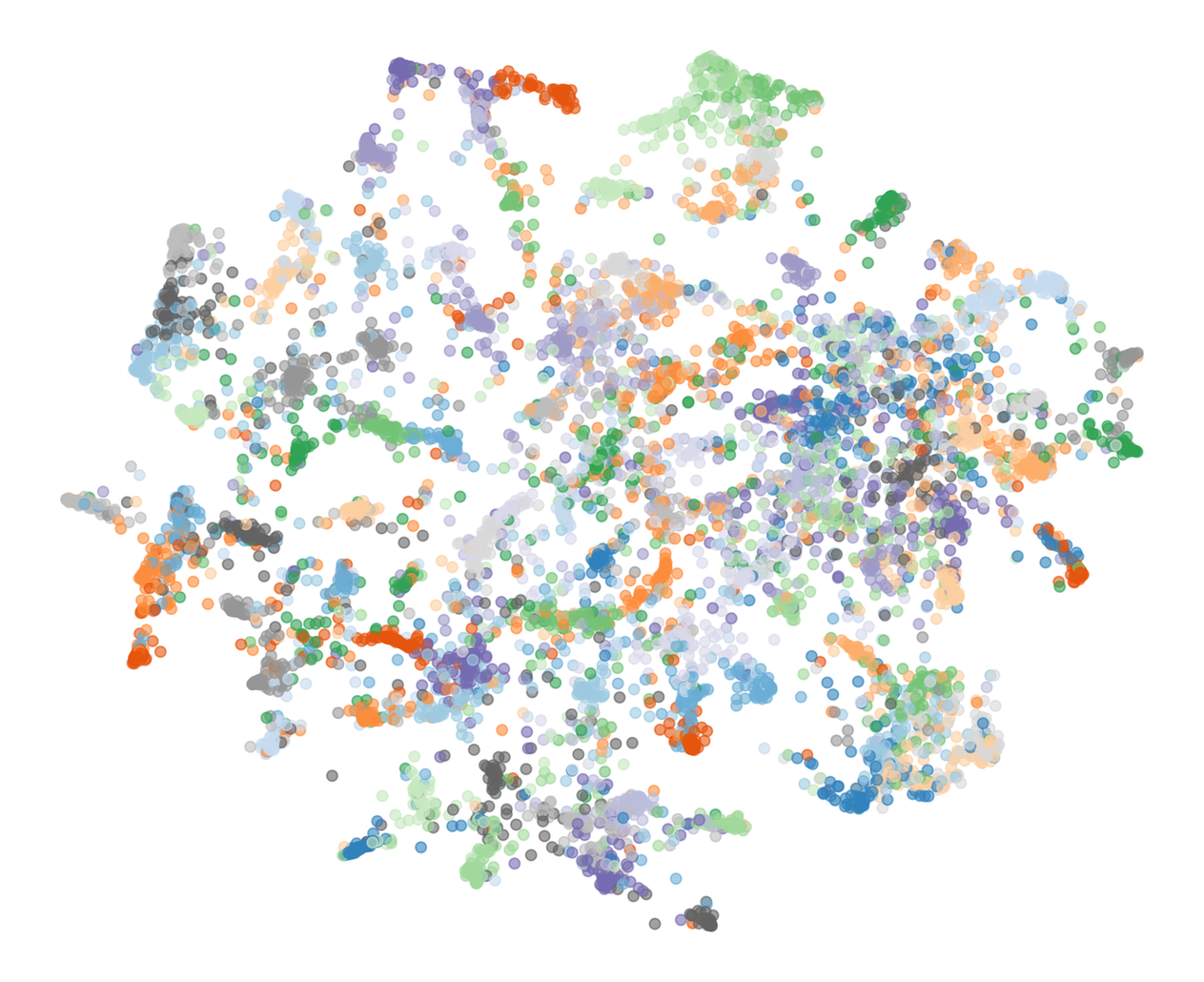}
  \subcaption{SP \cite{tung2019similarity}}
\end{subfigure}
\begin{subfigure}[b]{0.16\textwidth}
  \centering
  \includegraphics[width=\linewidth]{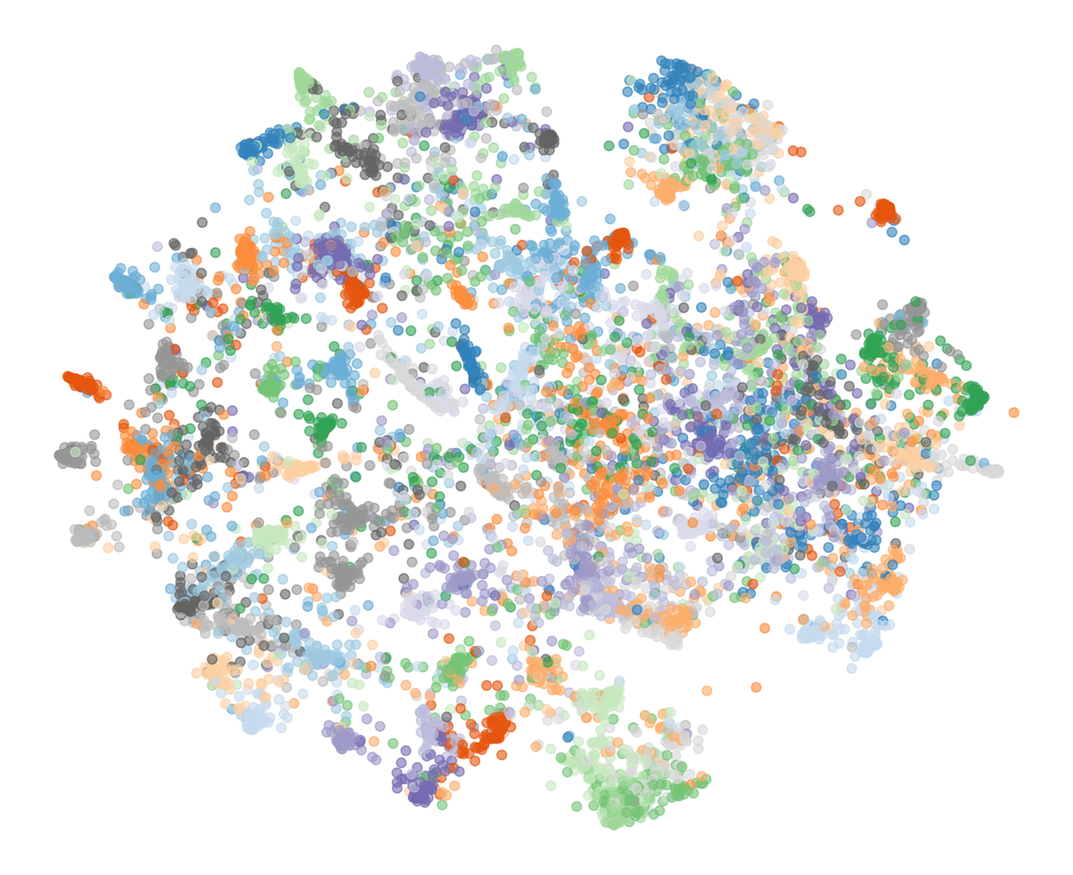}
  \subcaption{CC \cite{peng2019correlation}}
\end{subfigure}
\hfill
\begin{subfigure}[b]{0.16\textwidth}
  \centering
  \includegraphics[width=\linewidth]{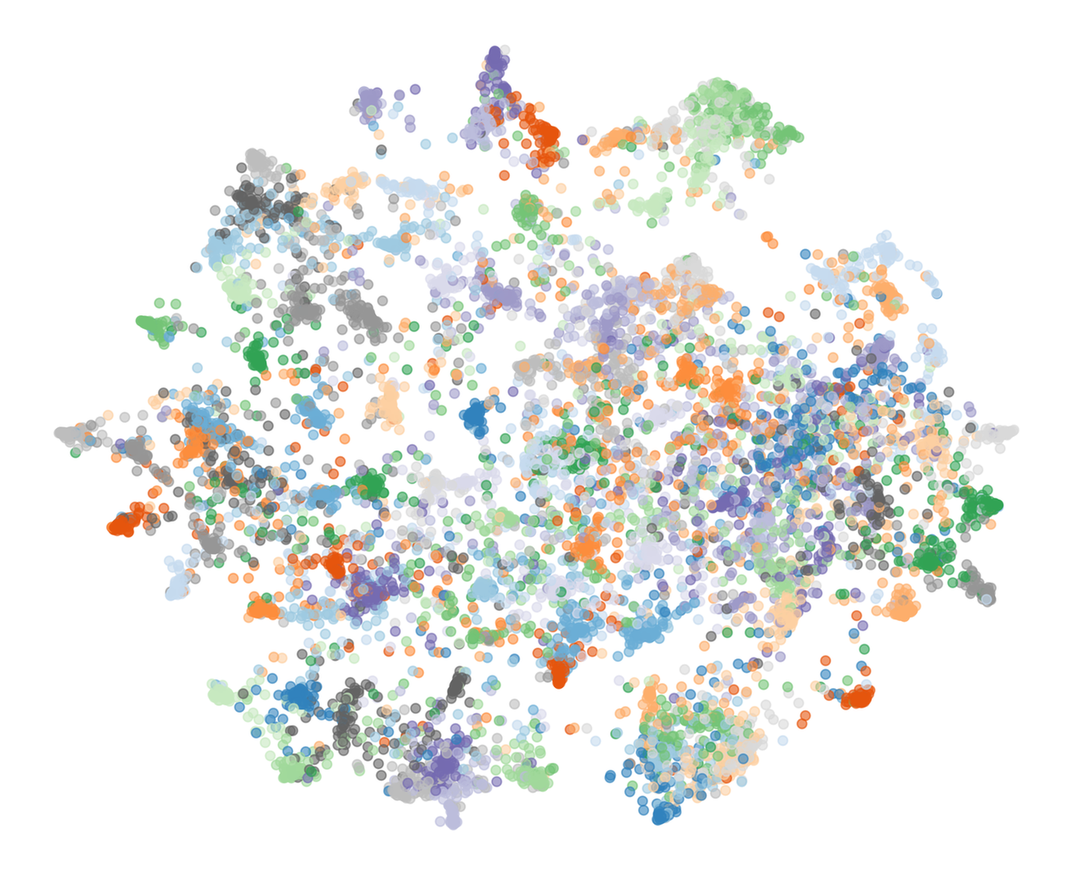}
  \subcaption{VID \cite{ahn2019variational}}
\end{subfigure}
\hfill
\begin{subfigure}[b]{0.16\textwidth}
  \centering
  \includegraphics[width=\linewidth]{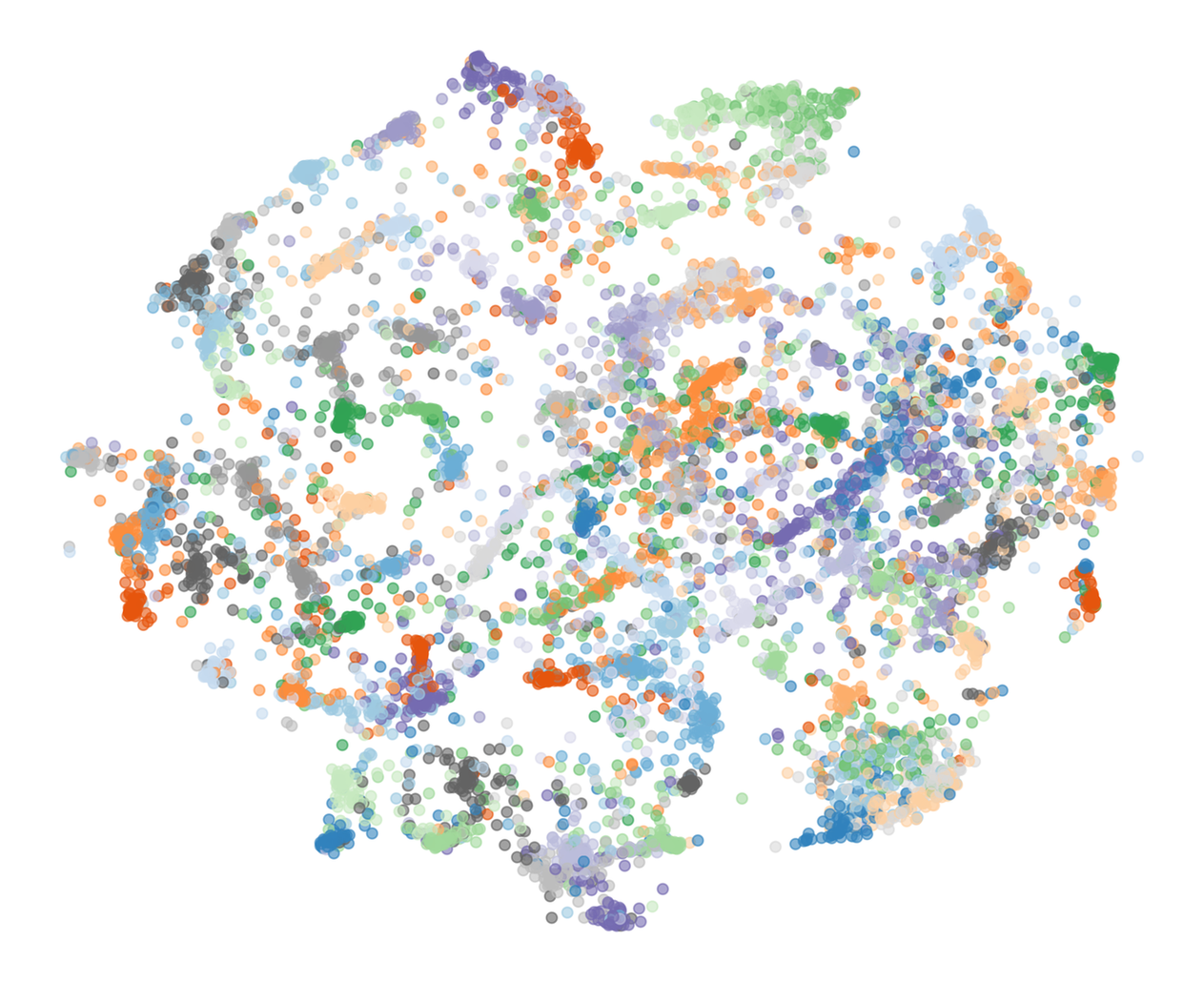}
  \subcaption{RKD \cite{park2019relational}}
\end{subfigure}
\hfill
\begin{subfigure}[b]{0.16\textwidth}
  \centering
  \includegraphics[width=\linewidth]{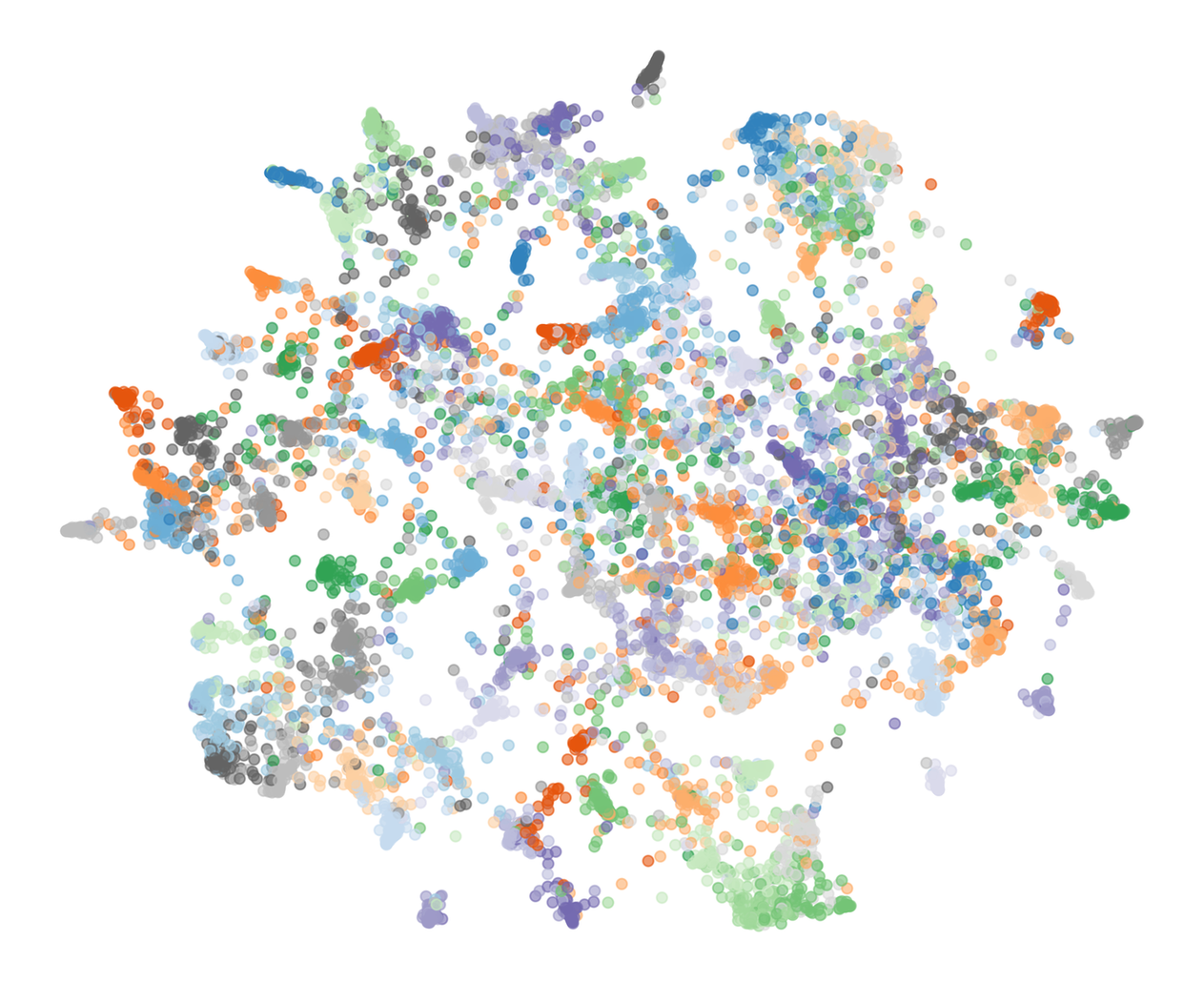}
  \subcaption{PKT \cite{passalis2018probabilistic}}
\end{subfigure}
\hfill
\begin{subfigure}[b]{0.16\textwidth}
  \centering
  \includegraphics[width=\linewidth]{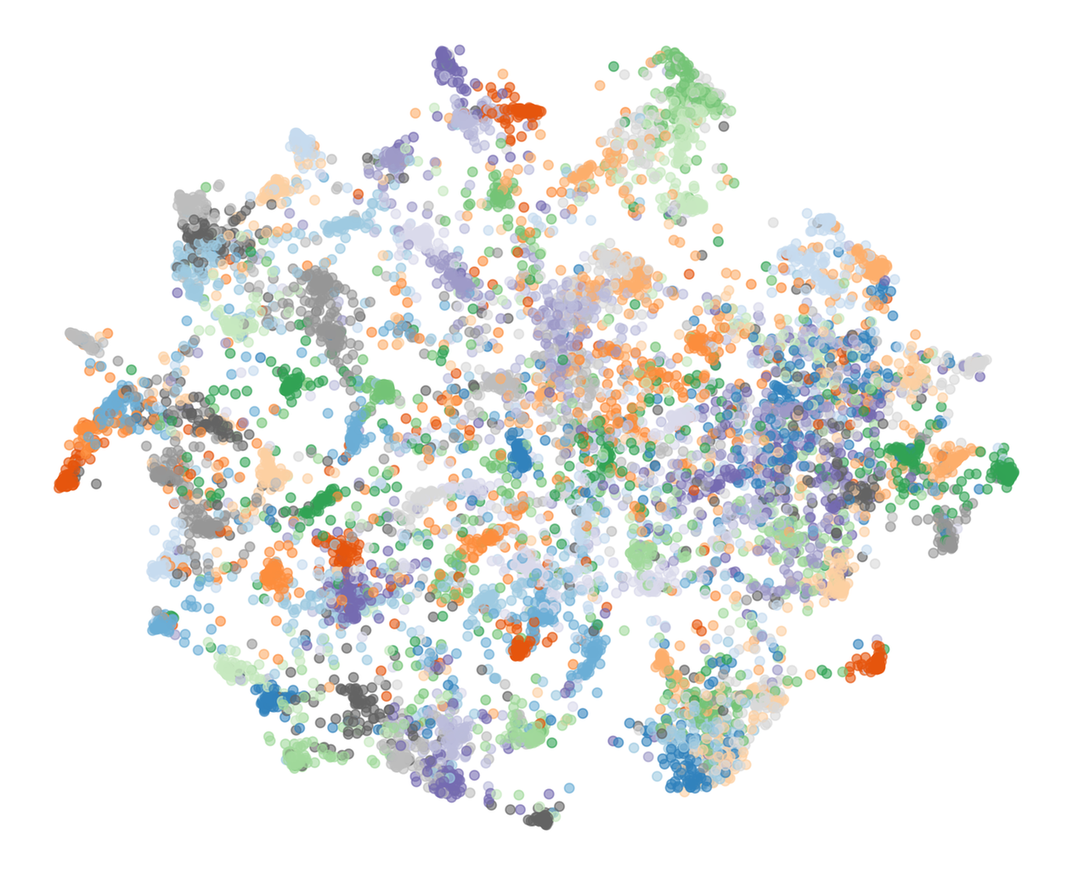}
  \subcaption{AB \cite{heo2019knowledge}}
\end{subfigure}
\hfill
\begin{subfigure}[b]{0.16\textwidth}
  \centering
  \includegraphics[width=\linewidth]{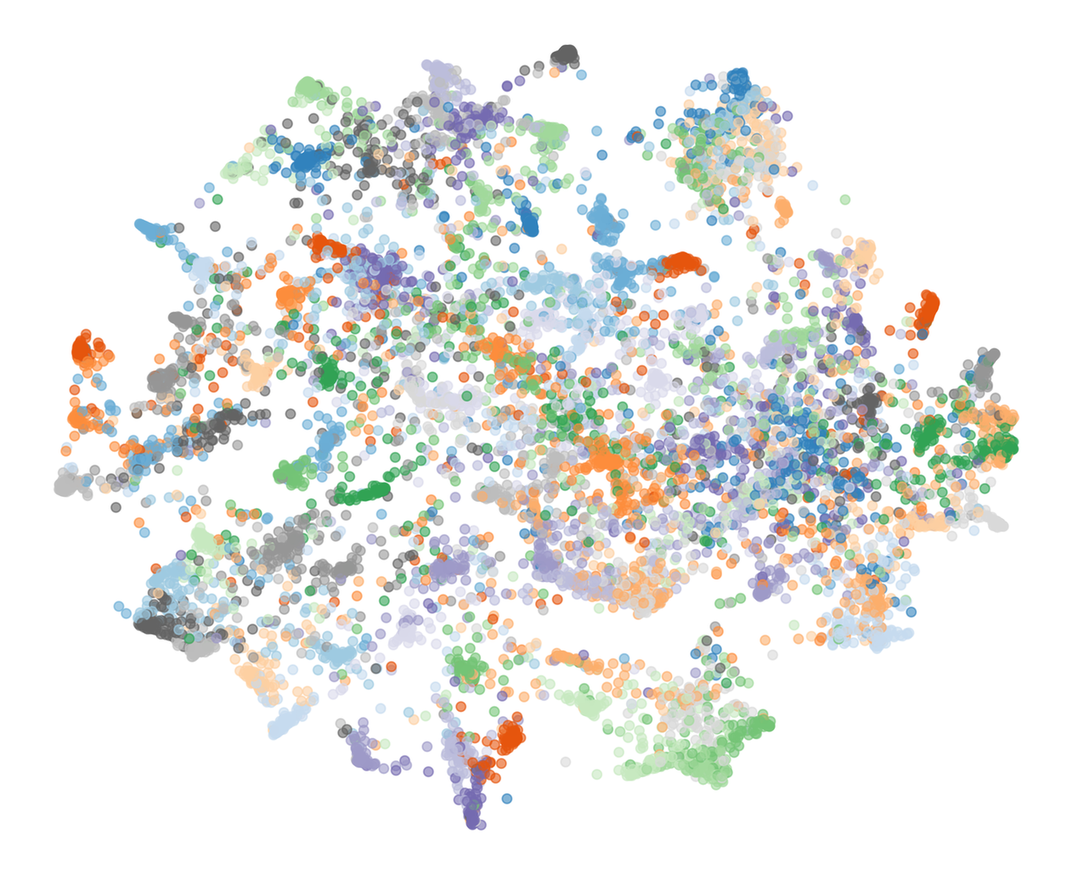}
  \subcaption{FT \cite{kim2018paraphrasing}}
\end{subfigure}
\begin{subfigure}[b]{0.16\textwidth}
  \centering
  \includegraphics[width=\linewidth]{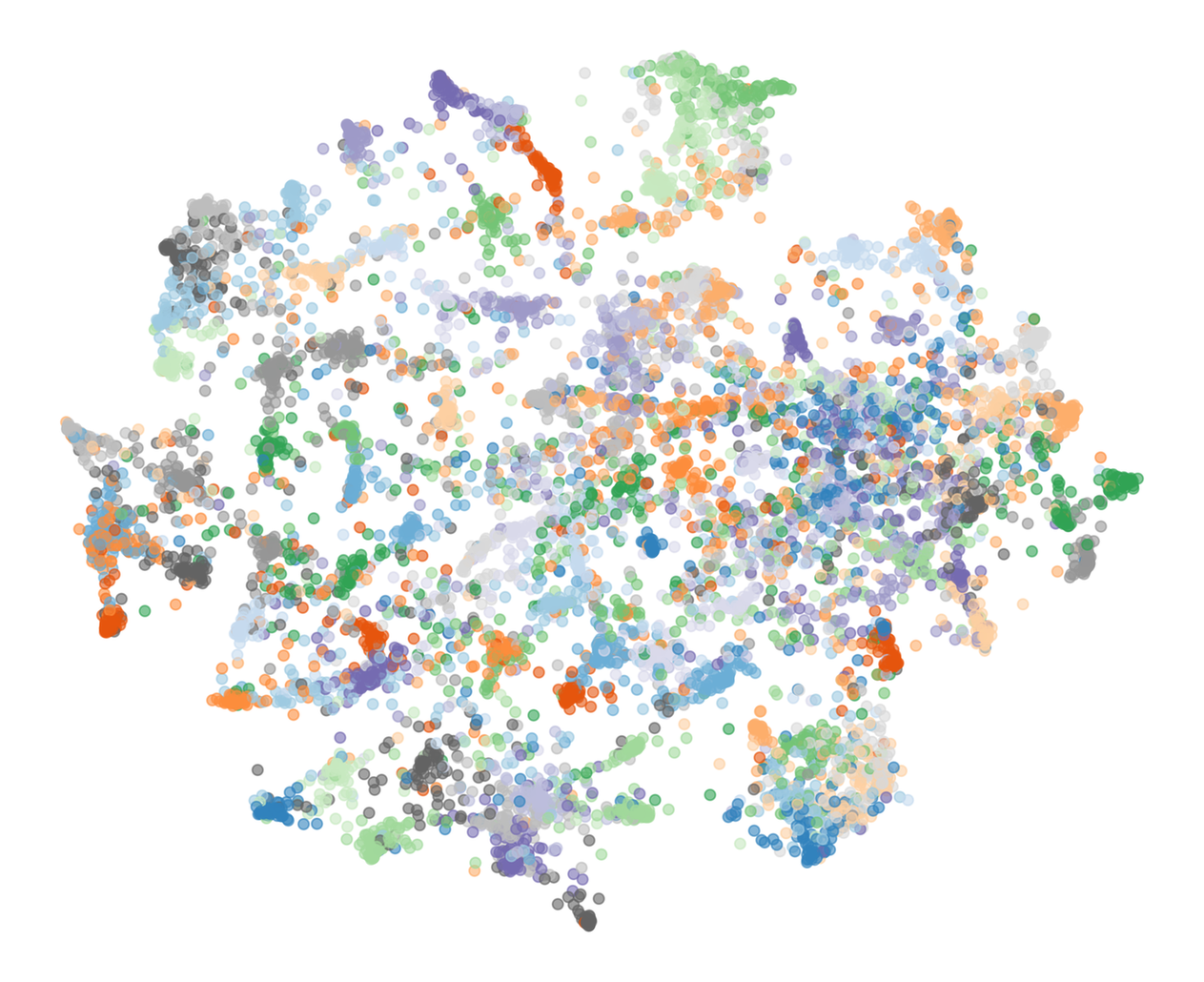}
  \subcaption{NST \cite{huang2017like}}
\end{subfigure}
\hfill
\begin{subfigure}[b]{0.16\textwidth}
  \centering
  \includegraphics[width=\linewidth]{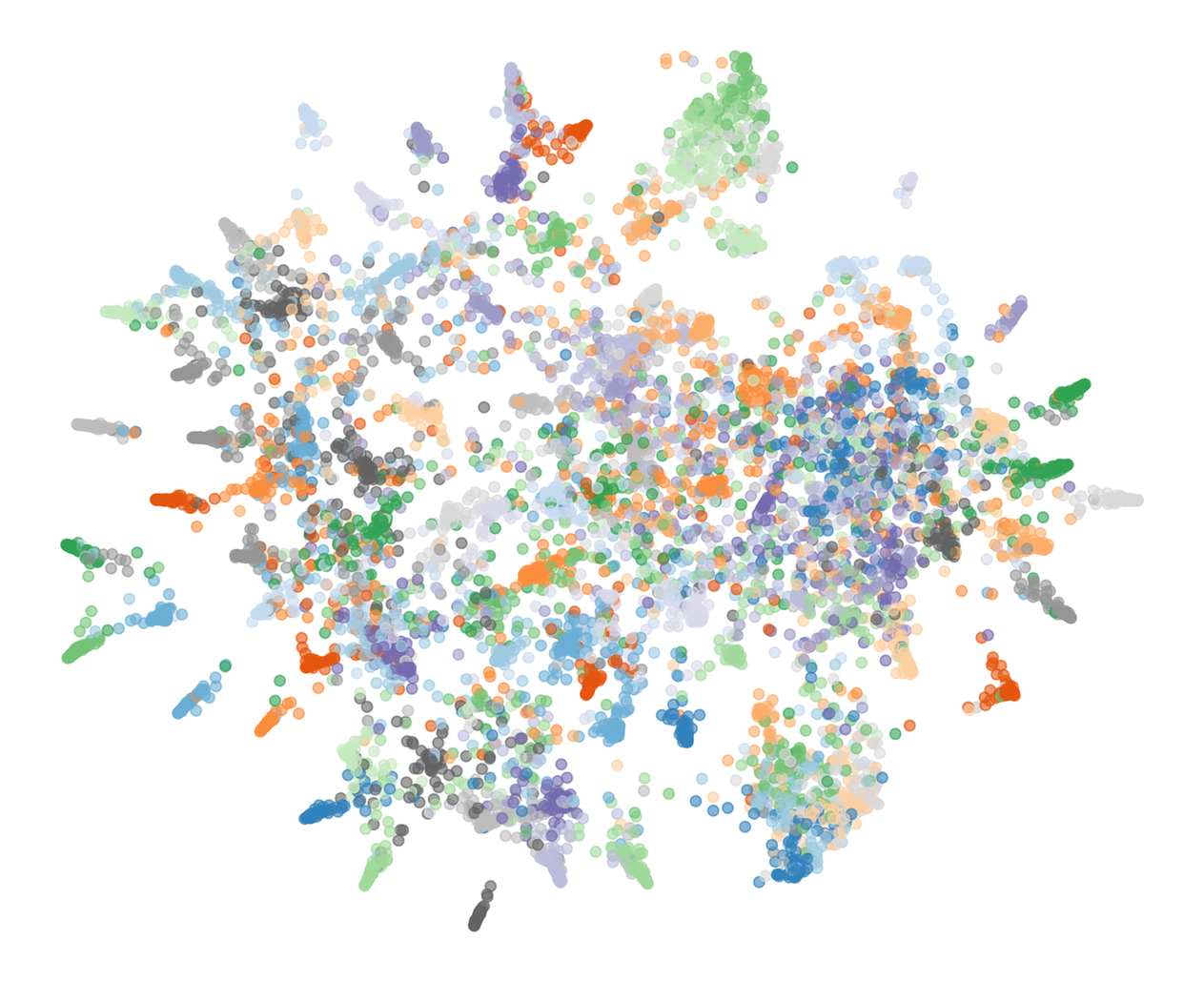}
  \subcaption{CRD \cite{tian2022crd}}
\end{subfigure}
\hfill
\begin{subfigure}[b]{0.16\textwidth}
  \centering
  \includegraphics[width=\linewidth]{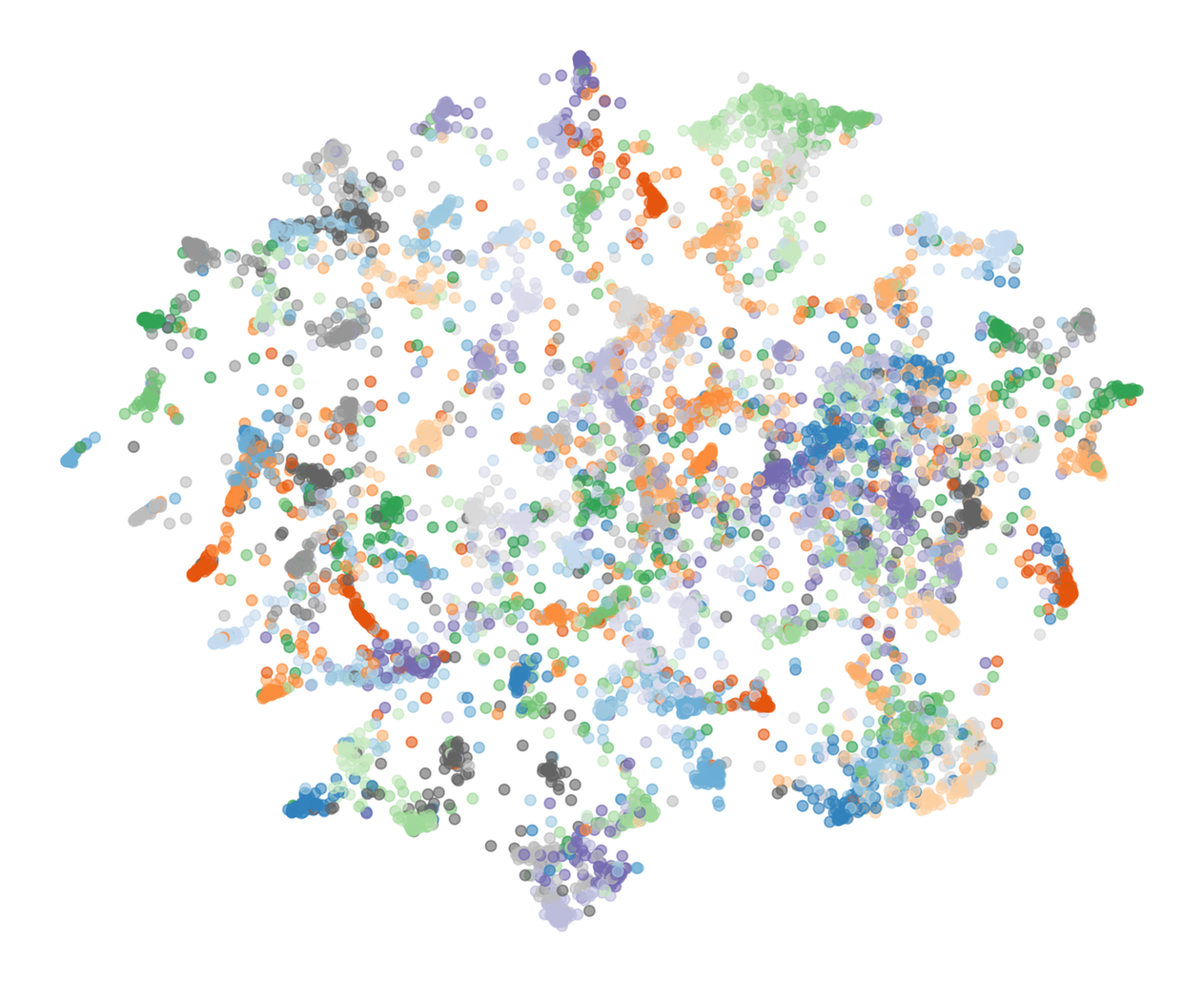}
  \subcaption{CRD+KD \cite{tian2022crd}}
\end{subfigure}
\hfill
\begin{subfigure}[b]{0.16\textwidth}
  \centering
  \includegraphics[width=\linewidth]{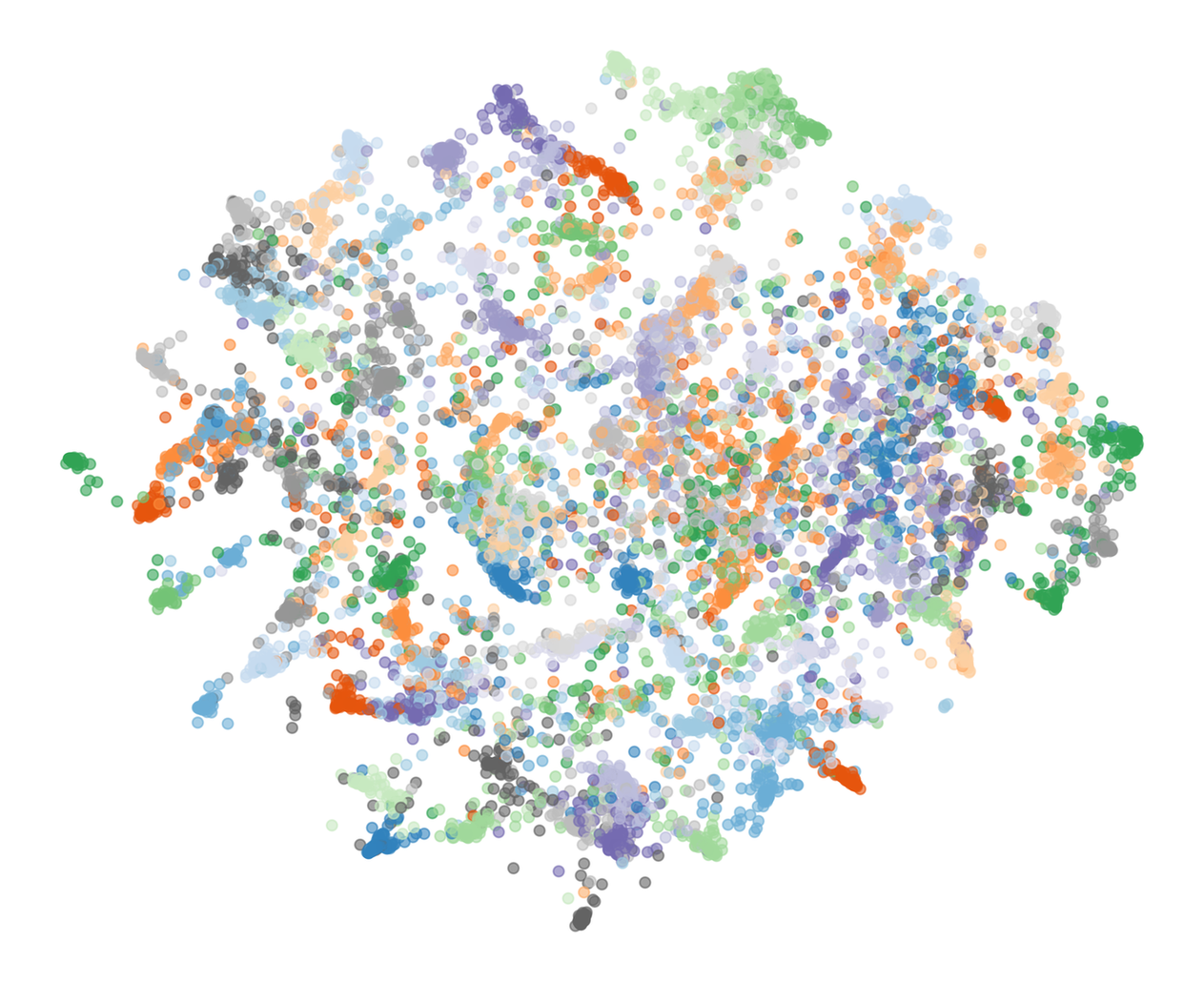}
  \subcaption{\model}
\end{subfigure}
\hfill
\begin{subfigure}[b]{0.16\textwidth}
  \centering
  \includegraphics[width=\linewidth]{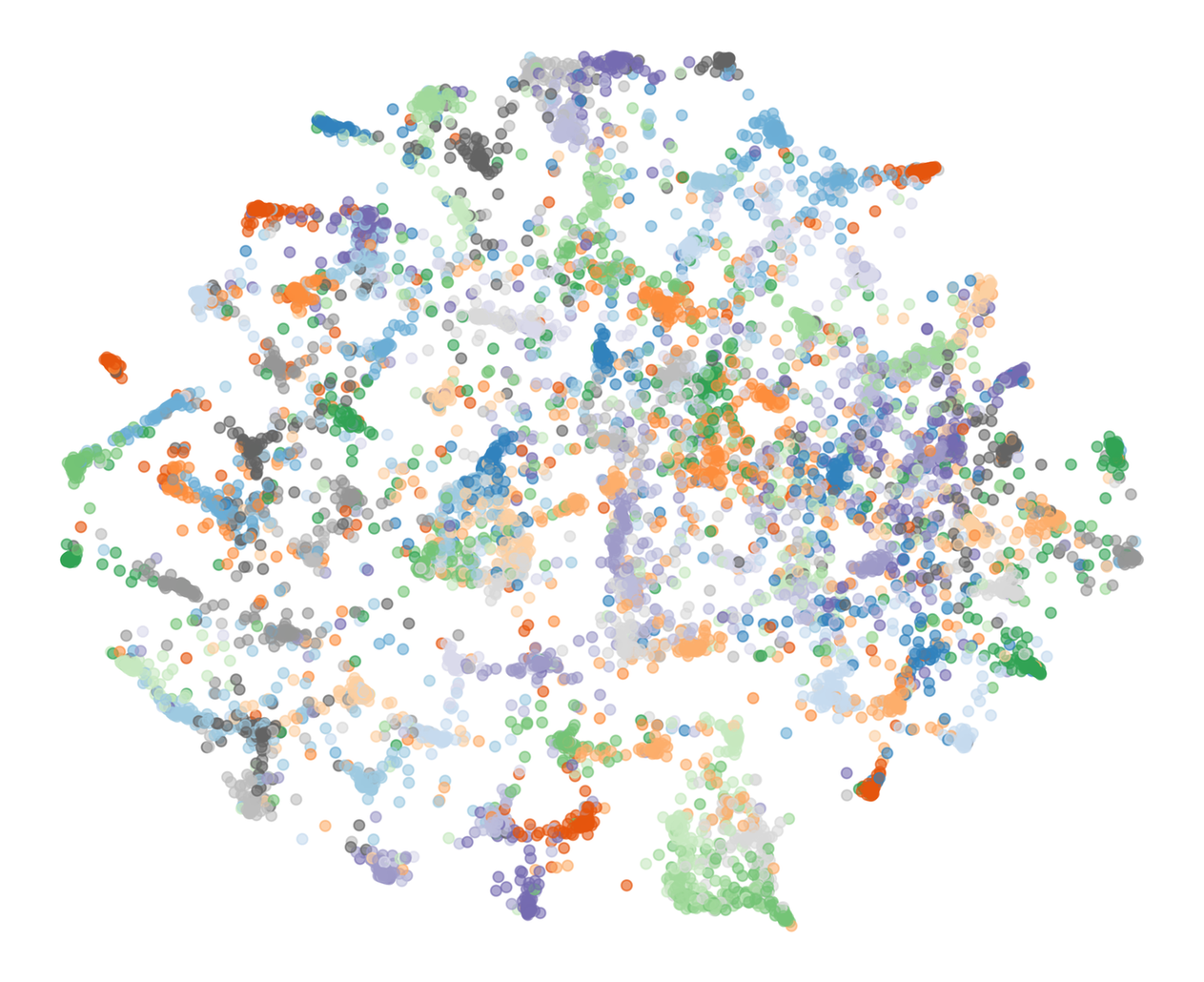}
  \subcaption{\model+KD}
\end{subfigure}
\caption{\textbf{t-SNE visualization of embedding spaces.} Comparison of feature distributions from the teacher and student networks for all 100 classes of CIFAR-100. We use a WRN-40-2 teacher and a WRN-40-1 student for the visualization.}
\label{fig:tsne2_appendix}
\end{figure*}

\Cref{fig:tsne1_appendix,fig:tsne2_appendix} present t-SNE visualizations of embeddings from the teacher (WRN-40-2) and student (WRN-40-1) networks on CIFAR-100. Specifically, \Cref{fig:tsne1_appendix} shows the t-SNE visualization focused on just the first 20 classes for clearer interpretation, and \Cref{fig:tsne2_appendix} shows the same visualizations for all 100 classes of CIFAR-100.

\clearpage
\begin{table*}[!p]
\centering
\setlength{\tabcolsep}{0.1mm}
\caption{\textbf{Results on CIFAR-100 (same architecture).} Test top-1 accuracy (\%) of student networks for same-architecture distillation. \textcolor{green}{$\uparrow$} and \textcolor{red}{$\downarrow$} denote performance relative to KD. Results for our method are averaged over five independent runs.}
\resizebox{\textwidth}{!}{
\begin{tabular}{lccccccc}
\toprule
Teacher & WRN-40-2 & WRN-40-2 & rn-56 & rn-110 & rn-110 & rn-32x4 & VGG-13 \\
Student & WRN-16-2 & WRN-40-1 & rn-20 & rn-20 & rn-32 & rn-8x4 & VGG-8 \\ 
\midrule
\textit{Teacher} & 75.61 & 75.61 & 72.34 & 74.31 & 74.31 & 79.42 & 74.64 \\
\textit{Student} & 73.26 & 71.98 & 69.06 & 69.06 & 71.14 & 72.50 & 70.36 \\ 
KD \cite{hinton2015distilling} & 74.92 & 73.54 & 70.66 & 70.67 & 73.08 & 73.33 & 72.98 \\
FitNet \cite{romero2014fitnets} & 73.58 (\textcolor{red}{$\downarrow$}) & 72.24 (\textcolor{red}{$\downarrow$}) & 69.21 (\textcolor{red}{$\downarrow$}) & 68.99 (\textcolor{red}{$\downarrow$}) & 71.06 (\textcolor{red}{$\downarrow$}) & 73.50 (\textcolor{green}{$\uparrow$}) & 71.02 (\textcolor{red}{$\downarrow$}) \\
AT \cite{zagoruyko2016paying} & 74.08 (\textcolor{red}{$\downarrow$}) & 72.77 (\textcolor{red}{$\downarrow$}) & 70.55 (\textcolor{red}{$\downarrow$}) & 70.22 (\textcolor{red}{$\downarrow$}) & 72.31 (\textcolor{red}{$\downarrow$}) & 73.44 (\textcolor{green}{$\uparrow$}) & 71.43 (\textcolor{red}{$\downarrow$}) \\
SP \cite{tung2019similarity} & 73.83 (\textcolor{red}{$\downarrow$}) & 72.43 (\textcolor{red}{$\downarrow$}) & 69.67 (\textcolor{red}{$\downarrow$}) & 70.04 (\textcolor{red}{$\downarrow$}) & 72.69 (\textcolor{red}{$\downarrow$}) & 72.94 (\textcolor{red}{$\downarrow$}) & 72.68 (\textcolor{red}{$\downarrow$}) \\
CC \cite{peng2019correlation} & 73.56 (\textcolor{red}{$\downarrow$}) & 72.21 (\textcolor{red}{$\downarrow$}) & 69.63 (\textcolor{red}{$\downarrow$}) & 69.48 (\textcolor{red}{$\downarrow$}) & 71.48 (\textcolor{red}{$\downarrow$}) & 72.97 (\textcolor{red}{$\downarrow$}) & 70.81 (\textcolor{red}{$\downarrow$}) \\
VID \cite{ahn2019variational} & 74.11 (\textcolor{red}{$\downarrow$}) & 73.30 (\textcolor{red}{$\downarrow$}) & 70.38 (\textcolor{red}{$\downarrow$}) & 70.16 (\textcolor{red}{$\downarrow$}) & 72.61 (\textcolor{red}{$\downarrow$}) & 73.09 (\textcolor{red}{$\downarrow$}) & 71.23 (\textcolor{red}{$\downarrow$}) \\
RKD \cite{park2019relational} & 73.35 (\textcolor{red}{$\downarrow$}) & 72.22 (\textcolor{red}{$\downarrow$}) & 69.61 (\textcolor{red}{$\downarrow$}) & 69.25 (\textcolor{red}{$\downarrow$}) & 71.82 (\textcolor{red}{$\downarrow$}) & 71.90 (\textcolor{red}{$\downarrow$}) & 71.48 (\textcolor{red}{$\downarrow$}) \\
PKT \cite{passalis2018probabilistic} & 74.54 (\textcolor{red}{$\downarrow$}) & 73.45 (\textcolor{red}{$\downarrow$}) & 70.34 (\textcolor{red}{$\downarrow$}) & 70.25 (\textcolor{red}{$\downarrow$}) & 72.61 (\textcolor{red}{$\downarrow$}) & 73.64 (\textcolor{green}{$\uparrow$}) & 72.88 (\textcolor{red}{$\downarrow$}) \\
AB \cite{heo2019knowledge} & 72.50 (\textcolor{red}{$\downarrow$}) & 72.38 (\textcolor{red}{$\downarrow$}) & 69.47 (\textcolor{red}{$\downarrow$}) & 69.53 (\textcolor{red}{$\downarrow$}) & 70.98 (\textcolor{red}{$\downarrow$}) & 73.17 (\textcolor{red}{$\downarrow$}) & 70.94 (\textcolor{red}{$\downarrow$}) \\
FT \cite{kim2018paraphrasing} & 73.25 (\textcolor{red}{$\downarrow$}) & 71.59 (\textcolor{red}{$\downarrow$}) & 69.84 (\textcolor{red}{$\downarrow$}) & 70.22 (\textcolor{red}{$\downarrow$}) & 72.37 (\textcolor{red}{$\downarrow$}) & 72.86 (\textcolor{red}{$\downarrow$}) & 70.58 (\textcolor{red}{$\downarrow$}) \\
FSP \cite{yim2017gift} & 72.91 (\textcolor{red}{$\downarrow$}) & n/a & 69.95 (\textcolor{red}{$\downarrow$}) & 70.11 (\textcolor{red}{$\downarrow$}) & 71.89 (\textcolor{red}{$\downarrow$}) & 72.62 (\textcolor{red}{$\downarrow$}) & 70.33 (\textcolor{red}{$\downarrow$}) \\
NST \cite{huang2017like} & 73.68 (\textcolor{red}{$\downarrow$}) & 72.24 (\textcolor{red}{$\downarrow$}) & 69.60 (\textcolor{red}{$\downarrow$}) & 69.53 (\textcolor{red}{$\downarrow$}) & 71.96 (\textcolor{red}{$\downarrow$}) & 73.30 (\textcolor{red}{$\downarrow$}) & 71.53 (\textcolor{red}{$\downarrow$}) \\
CRD \cite{tian2022crd} & 75.48 (\textcolor{green}{$\uparrow$}) & 74.14 (\textcolor{green}{$\uparrow$}) & 71.16 (\textcolor{green}{$\uparrow$}) & 71.46 (\textcolor{green}{$\uparrow$}) & 73.48 (\textcolor{green}{$\uparrow$}) & 75.51 (\textcolor{green}{$\uparrow$}) & 73.94 (\textcolor{green}{$\uparrow$}) \\
CRD+KD \cite{tian2022crd} & 75.64 (\textcolor{green}{$\uparrow$}) & 74.38 (\textcolor{green}{$\uparrow$}) & 71.63 (\textcolor{green}{$\uparrow$}) & 71.56 (\textcolor{green}{$\uparrow$}) & 73.75 (\textcolor{green}{$\uparrow$}) & 75.46 (\textcolor{green}{$\uparrow$}) & 74.29 (\textcolor{green}{$\uparrow$}) \\
OFD \cite{heo2019ofd} & 75.24 (\textcolor{green}{$\uparrow$}) & 74.33 (\textcolor{green}{$\uparrow$}) & 70.38 (\textcolor{red}{$\downarrow$}) & n/a & 73.23 (\textcolor{green}{$\uparrow$}) & 74.95 (\textcolor{green}{$\uparrow$}) & 73.95 (\textcolor{green}{$\uparrow$}) \\
WSLD \cite{zhou2021wsld} & n/a & 73.74 (\textcolor{green}{$\uparrow$}) & 71.53 (\textcolor{green}{$\uparrow$}) & n/a & 73.36 (\textcolor{green}{$\uparrow$}) & 74.79 (\textcolor{green}{$\uparrow$}) & n/a \\
IPWD \cite{niu2022ipwd} & n/a & 74.64 (\textcolor{green}{$\uparrow$}) & 71.32 (\textcolor{green}{$\uparrow$}) & n/a & 73.91 (\textcolor{green}{$\uparrow$}) & 76.03 (\textcolor{green}{$\uparrow$}) & n/a \\
SRRL \cite{yang2021srrl} & n/a & 74.64 (\textcolor{green}{$\uparrow$}) & n/a & n/a & n/a & 75.39 (\textcolor{green}{$\uparrow$}) & n/a \\
SemCKD \cite{chen2021semckd} & n/a & 74.41 (\textcolor{green}{$\uparrow$}) & n/a & n/a & n/a & 76.23 (\textcolor{green}{$\uparrow$}) & n/a \\
ReviewKD \cite{chen2021reviewkd} & 76.12 (\textcolor{green}{$\uparrow$}) & 75.09 (\textcolor{green}{$\uparrow$}) & 71.89 (\textcolor{green}{$\uparrow$}) & n/a & 73.89 (\textcolor{green}{$\uparrow$}) & 75.63 (\textcolor{green}{$\uparrow$}) & 74.84 (\textcolor{green}{$\uparrow$}) \\
SimKD \cite{chen2022simkd} & n/a & 75.56 (\textcolor{green}{$\uparrow$}) & n/a & n/a & n/a & {78.08} (\textcolor{green}{$\uparrow$}) & n/a \\
DistPro \cite{deng2022distpro} & 76.36 (\textcolor{green}{$\uparrow$}) & n/a & 72.03 (\textcolor{green}{$\uparrow$}) & n/a & 73.74 (\textcolor{green}{$\uparrow$}) & n/a & n/a \\
NORM \cite{liu2023norm} & 75.65 (\textcolor{green}{$\uparrow$}) & 74.82 (\textcolor{green}{$\uparrow$}) & 71.35 (\textcolor{green}{$\uparrow$}) & 71.55 (\textcolor{green}{$\uparrow$}) & 73.67 (\textcolor{green}{$\uparrow$}) & 76.49 (\textcolor{green}{$\uparrow$}) & 73.95 (\textcolor{green}{$\uparrow$}) \\
NORM+KD \cite{liu2023norm} & 76.26 (\textcolor{green}{$\uparrow$}) & 75.42 (\textcolor{green}{$\uparrow$}) & 71.61 (\textcolor{green}{$\uparrow$}) & 72.00 (\textcolor{green}{$\uparrow$}) & 74.95 (\textcolor{green}{$\uparrow$}) & 76.98 (\textcolor{green}{$\uparrow$}) & 74.46 (\textcolor{green}{$\uparrow$}) \\
NORM+CRD \cite{liu2023norm} & 76.02 (\textcolor{green}{$\uparrow$}) & 75.37 (\textcolor{green}{$\uparrow$}) & 71.51 (\textcolor{green}{$\uparrow$}) & 71.90 (\textcolor{green}{$\uparrow$}) & 73.81 (\textcolor{green}{$\uparrow$}) & 76.49 (\textcolor{green}{$\uparrow$}) & 73.58 (\textcolor{green}{$\uparrow$}) \\
WCoRD \cite{chen2021wcord} & 75.88 (\textcolor{green}{$\uparrow$}) & 74.73 (\textcolor{green}{$\uparrow$}) & 71.56 (\textcolor{green}{$\uparrow$}) & 71.57 (\textcolor{green}{$\uparrow$}) & 73.81 (\textcolor{green}{$\uparrow$}) & 75.95 (\textcolor{green}{$\uparrow$}) & 74.55 (\textcolor{green}{$\uparrow$}) \\
WCoRD+KD \cite{chen2021wcord} & 76.11 (\textcolor{green}{$\uparrow$}) & 74.72 (\textcolor{green}{$\uparrow$}) & 71.92 (\textcolor{green}{$\uparrow$}) & 71.88 (\textcolor{green}{$\uparrow$}) & 74.20 (\textcolor{green}{$\uparrow$}) & 76.15 (\textcolor{green}{$\uparrow$}) & 74.72 (\textcolor{green}{$\uparrow$}) \\
CRCD \cite{zhu2021crcd} & {76.67} (\textcolor{green}{$\uparrow$}) & {75.95} (\textcolor{green}{$\uparrow$}) & {73.21} (\textcolor{green}{$\uparrow$}) & {72.33} (\textcolor{green}{$\uparrow$}) & {74.98} (\textcolor{green}{$\uparrow$}) & 76.42 (\textcolor{green}{$\uparrow$}) & {74.97} (\textcolor{green}{$\uparrow$}) \\
FKD \cite{he2022fkd} & n/a & n/a & n/a & n/a & n/a & 75.57 (\textcolor{green}{$\uparrow$}) & 73.78 (\textcolor{green}{$\uparrow$}) \\
ITRD (corr) \cite{miles2022itrd} & 75.85 (\textcolor{green}{$\uparrow$}) & 74.90 (\textcolor{green}{$\uparrow$}) & 71.45 (\textcolor{green}{$\uparrow$}) & 71.77 (\textcolor{green}{$\uparrow$}) & 74.02 (\textcolor{green}{$\uparrow$}) & 75.63 (\textcolor{green}{$\uparrow$}) & 74.70 (\textcolor{green}{$\uparrow$}) \\
ITRD (corr+mi) \cite{miles2022itrd} & 76.12 (\textcolor{green}{$\uparrow$}) & 75.18 (\textcolor{green}{$\uparrow$}) & 71.47 (\textcolor{green}{$\uparrow$}) & 71.99 (\textcolor{green}{$\uparrow$}) & 74.26 (\textcolor{green}{$\uparrow$}) & 76.19 (\textcolor{green}{$\uparrow$}) & 74.93 (\textcolor{green}{$\uparrow$}) \\
TTM \cite{zheng2024ttm} & 76.23 (\textcolor{green}{$\uparrow$}) & 74.32 (\textcolor{green}{$\uparrow$}) & 71.83 (\textcolor{green}{$\uparrow$}) & 71.46 (\textcolor{green}{$\uparrow$}) & 73.97 (\textcolor{green}{$\uparrow$}) & 76.17 (\textcolor{green}{$\uparrow$}) & 74.33 (\textcolor{green}{$\uparrow$})\\
WTTM \cite{zheng2024ttm} & 76.37 (\textcolor{green}{$\uparrow$}) & 74.58 (\textcolor{green}{$\uparrow$}) & 71.92 (\textcolor{green}{$\uparrow$}) & 71.67 (\textcolor{green}{$\uparrow$}) & 74.13 (\textcolor{green}{$\uparrow$}) & 76.06 (\textcolor{green}{$\uparrow$}) & 74.44 (\textcolor{green}{$\uparrow$})\\
WTTM+CRD \cite{zheng2024ttm} & 76.61 (\textcolor{green}{$\uparrow$}) & 74.94 (\textcolor{green}{$\uparrow$}) & 72.20 (\textcolor{green}{$\uparrow$}) & 72.13 (\textcolor{green}{$\uparrow$}) & 74.52 (\textcolor{green}{$\uparrow$}) & 76.65 (\textcolor{green}{$\uparrow$}) & 74.71 (\textcolor{green}{$\uparrow$})\\
WTTM+ITRD \cite{zheng2024ttm} & 76.65 (\textcolor{green}{$\uparrow$}) & 75.34 (\textcolor{green}{$\uparrow$}) & 72.16 (\textcolor{green}{$\uparrow$}) & 72.20 (\textcolor{green}{$\uparrow$}) & 74.36 (\textcolor{green}{$\uparrow$}) & 77.36 (\textcolor{green}{$\uparrow$}) & 75.13 (\textcolor{green}{$\uparrow$})\\
DKD \cite{zhao2022dkd} & 76.24 (\textcolor{green}{$\uparrow$}) & 74.81 (\textcolor{green}{$\uparrow$}) & 71.97 (\textcolor{green}{$\uparrow$}) & n/a  &74.11 (\textcolor{green}{$\uparrow$}) & 76.32 (\textcolor{green}{$\uparrow$}) & 74.68 (\textcolor{green}{$\uparrow$}) \\
FCFD \cite{liu2023fcfd} & 76.34 (\textcolor{green}{$\uparrow$}) & 75.43 (\textcolor{green}{$\uparrow$}) & 71.68 (\textcolor{green}{$\uparrow$}) & n/a & n/a & 76.80 (\textcolor{green}{$\uparrow$}) & 74.86 (\textcolor{green}{$\uparrow$})\\
FCFD+KD \cite{liu2023fcfd} & 76.43 (\textcolor{green}{$\uparrow$}) & 75.46 (\textcolor{green}{$\uparrow$}) & 71.96 (\textcolor{green}{$\uparrow$}) & n/a & n/a & 76.62 (\textcolor{green}{$\uparrow$}) & 75.22 (\textcolor{green}{$\uparrow$})\\
CAT-KD \cite{guo2023cat} & 75.60 (\textcolor{green}{$\uparrow$}) & 74.82 (\textcolor{green}{$\uparrow$}) & 71.62 (\textcolor{green}{$\uparrow$}) & n/a & 73.62 (\textcolor{green}{$\uparrow$}) & 76.91 (\textcolor{green}{$\uparrow$}) & 74.65 (\textcolor{green}{$\uparrow$})\\
DIST \cite{huang2022dist} & n/a & 74.73 (\textcolor{green}{$\uparrow$}) & 71.75 (\textcolor{green}{$\uparrow$}) & n/a & n/a & 76.31 (\textcolor{green}{$\uparrow$}) & n/a \\
CTKD \cite{li2022ctkd} & 75.45 (\textcolor{green}{$\uparrow$}) & 73.93 (\textcolor{green}{$\uparrow$}) & 71.19 (\textcolor{green}{$\uparrow$}) & 70.99 (\textcolor{green}{$\uparrow$}) & 73.52 (\textcolor{green}{$\uparrow$}) & n/a & 73.52 (\textcolor{green}{$\uparrow$})\\
\model (ours) & 74.99 (\textcolor{green}{$\uparrow$}) & 73.69 (\textcolor{green}{$\uparrow$}) & 71.18 (\textcolor{green}{$\uparrow$}) & 71.00 (\textcolor{green}{$\uparrow$}) & 73.12 (\textcolor{green}{$\uparrow$}) & 74.23 (\textcolor{green}{$\uparrow$}) & 73.22 (\textcolor{green}{$\uparrow$}) \\
\model+KD (ours) & {76.06} (\textcolor{green}{$\uparrow$}) & {74.76} (\textcolor{green}{$\uparrow$}) & {71.81} (\textcolor{green}{$\uparrow$}) & {72.03} (\textcolor{green}{$\uparrow$}) & {73.62} (\textcolor{green}{$\uparrow$}) & 75.09 (\textcolor{green}{$\uparrow$}) & {73.95} (\textcolor{green}{$\uparrow$}) \\ 
\hline
\end{tabular}
}%
\label{tab:cifar100_same_appendix}
\end{table*}

\begin{table*}[!p]
\centering
\caption{\textbf{Results on CIFAR-100 (different architecture).} Test top-1 accuracy (\%) for teacher-student pairs with different architectures. \textcolor{green}{$\uparrow$} and \textcolor{red}{$\downarrow$} denote performance relative to KD. Results for our method are averaged over five independent runs.}
\begin{tabular}{lcccccc}
\toprule
Teacher & VGG-13 & RN-50 & RN-50 & RN-32x4 & RN-32x4 & WRN-40-2 \\
Student & MN-v2 & MN-v2 & VGG-8 & SN-v1 & SN-v2 & SN-v1 \\ 
\midrule
\textit{Teacher} & 74.64 & 79.34 & 79.34 & 79.42 & 79.42 & 75.61 \\
\textit{Student} & 64.60 & 64.60 & 70.36 & 70.5 & 71.82 & 70.5 \\ 
KD \cite{hinton2015distilling} & 67.37 & 67.35 & 73.81 & 74.07 & 74.45 & 74.83 \\
FitNet \cite{romero2014fitnets} & 64.14 (\textcolor{red}{$\downarrow$}) & 63.16 (\textcolor{red}{$\downarrow$}) & 70.69 (\textcolor{red}{$\downarrow$}) & 73.59 (\textcolor{red}{$\downarrow$}) & 73.54 (\textcolor{red}{$\downarrow$}) & 73.73 (\textcolor{red}{$\downarrow$}) \\
AT \cite{zagoruyko2016paying} & 59.40 (\textcolor{red}{$\downarrow$}) & 58.58 (\textcolor{red}{$\downarrow$}) & 71.84 (\textcolor{red}{$\downarrow$}) & 71.73 (\textcolor{red}{$\downarrow$}) & 72.73 (\textcolor{red}{$\downarrow$}) & 73.32 (\textcolor{red}{$\downarrow$}) \\
SP \cite{tung2019similarity} & 66.30 (\textcolor{red}{$\downarrow$}) & 68.08 (\textcolor{green}{$\uparrow$}) & 73.34 (\textcolor{red}{$\downarrow$}) & 73.48 (\textcolor{red}{$\downarrow$}) & 74.56 (\textcolor{green}{$\uparrow$}) & 74.52 (\textcolor{red}{$\downarrow$}) \\
CC \cite{peng2019correlation} & 64.86 (\textcolor{red}{$\downarrow$}) & 65.43 (\textcolor{red}{$\downarrow$}) & 70.25 (\textcolor{red}{$\downarrow$}) & 71.14 (\textcolor{red}{$\downarrow$}) & 71.29 (\textcolor{red}{$\downarrow$}) & 71.38 (\textcolor{red}{$\downarrow$}) \\
VID \cite{ahn2019variational} & 65.56 (\textcolor{red}{$\downarrow$}) & 67.57 (\textcolor{green}{$\uparrow$}) & 70.30 (\textcolor{red}{$\downarrow$}) & 73.38 (\textcolor{red}{$\downarrow$}) & 73.40 (\textcolor{red}{$\downarrow$}) & 73.61 (\textcolor{red}{$\downarrow$}) \\
RKD \cite{park2019relational} & 64.52 (\textcolor{red}{$\downarrow$}) & 64.43 (\textcolor{red}{$\downarrow$}) & 71.50 (\textcolor{red}{$\downarrow$}) & 72.28 (\textcolor{red}{$\downarrow$}) & 73.21 (\textcolor{red}{$\downarrow$}) & 72.21 (\textcolor{red}{$\downarrow$}) \\
PKT \cite{passalis2018probabilistic} & 67.13 (\textcolor{red}{$\downarrow$}) & 66.52 (\textcolor{red}{$\downarrow$}) & 73.01 (\textcolor{red}{$\downarrow$}) & 74.10 (\textcolor{green}{$\uparrow$}) & 74.69 (\textcolor{green}{$\uparrow$}) & 73.89 (\textcolor{red}{$\downarrow$}) \\
AB \cite{heo2019knowledge} & 66.06 (\textcolor{red}{$\downarrow$}) & 67.20 (\textcolor{red}{$\downarrow$}) & 70.65 (\textcolor{red}{$\downarrow$}) & 73.55 (\textcolor{red}{$\downarrow$}) & 74.31 (\textcolor{red}{$\downarrow$}) & 73.34 (\textcolor{red}{$\downarrow$}) \\
FT \cite{kim2018paraphrasing} & 61.78 (\textcolor{red}{$\downarrow$}) & 60.99 (\textcolor{red}{$\downarrow$}) & 70.29 (\textcolor{red}{$\downarrow$}) & 71.75 (\textcolor{red}{$\downarrow$}) & 72.50 (\textcolor{red}{$\downarrow$}) & 72.03 (\textcolor{red}{$\downarrow$}) \\
NST \cite{huang2017like} & 58.16 (\textcolor{red}{$\downarrow$}) & 64.96 (\textcolor{red}{$\downarrow$}) & 71.28 (\textcolor{red}{$\downarrow$}) & 74.12 (\textcolor{green}{$\uparrow$}) & 74.68 (\textcolor{green}{$\uparrow$}) & 76.09 (\textcolor{green}{$\uparrow$}) \\
CRD \cite{tian2022crd} & 69.73 (\textcolor{green}{$\uparrow$}) & 69.11 (\textcolor{green}{$\uparrow$}) & 74.3 (\textcolor{green}{$\uparrow$}) & 75.11 (\textcolor{green}{$\uparrow$}) & 75.65 (\textcolor{green}{$\uparrow$}) & 76.05 (\textcolor{green}{$\uparrow$}) \\
CRD+KD \cite{tian2022crd} & 69.94 (\textcolor{green}{$\uparrow$}) & 69.54 (\textcolor{green}{$\uparrow$}) & 74.58 (\textcolor{green}{$\uparrow$}) & 75.12 (\textcolor{green}{$\uparrow$}) & 76.05 (\textcolor{green}{$\uparrow$}) & 76.27 (\textcolor{green}{$\uparrow$}) \\
OFD \cite{heo2019ofd} & 69.48 (\textcolor{green}{$\uparrow$}) & 69.04 (\textcolor{green}{$\uparrow$}) & n/a & 75.98 (\textcolor{green}{$\uparrow$}) & 76.82 (\textcolor{green}{$\uparrow$}) & 75.85 (\textcolor{green}{$\uparrow$}) \\
WSLD \cite{zhou2021wsld} & n/a & 68.79 (\textcolor{green}{$\uparrow$}) & 73.80 (\textcolor{red}{$\downarrow$}) & 75.09 (\textcolor{green}{$\uparrow$}) & n/a & 75.23 (\textcolor{green}{$\uparrow$}) \\
IPWD \cite{niu2022ipwd} & n/a & 70.25 (\textcolor{green}{$\uparrow$}) & 74.97 (\textcolor{green}{$\uparrow$}) & 76.03 (\textcolor{green}{$\uparrow$}) & n/a & 76.44 (\textcolor{green}{$\uparrow$}) \\
SRRL \cite{yang2021srrl} & n/a & n/a & n/a & 75.18 (\textcolor{green}{$\uparrow$}) & n/a & n/a \\
SemCKD \cite{chen2021semckd} & n/a & n/a & n/a & n/a & 77.62 (\textcolor{green}{$\uparrow$}) & n/a \\
ReviewKD \cite{chen2021reviewkd} & 70.37 (\textcolor{green}{$\uparrow$}) & 69.89 (\textcolor{green}{$\uparrow$}) & n/a & 77.45 (\textcolor{green}{$\uparrow$}) & 77.78 (\textcolor{green}{$\uparrow$}) & 77.14 (\textcolor{green}{$\uparrow$}) \\
SimKD \cite{chen2022simkd} & n/a & n/a & n/a & 77.18 (\textcolor{green}{$\uparrow$}) & n/a & n/a \\
DistPro \cite{deng2022distpro} & n/a & n/a & n/a & 77.18 (\textcolor{green}{$\uparrow$}) & 77.54 (\textcolor{green}{$\uparrow$}) & 77.24 (\textcolor{green}{$\uparrow$}) \\
NORM \cite{liu2023norm} & 68.94 (\textcolor{green}{$\uparrow$}) & 70.56 (\textcolor{green}{$\uparrow$}) & 75.17 (\textcolor{green}{$\uparrow$}) & 77.42 (\textcolor{green}{$\uparrow$}) & 78.07 (\textcolor{green}{$\uparrow$}) & 77.06 (\textcolor{green}{$\uparrow$}) \\
NORM+KD \cite{liu2023norm} & 69.38 (\textcolor{green}{$\uparrow$}) & 71.17 (\textcolor{green}{$\uparrow$}) & 75.67 (\textcolor{green}{$\uparrow$}) & {77.79} (\textcolor{green}{$\uparrow$}) & {78.32} (\textcolor{green}{$\uparrow$}) & {77.63} (\textcolor{green}{$\uparrow$}) \\
NORM+CRD \cite{liu2023norm} & 69.17 (\textcolor{green}{$\uparrow$}) & 71.08 (\textcolor{green}{$\uparrow$}) & 75.51 (\textcolor{green}{$\uparrow$}) & 77.50 (\textcolor{green}{$\uparrow$}) & 77.96 (\textcolor{green}{$\uparrow$}) & 77.09 (\textcolor{green}{$\uparrow$}) \\
WCoRD \cite{chen2021wcord} & 69.47 (\textcolor{green}{$\uparrow$}) & 70.45 (\textcolor{green}{$\uparrow$}) & 74.86 (\textcolor{green}{$\uparrow$}) & 75.40 (\textcolor{green}{$\uparrow$}) & 75.96 (\textcolor{green}{$\uparrow$}) & 76.32 (\textcolor{green}{$\uparrow$}) \\
WCoRD+KD \cite{chen2021wcord} & 70.02 (\textcolor{green}{$\uparrow$}) & 70.12 (\textcolor{green}{$\uparrow$}) & 74.68 (\textcolor{green}{$\uparrow$}) & 75.77 (\textcolor{green}{$\uparrow$}) & 76.48 (\textcolor{green}{$\uparrow$}) & 76.68 (\textcolor{green}{$\uparrow$}) \\
CRCD \cite{zhu2021crcd} & n/a & n/a & n/a & n/a & n/a & n/a \\
FKD \cite{he2022fkd} & n/a & n/a & 74.61 (\textcolor{green}{$\uparrow$}) & 75 (\textcolor{green}{$\uparrow$}) & n/a & n/a \\
ITRD (corr) \cite{miles2022itrd} & 69.97 (\textcolor{green}{$\uparrow$}) & {71.41} (\textcolor{green}{$\uparrow$}) & {75.71} (\textcolor{green}{$\uparrow$}) & 76.8 (\textcolor{green}{$\uparrow$}) & 77.27 (\textcolor{green}{$\uparrow$}) & 77.35 (\textcolor{green}{$\uparrow$}) \\
ITRD (corr+mi) \cite{miles2022itrd} & {70.39} (\textcolor{green}{$\uparrow$}) & 71.34 (\textcolor{green}{$\uparrow$}) & 75.49 (\textcolor{green}{$\uparrow$}) & 76.91 (\textcolor{green}{$\uparrow$}) & 77.40 (\textcolor{green}{$\uparrow$}) & 77.09 (\textcolor{green}{$\uparrow$}) \\ 
TTM \cite{zheng2024ttm} & 68.98 (\textcolor{green}{$\uparrow$}) & 69.24 (\textcolor{green}{$\uparrow$}) & 74.87 (\textcolor{green}{$\uparrow$}) & 74.18 (\textcolor{green}{$\uparrow$}) & 76.57 (\textcolor{green}{$\uparrow$}) & 75.39 (\textcolor{green}{$\uparrow$})\\
WTTM \cite{zheng2024ttm} & 69.16 (\textcolor{green}{$\uparrow$}) & 69.59 (\textcolor{green}{$\uparrow$}) & 74.82 (\textcolor{green}{$\uparrow$}) & 74.37 (\textcolor{green}{$\uparrow$}) & 76.55 (\textcolor{green}{$\uparrow$}) & 75.42 (\textcolor{green}{$\uparrow$})\\
WTTM+CRD \cite{zheng2024ttm} & 70.30 (\textcolor{green}{$\uparrow$}) & 70.84 (\textcolor{green}{$\uparrow$}) & 75.30 (\textcolor{green}{$\uparrow$}) & 75.82 (\textcolor{green}{$\uparrow$}) & 77.04 (\textcolor{green}{$\uparrow$}) & 76.86 (\textcolor{green}{$\uparrow$})\\
WTTM+ITRD \cite{zheng2024ttm} & 70.70 (\textcolor{green}{$\uparrow$}) & 71.56 (\textcolor{green}{$\uparrow$}) & 76.00 (\textcolor{green}{$\uparrow$}) & 77.03 (\textcolor{green}{$\uparrow$}) & 77.68 (\textcolor{green}{$\uparrow$}) & 77.44 (\textcolor{green}{$\uparrow$})\\
DKD \cite{zhao2022dkd} & 69.71 (\textcolor{green}{$\uparrow$}) & 70.35 (\textcolor{green}{$\uparrow$}) & n/a & 76.45 (\textcolor{green}{$\uparrow$}) & 77.07 (\textcolor{green}{$\uparrow$}) & 76.70 (\textcolor{green}{$\uparrow$})\\
FCFD \cite{liu2023fcfd} & 70.67 (\textcolor{green}{$\uparrow$}) & 71.07 (\textcolor{green}{$\uparrow$}) & n/a & 78.12 (\textcolor{green}{$\uparrow$}) & 78.20 (\textcolor{green}{$\uparrow$}) & 77.81 (\textcolor{green}{$\uparrow$})\\
FCFD+KD \cite{liu2023fcfd} & 70.65 (\textcolor{green}{$\uparrow$}) & 71.00 (\textcolor{green}{$\uparrow$}) & n/a & 78.12 (\textcolor{green}{$\uparrow$}) & 78.18 (\textcolor{green}{$\uparrow$}) & 77.99 (\textcolor{green}{$\uparrow$})\\
CAT-KD \cite{guo2023cat} & 69.13 (\textcolor{green}{$\uparrow$}) & 71.36 (\textcolor{green}{$\uparrow$}) & n/a & 78.26 (\textcolor{green}{$\uparrow$}) & 78.41 (\textcolor{green}{$\uparrow$}) & 77.35 (\textcolor{green}{$\uparrow$})\\
DIST \cite{huang2022dist} & n/a & 68.66 (\textcolor{green}{$\uparrow$}) & n/a & 76.34 (\textcolor{green}{$\uparrow$}) & 77.35 (\textcolor{green}{$\uparrow$}) & n/a \\
CTKD \cite{li2022ctkd} & 68.46 (\textcolor{green}{$\uparrow$}) & 68.47 (\textcolor{green}{$\uparrow$}) & n/a & 74.78 (\textcolor{green}{$\uparrow$}) & 75.31 (\textcolor{green}{$\uparrow$}) & 75.78 (\textcolor{green}{$\uparrow$})\\
\model (ours) & 68.35 (\textcolor{green}{$\uparrow$}) & 67.39 (\textcolor{green}{$\uparrow$}) & 73.85 (\textcolor{green}{$\uparrow$}) & 74.26 (\textcolor{green}{$\uparrow$}) & 75.26 (\textcolor{green}{$\uparrow$}) & 74.98 (\textcolor{green}{$\uparrow$}) \\
\model+KD (ours) & 69.77 (\textcolor{green}{$\uparrow$}) & {70.03} (\textcolor{green}{$\uparrow$}) & 74.08 (\textcolor{green}{$\uparrow$}) & {76.01} (\textcolor{green}{$\uparrow$}) & {76.95} (\textcolor{green}{$\uparrow$}) & {76.51} (\textcolor{green}{$\uparrow$}) \\ 
\hline
\end{tabular}
\label{tab:cifar100_diff_appendix}
\end{table*}



\end{document}